\documentclass[11pt, letterpaper]{article}

\usepackage[left=1in, right=1in, top=1in, bottom=1in]{geometry}
\usepackage[utf8]{inputenc}
\usepackage[T1]{fontenc}
\usepackage{bm}
\usepackage{type1cm}
\usepackage{lettrine}
\usepackage{amsmath,amssymb,amsthm}
\usepackage{moreverb}
\usepackage{mathtools}
\usepackage{amsmath}
\usepackage{amssymb}
\usepackage{algorithmic}
\usepackage{graphics}
\usepackage{graphicx}
\usepackage{subfigure}
\usepackage{caption}
\usepackage{extarrows}
\usepackage{color}
\usepackage{framed}
\usepackage{wrapfig}
\usepackage{bm}
\usepackage{mathrsfs}
\usepackage{mathabx}
\usepackage{multirow}
\usepackage{longtable}
\usepackage{hyperref}
\usepackage{paralist}
\usepackage{indentfirst}
\usepackage{relsize}
\usepackage{extarrows}
\usepackage{upgreek}
\usepackage{bm}
\usepackage{mwe}
\usepackage[dvipsnames,table]{xcolor}
\usepackage{booktabs}
\usepackage{authblk}
\usepackage{lettrine}
\usepackage{type1cm}
\usepackage{threeparttable}
\usepackage[sort&compress,numbers]{natbib}
\usepackage[figurename=Fig.]{caption}
\usepackage[normalem]{ulem}

\usepackage{graphicx}
\usepackage{svg}

\usepackage{multirow}
\usepackage{wrapfig}
\usepackage{array} 
\usepackage{subfigure}
\usepackage{enumerate}
\usepackage{scalerel} 
\usepackage{nameref}

\usepackage{svg}

\DeclareMathOperator*{\concat}{\scalerel*{\Vert}{\sum}}
\newcolumntype{H}{>{\setbox0=\hbox\bgroup}c<{\egroup}@{}}

\graphicspath{ {./Fig./} }
\usepackage[font=footnotesize,labelfont=bf]{caption}
\newcommand{\eref}[1]{(\ref{#1})}

\newcommand{\kredit}[1]{{\color{black} #1}}

\hypersetup{
bookmarks=true,
bookmarksopen=true,
bookmarksnumbered=true,
unicode=false,
pdftoolbar=true,
pdfmenubar=true,
pdffitwindow=false,
pdfstartview={FitH},
pdftitle={My title},
pdfauthor={Author},
pdfsubject={Subject},
pdfcreator={Creator},
pdfproducer={Producer},
pdfkeywords={keywords},
pdfnewwindow=true,
colorlinks=true,
linkcolor=blue,
citecolor=blue,
filecolor=blue,
urlcolor=blue
}

\begin{document}

\title{\textbf{Discovering physical laws with parallel symbolic enumeration}}

\author[1]{Kai Ruan}
\author[1]{Yilong Xu}
\author[1]{Ze-Feng Gao}
\author[2,3]{Yang Liu}
\author[4]{Yike Guo} 
\author[1]{Ji-Rong Wen} 
\author[1,$^*$]{Hao Sun}

\affil[1]{\footnotesize Gaoling School of Artificial Intelligence, Renmin University of China, Beijing, China}
\affil[2]{\footnotesize School of Engineering Science, University of Chinese Academy of Sciences, Beijing, China}
\affil[3]{\footnotesize State Key Laboratory of Nonlinear Mechanics, Institute of Mechanics, Chinese Academy of Sciences, Beijing, China}
\affil[4]{\footnotesize Department of Computer Science and Engineering, HKUST, Hong Kong, China \vspace{18pt}} 

\affil[*]{Corresponding author.\vspace{12pt}}

\date{}

\maketitle

\normalsize

\vspace{-28pt} 
\begin{abstract}
\small

Symbolic regression plays a crucial role in modern scientific research thanks to its capability of discovering concise and interpretable mathematical expressions from data. A key challenge lies in the search for parsimonious and generalizable mathematical formulas, in an infinite search space, while intending to fit the training data. Existing algorithms have faced a critical bottleneck of accuracy and efficiency over a decade when handling problems of complexity, which essentially hinders the pace of applying symbolic regression for scientific exploration across interdisciplinary domains. To this end, we introduce parallel symbolic enumeration (PSE) to efficiently distill generic mathematical expressions from limited data. Experiments show that PSE achieves higher accuracy and faster computation compared to the state-of-the-art baseline algorithms across over 200 synthetic and experimental problem sets (e.g., improving the recovery accuracy by up to 99\% and reducing runtime by an order of magnitude). PSE represents an advance in accurate and efficient data-driven discovery of symbolic, interpretable models (e.g., underlying physical laws), and improves the scalability of symbolic learning.

\end{abstract}

\vspace{12pt}

\section*{Introduction}
\label{sec:introduction}

Over the centuries, scientific discovery has never departed from the use of interpretable mathematical equations or analytical models to describe complex phenomena in nature. Pioneering scientists discovered that behind many sets of empirical data in the real world lay succinct governing equations or physical laws. A famous example of this is Kepler's discovery of three laws of planetary motion using Tycho Brahe's observational data, which laid the foundation for Newton's discovery of universal gravitation. Automated extraction of these natural laws from data, as a class of typical symbolic regression (SR)  problems \cite{schmidt2009distilling, liu2024automated}, stands at the forefront of data-driven scientific exploration in natural sciences and engineering applications \cite{weng2020simple_NC_Catalyst,  cornelio2023combining_NC_AI-Descartes, wadekar2023augmenting_SRfor_astro,li2023electron,Machineguided2023hafner, Machine2023Chong_PNAS, jung2023machine_NCS, golden2023physically, chen2023constructing_NCS}. However, uncovering parsimonious closed-form equations that govern complex systems (e.g., nonlinear dynamics) is always challenging. The data revolution, which is rooted in advanced machine intelligence, has offered an alternative to tackle this issue. Although well-known regression methods \cite{hosmer2013applied} have been widely applied to identify the coefficients of given equations in fixed forms, they are no longer effective for natural systems where our prior knowledge of the explicit model structure is vague.

Attempts have been made to develop evolutionary computational methods to uncover symbolic formulas that best interpret data. Unlike traditional linear/nonlinear regression methods which fit parameters to equations of a given form, SR allows free combination of mathematical operators to obtain an open-ended solution and thus makes it possible to automatically infer an analytical model from data (e.g., by simultaneously discovering both the form of equations and controlling parameters). Monte Carlo sampling \cite{guimera2020bayesian} and evolutionary algorithms, such as genetic programming (GP), have been widely applied to distill mathematical expressions and governing laws that best fit available measurement data for nonlinear dynamical systems \cite{forrest1993genetic, schmidt2009distilling, bongard2007automated, ly2012learning, quade2016prediction, vaddireddy2020feature, ma2022evolving_SA_google_DFT}. However, this type of approach is known to scale poorly to problem's dimensionality, exhibits sensitivity to hyperparameters, and generally suffers from extensive computational cost in an extensively large search space.

Another remarkable breakthrough leverages sparse regression, in a restricted search space based on a pre-defined library of candidate functions, to select an optimal analytical model \cite{brunton2016discovering_SINDy}. Such an approach quickly became one of the state-of-art methods and kindled significant enthusiasm in data-driven discovery of ordinary or partial differential equations \cite{rudy2017data_PDE-FIND, long2018pde_Pde-net, champion2019data, wang2019variational, chen2021physics_PINN-SR, golden2023physically, rao2023encoding_PeRCNN, gao2022autonomous_NCS} as well as state estimation \cite{course2023state}.  However, the success of this sparsity-promoting approach relies on a properly defined candidate function library that operates on a fit-complexity Pareto front. It is further restricted by the fact that the linear combination of candidate functions is usually insufficient to express complicated mathematical formulas. Moreover, when the library size is overly massive, this approach generally fails to hold the sparsity constraint.

Deep learning has also been employed to uncover generic symbolic formulas from data, e.g., recurrent neural networks with risk-seeking policy gradient formulation \cite{petersen2021deep}, variational grammar autoencoders \cite{kusner2017grammarvae}, neural-network-based graph modularity \cite{udrescu2020ai_AIFeynman, udrescu2020ai_AIFeynman2.0}, pre-trained transformers \cite{2021BiggioNeural_NeSymReS, kamienny2022end_E2E, holt2023deep_DGSR}, etc. Another notable work has leveraged symbolic neural networks to distill physical laws of dynamical systems \cite{sahoo2018learning_EQLdiv, long2019pde_Pde-net2.0}, where commonly seen mathematical operators are employed as symbolic activation functions to establish intricate formulas via weight pruning. Nevertheless, this framework is primarily built on empirical pruning of the weights, thus exhibits sensitivity to user-defined thresholds and may fall short of producing parsimonious equations for noisy and scarce data.

Very recently, Monte Carlo tree search (MCTS) \cite{coulom2006efficient, kocsis2006bandit}, which gained acclaim for powering the decision-making algorithms in AlphaGo \cite{silver2017mastering_alphaGo}, AlphaZero \cite{silver2018general_AlphaZero} and AlphaTensor \cite{fawzi2022discovering_AlphaTensor}, has shown great potential in navigating the expansive search space inherent in SR \cite{sun2023symbolic_SPL}. This method uses stochastic simulations to meticulously evaluate the merit of each node in the search tree and has empowered several SR techniques \cite{Pierre2023DGSRMCTS_ICML, xu2024reinforcement_RSRM, Shojaee_reviewer_NIPS2023_TPSR, li2024discoveringmathematicalformulasdata}. However, the conventional application of MCTS maps each node to a unique expression, which tends to impede the rapid recovery of accurate symbolic representations. This narrow mapping may limit the strategy's ability to efficiently parse through the complex space of potential expressions, thus presenting a bottleneck in the quest for swiftly uncovering underlying mathematical equations. 

SR involves evaluating a large number of complex symbolic expressions composed of various operators, variables, and constants. The operators and variables are discrete, while the value of the coefficients is continuous. The NP-hardness of a typical SR process, noted by many scholars \cite{udrescu2020ai_AIFeynman, mundhenk2021symbolic_NGGP, petersen2021deep}, has been formally established \cite{virgolin2022symbolic_NPhard}. This characteristic requires the algorithm to traverse various possible combinations, resulting in a huge and even infinite search space and thus facing the problem of combinatorial explosion. Unfortunately, all the existing SR methods evaluate each candidate expression independently, leading to significantly low computational efficiency. Although some works considered caching evaluated expressions to prevent recomputation \cite{Shojaee_reviewer_NIPS2023_TPSR}, they do not reuse these results as subtree values for evaluating deeper expressions. Consequently, the majority of these methods either rely on meta-heuristic search strategies, narrow down the search space based on specific assumptions, or incorporate pre-trained models to discover relatively complex expressions, which make the algorithms prone to producing specious results (e.g., local optima). Therefore, the efficiency of candidate expression evaluation is of paramount importance. By enhancing the efficiency of candidate expression evaluation, it becomes possible to design new SR algorithms that are less reliant on particular optimization methods, while increasing the likelihood of directly finding the global optima in the grand search space. Thus, we can improve the SR accuracy (in particular, the symbolic recovery rate) and, meanwhile, drastically reduce the computational time.

To this end, we propose novel parallel symbolic enumeration (PSE) to automatically distill symbolic expressions from limited data. Such a model is capable of efficiently evaluating potential expressions, thereby facilitating the exploration of hundreds of millions of candidate expressions simultaneously in parallel within a mere few seconds. In particular, we propose a parallel symbolic regression network (PSRN) as the cornerstone search engine, which (1) automatically captures common subtrees of different math expression trees for shared evaluation that avoids redundant computations, and (2) capitalizes on graphics processing unit (GPU) based parallel search that results in a notable performance boost. It is notable that, in a standard SR process for equation discovery, many candidate expressions share common subtrees, which leads to repeatedly redundant evaluations and, consequently, superfluous computation. To address this issue, we propose a strategy to automatically reuse common subtrees for shared evaluation, effectively circumventing significantly redundant computation. We further execute PSE on a GPU to perform a large-scale evaluation of candidate expressions in parallel, thereby augmenting the efficiency of the evaluation process. Notably, the synergy of these two techniques could yield up to four orders of magnitude efficiency improvement in the context of expression evaluation. In addition, to expedite the convergence and enhance the capacity of PSRN for exploration and identification of more intricate expressions, we amalgamate it with a token generator strategy (e.g., MCTS or GP) which identifies a set of admissible base expressions as tokenized input. The effectiveness and efficiency of the proposed PSE model have been demonstrated on a variety of benchmark and lab test datasets. The results show that PSE surpasses multiple existing baseline methods, achieving higher symbolic recovery rates and efficiency.

\section*{Methods}
\label{sec:methods}

\begin{figure}[htbp]
  \hspace*{-0.0in}
  \centering
  \includegraphics[width=0.9\textwidth]{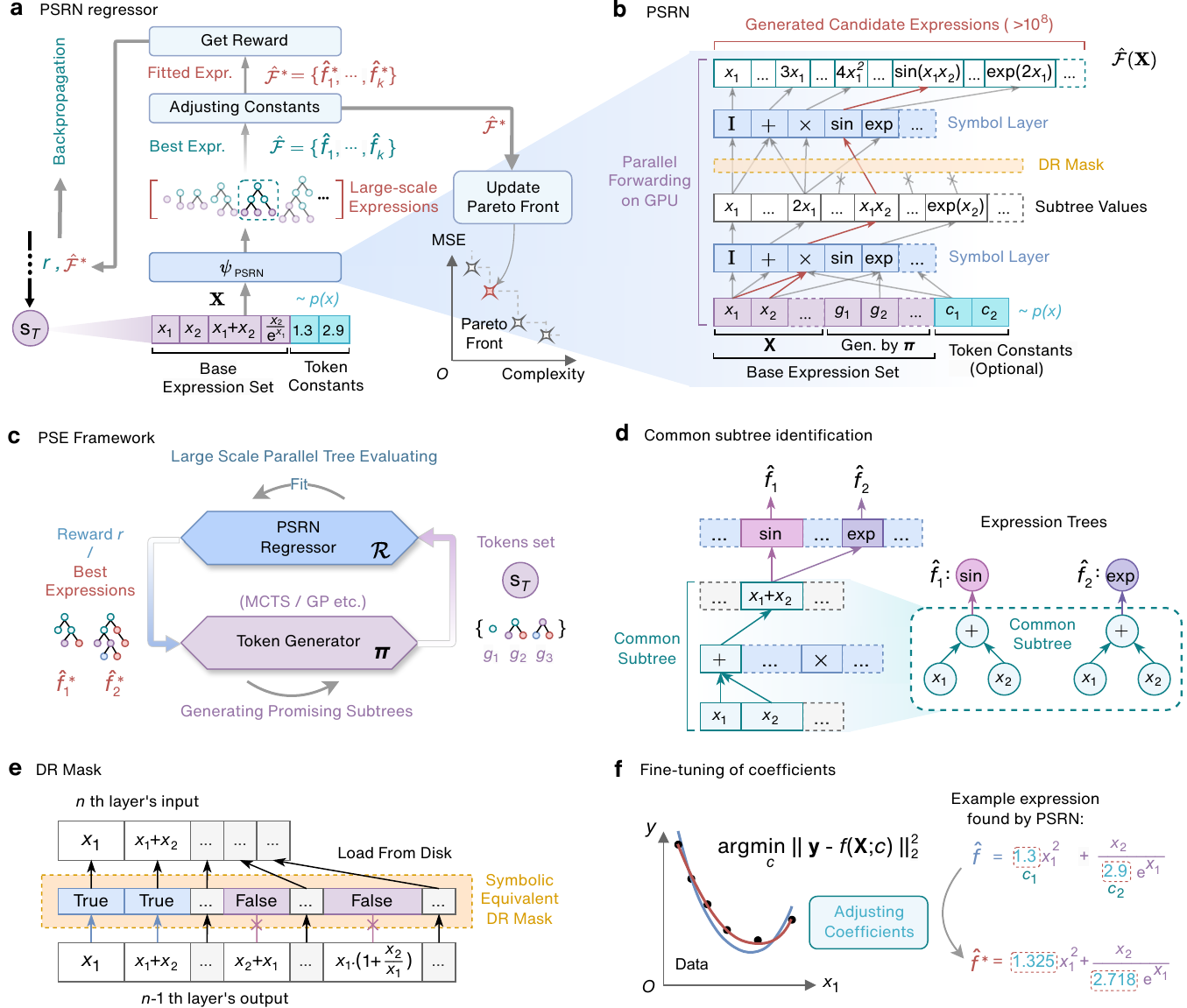}
  \caption{\textbf{Overview of the proposed PSE model}. \textbf{a}, Schematic of PSRN regressor for discovering optimal expressions from data. The base expression set $s_{\mathcal{T}}$ from terminal nodes, along with optionally sampled token constant values, is fed into PSRN (denoted as $\psi_\text{PSRN}$) for forward computation. After evaluating the errors of large-scale candidate expressions, $\psi_\text{PSRN}$ provides the optimal (or top-$k$) expressions denoted as $\hat{\mathcal{F}}$. Subsequently, the least squares method is employed for the identification and fine-tuning of the coefficients, resulting in adjusted expressions $\hat{\mathcal{F}}^*$, which are then utilized for updating the Pareto front as computing the complexity and reward. \textbf{b}, Forward computation in PSRN. The base expression set and sampled constants, along with the corresponding data tensor $\mathbf{X}$, are fed into the network. The network comprises multiple Symbol Layers (e.g., typically three, but only two are shown for simplicity). Each layer provides all possible subtree values resulting from one-operator computations among all input expressions. This process can be efficiently parallelized on a GPU for rapid computation. Upon completion of PSRN's forward computation, a myriad of candidate expressions $\hat{\mathcal{F}}$  (e.g., up to hundreds of millions) corresponding to the base expression set $s_{\mathcal{T}}$, along with their associated error tensors on the given data, are generated. Then, the expression with the minimal error (or top-$k$ expressions) will be selected. \textbf{c}, The token generator $\pi$ (e.g., MCTS, GP) creates promising subtrees from a token set $s_T$. These subtrees are evaluated by the PSRN Regressor $\mathcal{R}$ through large-scale parallel expression evaluation. $\mathcal{R}$ fits the input data, producing a reward $r$ and the best expressions $\hat{\mathcal{F}}^*$, which are fed back to $\pi$. This feedback loop drives continuous improvement in generating promising subtrees. The process combines $\pi$'s ability to explore the token space with $\mathcal{R}$'s capacity to assess expression quality based on data fit. This iterative system efficiently discovers high-quality expressions by leveraging both heuristic search and large-scale parallel evaluation. The overall iterative PSE process is detailed in \textcolor{blue}{Supplementary Algorithm 2}, with specific token generators $\pi$ described in \textcolor{blue}{Supplementary Algorithms 3--5}. \textbf{d}, Schematic of common subtree identification. Expressions sharing common subtrees, e.g., $\sin(x_1+x_2)$ and $\exp(x_1+x_2)$, leverage identical subtree computation outcomes. This approach circumvents extensive redundant calculations, thereby increasing the efficiency of SR. \textbf{e}, Schematic of duplicate removal mask (DR Mask). We designed the DR Mask layer to mask the output tensor of the penultimate layer in PSRN, thereby excluding subtree expressions that are symbolically equivalent. This significantly reduces PSRN's usage of GPU memory. \textbf{f}, Estimation and fine-tuning of constant coefficients. The least squares method is used to fine-tune the coefficients of preliminary expressions discovered by PSRN using the complete dataset.}
  \label{fig_overall}
\end{figure}

The NP-hardness of SR, noted in existing studies \cite{udrescu2020ai_AIFeynman,petersen2021deep,mundhenk2021symbolic_NGGP}, was formally proven recently \cite{virgolin2022symbolic_NPhard}. This implies the absence of a known solution that can be determined in polynomial time for solving SR problems across all instances, and reflects that the search space for SR is vast and intricate. Almost all the existing SR methods involve a common procedure, which is to assess the quality of candidate expressions $\hat{\mathcal{F}}=\{\hat{f}_1,\hat{f}_2,...\}$ based on the given data $\mathcal{D} = (\mathbf{X}, \mathbf{y})$, where $\mathbf{X}\in \mathbb{R}^{n\times m}$, $\mathbf{y}\in \mathbb{R}^{n\times 1}$. Here, $\hat{f}$ is usually constructed by a given operator set $\mathcal{O}$ (e.g., $\{+,\times,-,\div, \sin, \cos, \exp, \log\}$). Typically, this evaluation is guided by the mean square error (MSE) expressed as:
\begin{equation}\label{mse}
    \text{MSE}=\frac{1}{n}||\hat{f}(\mathbf{X})-\mathbf{y}||^2_2.
\end{equation}
During the search process, a large number of candidate expressions need to be evaluated, while the available operators and variables are limited, leading to the inevitable existence of vast repeated sub-expressions. Existing methods evaluate candidate expressions sequentially and independently. As a result, the common intermediate expressions are repeatedly calculated, leading to significant computational burden (see \textcolor{blue}{Supplementary Fig. S.1}). By reducing the amount of repeated computation, the search process can be significantly accelerated.

To address these challenges and enhance both the efficiency and accuracy of SR, we propose a novel parallel symbolic enumeration (PSE) model, as shown in Fig. \ref{fig_overall}, to automatically discover mathematical expressions from limited data. The core of PSE is a Parallel Symbolic Regression Network (PSRN) regressor, depicted in Fig. \ref{fig_overall}\textbf{b}--\textbf{c}, designed to significantly enhance the efficiency of candidate expression evaluation. The architecture of PSRN is inherently parallelism-friendly, enabling rapid GPU-based parallel computation (see Fig. \ref{fig_overall}\textbf{c}). A key innovation of PSRN is its ability to automatically identify and reuse the intermediate calculations of common subtrees (see Fig. \ref{fig_overall}\textbf{d}), thus avoiding redundant computations.

To further bolster the model's discovery capability for more complex expressions, we integrate a token generator (e.g., MCTS or GP) which identifies a set of admissible base expressions as tokenized input to PSRN (see Fig. \ref{fig_overall}\textbf{a}). The PSE model operates iteratively: PSRN continually activates the token generator, starting with a base expression set that includes available independent variables. During these iterations, this set of base expressions is updated by progressively incorporating more complex expressions generated by the token generator. These base expressions, along with sampled token constants, are fed into PSRN for evaluation and the search of optimal symbolic expressions (see Fig. \ref{fig_overall}\textbf{b}). PSRN can rapidly (e.g., within seconds) evaluate hundreds of millions of candidate symbolic expressions, identifying the one with the smallest error or a few best candidates. This represents a significant speed improvement compared to approaches that evaluate each candidate expression independently. Essentially, PSRN performs a large-scale and parallel symbolic enumeration of expression trees with scalability. Additionally, a duplicate removal mask (DR Mask) is designed for PSRN to reduce memory usage (see Fig. \ref{fig_overall}\textbf{e}). Within the PSRN regressor, the coefficients of the most promising expressions are identified and fine-tuned using least squares estimation (see Fig. \ref{fig_overall}\textbf{f}). Rewards computed on the given data are then back-propagated for subsequent token generator updates. As the search progresses, the Pareto front representing the optimal set of expressions is continuously updated to report the final result.

\subsection*{Symbol layer}
\label{subsec:symbol_layer}

A mathematical expression can be equivalently represented as an expression tree \cite{forrest1993genetic}, where the internal nodes denote mathematical operators and the leaf nodes the variables and constants. The crux of the aforementioned computational issue lies in the absence of temporary storage for subtree values and parallel evaluation. Consequently, the common subtrees in different candidate expressions are repeatedly evaluated, resulting in significant computational wastage.

To this end, we introduce the concept of Symbol Layer (see Fig. \ref{fig_overall}\textbf{c}), which consists of a series of mathematical operators. The Symbol Layer serves to transform the computation results of shallow expression trees into deeper ones. From the perspective of avoiding redundant computations, the results of the Symbol Layer are reused by expression trees with greater heights. From the view of parallel computation, the Symbol Layer can leverage the power of GPU to compute the results of common subtrees in parallel, significantly improving the overall speed (see \textcolor{blue}{Supplementary Fig. S.1}). We categorize the mathematical operators in the Symbol Layer into three types: (1) unary operators, (e.g., sin and exp); (2) binary-squared operators (e.g., $-$ and $\div$), representing non-commutative operators; (3) binary-triangled operators, representing commutative operators (e.g., $+$ and $\times$) or a variant of non-commutative operators with low-memory footprint (e.g., SemiSub and SemiDiv that output $x_i - x_j$ and $x_i \div x_j$ for $i \le j$, respectively). Note that the binary-triangled operators only take up half of the space compared with the binary-squared operators. We denote these three types of operators as ${u}$, ${b}_{S}$, and ${b}_{T}$, respectively. Mathematically, a Symbol Layer located at the $l$-th level can be represented as follows: 

\begin{equation}
    \mathbf{h}^{(l)}=\left(
\concat_{i=1}^{N_u}{\mathbf{u}_i\left(\mathbf{h}^{(l-1)}\right)}\right)\concat\left(
\concat_{i=1}^{N_{b_S}}{{\mathbf{b}_{S}}_i\left(\mathbf{h}^{(l-1)}\right)}\right)\concat\left(
\concat_{i=1}^{N_{b_T}}{{\mathbf{b}_{T}}_i\left(\mathbf{h}^{(l-1)}\right)}
\right),
\end{equation}
where
\begin{equation}
    \mathbf{u}(\mathbf{h})=\concat_{i}{u(h_{i})},~~\mathbf{b}_{S}(\mathbf{h})=\concat_{i,j}{{b_S}(h_{i},h_{j})},~~\mathbf{b}_{T}(\mathbf{h})=\concat_{i\le j}{{b_T}(h_{i},h_{j})}.
\end{equation}
Here, $\concat$ represents the concatenation operation; $N_u$, $N_{b_S}$ and $N_{b_T}$ denote the numbers of unary operators, binary-squared operators and binary-triangled operators.

For example, let us consider independent variables $\{x_1, x_2\}$ and dependent variable $z$, with the task of finding an expression that satisfies $z=f(x_1,x_2)$. When the independent variables $\{x_1,x_2\}$ are fed into a Symbol Layer, we obtain the combinatorial results (e.g., $\{x_1, x_2, \sin{(x_1)}, \sin{(x_2)}, \ldots, x_1+x_1, x_1+x_2, x_2+x_2, \ldots\}$). Notably, each value in the output tensor of the Symbol Layer corresponds to a distinct sub-expression. This enables PSE to compute the common subtree values just once, avoiding redundant calculations. Additionally, the Symbol Layer can be leveraged on the GPU for parallel computation, further enhancing the speed of expression searches significantly.

\begin{figure}[htbp]
  \hspace*{-0.0in}
  \centering
  \includegraphics[width=0.99\textwidth]{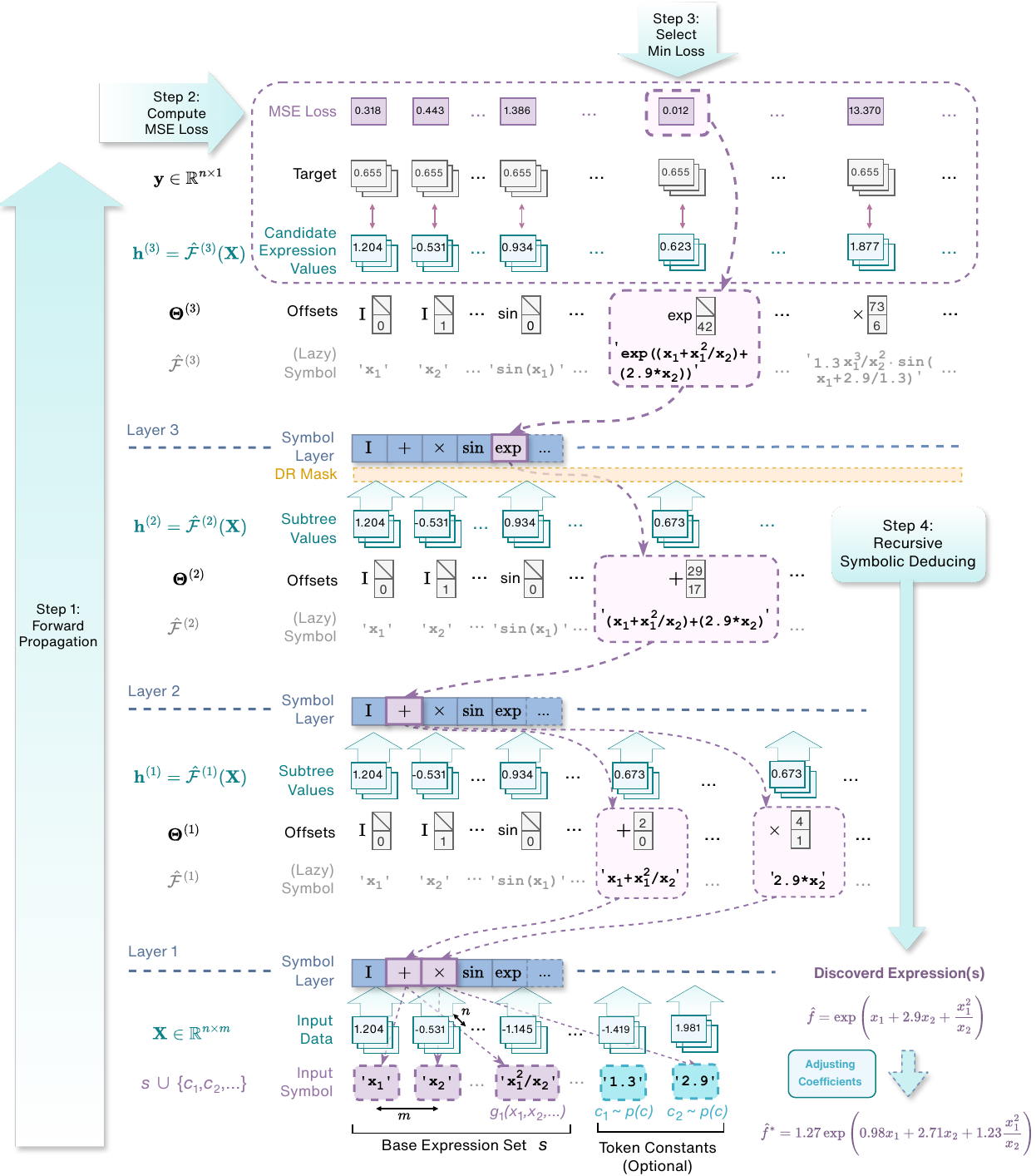}
  \caption{\textbf{The detailed process of PSRN forward propagation for obtaining the optimal expression.} Step 1: The input data $\mathbf{X}$ and the base expression set $s$, together with optional token constants sampled from a pre-defined distribution $p(c)$, are fed into PSRN for forward propagation. Based on the designated operator categories, a large number of distinct subtree values $\mathbf{h}$, e.g., $\hat{\mathcal{F}}(\mathbf{X})$ are rapidly calculated layer by layer. Step 2: The MSE is computed between the final layer's output  and the broadcast target tensor $\mathbf{y}$, resulting in the loss tensor. Step 3: The position of the minimum value in the loss tensor is selected. Step 4: Starting from the optimal position, a recursive symbolic deducing is performed using the offset tensor $\mathbf{\Theta}$ generated when building the network, ultimately yielding the optimal expression. If necessary, the coefficients of this expression will be adjusted in post-processing. The algorithmic procedure for PSRN evaluation and symbolic reconstruction is presented in \textcolor{blue}{Supplementary Algorithm 1}.}
  \label{fig_detailed_symbol_layer}
\end{figure}

When establishing a Symbol Layer initially, an inherent offset tensor $\mathbf{\Theta}$ is inferred and stored within the layer for subsequent lazy symbolic deducing during the backward pass (see Fig. \ref{fig_detailed_symbol_layer}). For the $l$-th Symbol Layer, its offset tensor can be represented as follows:
\begin{equation}
    \mathbf{\Theta}^{(l)}=\left( \concat_{i=1}^{N_u}\mathbf{\Theta}_u\right)\concat\left(\concat_{i=1}^{N_{b_{S}}}\mathbf{\Theta}_{b_{S}}\right)\concat\left(\concat_{i=1}^{N_{b_{T}}}\mathbf{\Theta}_{b_{T}}  \right),
\end{equation}
where

\vspace{-12pt}
\begin{subequations}
\begin{align}
    \mathbf{\Theta}_{u}&=\begin{bmatrix}1\ \cdots\ \omega\\\emptyset\ \cdots\ \emptyset\end{bmatrix}_{2\times \omega}, \\
    \mathbf{\Theta}_{b_{S}}&=\concat_{j=1}^{\omega}\begin{bmatrix}1\  &\cdots &\ \omega\\j\ &\cdots &\ j\end{bmatrix}_{2\times \omega^2}, \\
    \mathbf{\Theta}_{b_{T}}&=\concat_{j=1}^{\omega}\begin{bmatrix}1\  &\cdots &\ \omega-j+1\\j\ &\cdots &\ j\end{bmatrix}_{2\times \frac{\omega(\omega+1)}{2}}.
\end{align}
\end{subequations}
Here, $\omega$ denotes the output dimension (the number of symbolic expression trees) of the previous layer.

Each position in the output tensor $\mathbf{h}^{(l)}$ of the Symbol Layer corresponds to a unique column in the offset tensor $\mathbf{\Theta}^{(l)}$. In other words, it corresponds to two child indices (for unary operators, only one), representing the left and right child nodes of the current output tensor position from the previous layer. These offset tensors are used for recursive backward symbol deduction after the position of a minimum MSE value is found.

\subsection*{Parallel symbolic regression network}
\label{subsec:PSRN}

By stacking multiple Symbol Layers, we construct a Parallel Symbolic Regression Network (PSRN), denoted as $\psi_{\mathrm{PSRN}}$ (Fig. \ref{fig_overall}\textbf{c}). PSRN takes a set of $N_{in}$ input base expressions $s = \{s_1, s_2, ..., s_{N_{in}}\}$ (which can be raw variables, constants, or more complex tokens generated as described in Subsection \nameref{subsec:PSRN_mcts}) and their corresponding data tensor $\mathbf{X}$. It then performs a rapid, parallel forward computation on the GPU. Based on the predefined operator set $\mathcal{O}$, PSRN efficiently evaluates a vast number of distinct candidate expressions by systematically combining its inputs through its layers. For instance, with $l$ Symbol Layers, PSRN can generate and evaluate all expressions representable as expression trees of depth up to $l$ using the provided base expressions $s$ as leaf nodes. A key advantage is that PSRN computes the values of common subtrees only once, avoiding redundant calculations and significantly speeding up the evaluation of potentially hundreds of millions of expressions.

Once the final layer's computations are complete, PSRN outputs the values $\hat{\mathcal{F}}(\mathbf{X})=\{\hat{f_1}(\mathbf{X}), \hat{f_2}(\mathbf{X}), \ldots\}$ for all generated candidate expressions. These are then used to compute the Mean Squared Error (MSE, Eq. (\ref{mse})) against the target data $\mathbf{y}$, allowing the identification of the expression(s) with the minimum error. The pre-generated offset tensors $\mathbf{\Theta}$ within each Symbol Layer are then used to recursively reconstruct the symbolic form of the optimal candidate expression(s) (see Fig. \ref{fig_detailed_symbol_layer}). The detailed steps for PSRN's evaluation and symbolic derivation are provided in \textcolor{blue}{Supplementary Algorithm 1}.

The choice of operator set and the number of input slots for PSRN can be adapted based on the problem complexity and available GPU memory. In the SR benchmark tasks, we employ a 5-input PSRN with the operator set $\mathcal{O}_\mathrm{Koza}=\{+,\times,-,\div,\mathrm{identity},\sin,\cos,\exp,\log\}$, except for the Feynman expression set which uses a 6-input PSRN with the operator set $\mathcal{O}_\mathrm{SemiKoza}=\{+,\times,\mathrm{SemiSub},\mathrm{SemiDiv},\mathrm{identity},\mathrm{neg},\mathrm{inv},\sin,\cos,\exp,\log\}$. Such a setting aims to conserve GPU memory for handling more input variables. The distinguishing feature of the $\mathcal{O}_\mathrm{SemiKoza}$ operator set is that it treats division and subtraction as binary-triangled operators and allows only one direction of operation (see \nameref{sec:methods}: \nameref{subsec:symbol_layer}). This trade-off reduces expressive power but conserves GPU memory, which enables to tackle larger scale SR tasks. To discover more deeply nested or complex expressions, PSRN is integrated with a token generator, as described next.

\subsection*{PSRN regressor with token generator}
\label{subsec:PSRN_mcts}

While PSRN can evaluate expressions up to a depth of $l$ (number of Symbol Layers) from its direct inputs in a single pass, discovering more complex real-world equations often requires exploring deeper expression trees. To extend PSRN's reach and enable the discovery of such intricate symbolic expressions, we integrate the PSRN into a larger iterative framework driven by a token generator, $\pi$. This system operates through a synergistic loop. In each iteration, the token generator first proposes a new set of promising base expressions (tokens), denoted as $s_T$. Subsequently, the PSRN regressor $\mathcal{R}$ takes $s_T$ as input and performs a rapid, large-scale parallel enumeration to evaluate millions of candidate expressions, identifying the top-$k$ expressions, $\{\hat{f}_1, ..., \hat{f}_k\}$, that best fit the data. A reward signal is then computed based on the performance (e.g., accuracy and complexity) of these discovered expressions. Finally, this reward is fed back to the token generator $\pi$, guiding its strategy for generating even better tokens in the subsequent iteration. This iterative process allows PSE to progressively build and refine complex equations that would be unreachable by a single pass of PSRN alone. The overall PSE procedure is detailed in \textcolor{blue}{Supplementary Algorithm 2}.

\paragraph{Token Generation Strategies.} The token generator, $\pi$, is a modular component responsible for exploring the vast space of possible sub-expressions. It is initialized with a base set of operators $\mathcal{O}$ and variables $\mathcal{V}$. In each iteration, it produces a token set, $s_T$, which can include raw variables, numerical constants, and composite expressions (e.g., $\{x_1, x_2, x_1^2+x_2, \frac{x_1}{\exp(x_2)},1.3, 2.9\}$) built from previously successful tokens. In this work, we explore several strategies for $\pi$. One approach is Genetic Programming (GP), which maintains and evolves a population of expression trees using operations like crossover and mutation to generate new tokens. The GP-based token generation approach is outlined in \textcolor{blue}{Supplementary Algorithm 3}. Another method is Monte Carlo Tree Search (MCTS), which builds a search tree over the expression space and uses simulation-based feedback to explore promising branches. Our MCTS token generator is detailed in \textcolor{blue}{Supplementary Algorithm 4}. We also use a random generation strategy as a baseline, which simply combines operators and existing tokens at random. The random token generation strategy is presented in \textcolor{blue}{Supplementary Algorithm 5}. A detailed comparison of these strategies is presented in our ablation study (see \textcolor{blue}{Results:} \nameref{subsec:ablation}). As the GP-based generator was found to be the most effective, it is used as the default configuration in our primary experiments.

\paragraph{Reward Calculation and Pareto Front.} To guide the search, each candidate expression $\hat{f}$ is assigned a reward $r$ that balances its accuracy and complexity:
\begin{equation}
\label{eq:reward} 
r=\frac{{\eta}^{\alpha}}{1+{\sqrt{\mathrm{MSE}}}},
\end{equation}
where $\eta$ is a discount factor (default 0.99) penalizing complexity, and $\alpha$ is the expression's complexity. The set of non-dominated solutions—expressions for which no other expression is better in both accuracy and simplicity—is maintained on a Pareto front. This front is updated in each iteration and represents the final output of the algorithm. The maximum reward achieved is used as the feedback signal for the token generator $\pi$.

\subsection*{Coefficients tuning}
\label{subsec:deal_const}

Many physical laws and mathematical expressions involve numerical coefficients. PSE incorporates a two-stage approach to handle these constants effectively.

\paragraph{Initial Constant Representation in PSRN.} PSRN can incorporate numerical constants during its expression generation phase. As shown in Fig. \ref{fig_overall}\textbf{a}--\textbf{b} and Fig. \ref{fig_detailed_symbol_layer}, a predefined number of PSRN input slots can be reserved for token constants. In each iteration, before PSRN's forward propagation, constant values $c$ are sampled from a pre-selected distribution $p$ (e.g., $c\sim U(-1,1)$, $c\sim U(-3,3)$) and fed into these constant slots. PSRN can then use these explicit numerical values directly or combine them with operators (e.g., $1.3+2.9$, $\exp(1.3)$, $1.3/\sin(2.9)$) to form more diverse embedded constants within the candidate expressions $\hat{\mathcal{F}}$ it generates. This allows PSRN to explore structures that inherently include numerical values.

\paragraph{Post-PSRN \kredit{Coefficients} Refinement.} After PSRN identifies a set of promising candidate symbolic structures $\hat{\mathcal{F}}=\{\hat{f_1}, \ldots, \hat{f_k}\}$ (which may contain the initially sampled or derived constants), a more precise coefficient optimization is performed (see Step 4 in Fig. \ref{fig_detailed_symbol_layer}). For each selected raw expression $\hat{f}_i$, we parse it to identify all numerical coefficient positions. These coefficients are then treated as free parameters and are collectively optimized using a standard Least Squares (LS) method against the entire training dataset $(\mathbf{X}, \mathbf{y})$:
\begin{equation}
    \hat{f}^{*}_i=\mathrm{LS}\big(\hat{f}_i,\mathbf{X},\mathbf{y}\big),\ i=1,\ldots,k.
\end{equation}
This LS step fine-tunes the initial constant values found or incorporated by PSRN, leading to the final optimized expressions $\hat{\mathcal{F}}^* = \big\{\hat{f}^*_1, ..., \hat{f}^*_k \big\}$. This ensures that the numerical values in the discovered equations are optimally fitted to the data, given the symbolic structure. \kredit{While we employ the LS method, this refinement step can be generalized to the maximum likelihood estimation (MLE) for more complex noise models or maximum \textit{a posteriori} optimization to incorporate regularization.}
During the warm-up phase of PSE, a simple linear regression step using the independent variables is also performed to expedite the discovery of simpler expressions that might involve basic coefficient fitting.

\subsection*{Duplicate removal mask}
\label{subsec:drmask}

One major limitation of standard PSRN lies in its high demand for graphics memory. When using binary operators such as addition and multiplication, the graphics memory required for each layer grows quadratically with respect to the tensor dimensions of the previous layer. Therefore, reducing the output dimension of the penultimate layer can significantly reduce the required graphics memory. We propose a technique called Duplicate Removal Mask (DR Mask) as shown in Fig. \ref{fig_overall}\textbf{e}. Before the last PSRN layer, we employ the DR Mask to remove repeated terms and generate a dense input set. Specifically, assuming the input variables $x_1, x_2, ..., x_m$ are independent, we extract the output expressions from the last second layer, parse them using the symbolic computing library SymPy, and remove duplicate expressions to obtain a mask of unique expressions. SymPy's efficient hash value computation facilitates the comparison of mathematical expressions for symbolic equivalence. Typically, this process takes less than a minute. Once the DR Mask is identified, it is reusable for PSRNs with the same architecture (e.g., total number of network layers and operator sets). During the search process, the expressions marked by the mask are extracted from the penultimate layer's output tensor and then passed to the final layer for the most graphics-intensive binary operator computations. The percentage of graphics memory saved by the DR Mask technique depends on the input size, the number of layers, and the operators used in the PSRN architecture. Detailed results are shown in \textcolor{blue}{Results:} \nameref{subsec:ablation}.

\subsection*{Baseline models}

We consider nine main baseline SR algorithms used in the comparasion analysis: DGSR \cite{holt2023deep_DGSR}, NGGP \cite{mundhenk2021symbolic_NGGP}, PySR \cite{cranmer2023interpretable_PySR}, BMS \cite{guimera2020bayesian}, SPL \cite{sun2023symbolic_SPL}, wAIC \cite{gao2022autonomous_NCS}, uDSR \cite{uDSR_landajuela2022unified}, TPSR \cite{Shojaee_reviewer_NIPS2023_TPSR} and Operon \cite{Operon} Additionally, we compare with ESR \cite{ESR_Bartlett_2022} on single-variable problems only, since it does not support multivariate regression. They represent a selection of SR algorithms, ranging from recent advancements to established, powerful methods. The introduction and detailed settings of the baseline models are found in \textcolor{blue}{Supplementary Note 2.3}.

\section*{Results}
\label{sec:results}

To demonstrate the effectiveness and efficiency of the proposed PSE model, we conduct a comprehensive evaluation across a diverse range of challenging tasks. These include standard symbolic regression benchmarks, the data-driven discovery of governing equations for chaotic dynamical systems, and the modeling of real-world physical phenomena from experimental data. It is noteworthy that, unless specified otherwise in the ablation studies, all experiments presented hereafter utilize the GP variant of our token generator to effectively explore the vast search space. For real-world datasets like the electro-mechanical positioning system and the turbulent friction experiments, we assess the plausibility of discovered equations using MSE and parsimony due to the absence of ground truth expressions. For other benchmark experiments, we employ the symbolic recovery rate as the key measure of accuracy. Although the generated expression, further simplified by SymPy, is not guaranteed to be the best-fitting solution matching the ground truth especially for noisy datasets, this metric remains widely adopted in the SR community. It enables direct and standardized comparisons with a substantial body of prior work on established benchmark problem sets that contain ground-truth expressions. The collective results presented as follows consistently highlight the superiority of PSE in both symbolic recovery accuracy and computational performance over exisiting baseline methods.

\subsection*{Symbolic regression benchmarks}
\label{subsec:symbolic_regression}

We firstly demonstrate the efficacy of PSE on recovering specified mathematical formulas given multiple benchmark problem sets (including Nguyen \cite{uy2011semantically_Nguyen}, Nguyen-c \cite{mcdermott2012genetic}, R \cite{krawiec2013approximating_Rbenchmark}, Livermore \cite{mundhenk2021symbolic_NGGP} and Feynman \cite{udrescu2020ai_AIFeynman}, as described in \textcolor{blue}{Supplementary Note 2.1}), commonly used to evaluate the performance of SR algorithms. Each SR puzzle consists of a ground truth equation, a set of available math operators, and a corresponding dataset. These benchmark data sets contains various math expressions, e.g., $x^3+x^2+x$ (Nguyen-1), $3.39x^3+2.12x^2+1.78x$ (Nguyen-1c), $(x+1)^3/(x^2-x+1)$ (R-1), $1/3+x+\sin{(x^2)}$ (Livermore-1), $x_1^4-x_1^3+x_1^2-x_2$ (Livermore-5), and $x_1x_2x_3\log{(x_5/x_4)}$ (Feynman-9), which are listed in detail in \textcolor{blue}{Supplementary Tables S.1}--\textcolor{blue}{S.2}. Our objective is to uncover the Pareto front of optimal mathematical expressions that balance the equation complexity and error. The performance of PSE is compared with eight baseline methods, e.g., 
symbolic physics learner (SPL) \cite{sun2023symbolic_SPL}, neural-guided genetic programming (NGGP) \cite{mundhenk2021symbolic_NGGP}, deep generative symbolic regression (DGSR) \cite{holt2023deep_DGSR}, PySR \cite{cranmer2023interpretable_PySR}, Bayesian machine scientist (BMS) \cite{guimera2020bayesian}, a unified framework for deep symbolic regression (uDSR) \cite{uDSR_landajuela2022unified}, Transformer-based planning for symbolic regression (TPSR) \cite{Shojaee_reviewer_NIPS2023_TPSR} and Operon \cite{Operon}. For the Nguyen-c dataset, each model is run for at least 20 independent trials with different random seeds, while for all other puzzles, 100 distinct random seeds are utilized. We mark the successful case if the ground truth equation lies in the discovered Pareto front set. 

\begin{figure}[t!]
  \hspace*{-0.0in}
  \centering
  \includegraphics[width=6.4in]{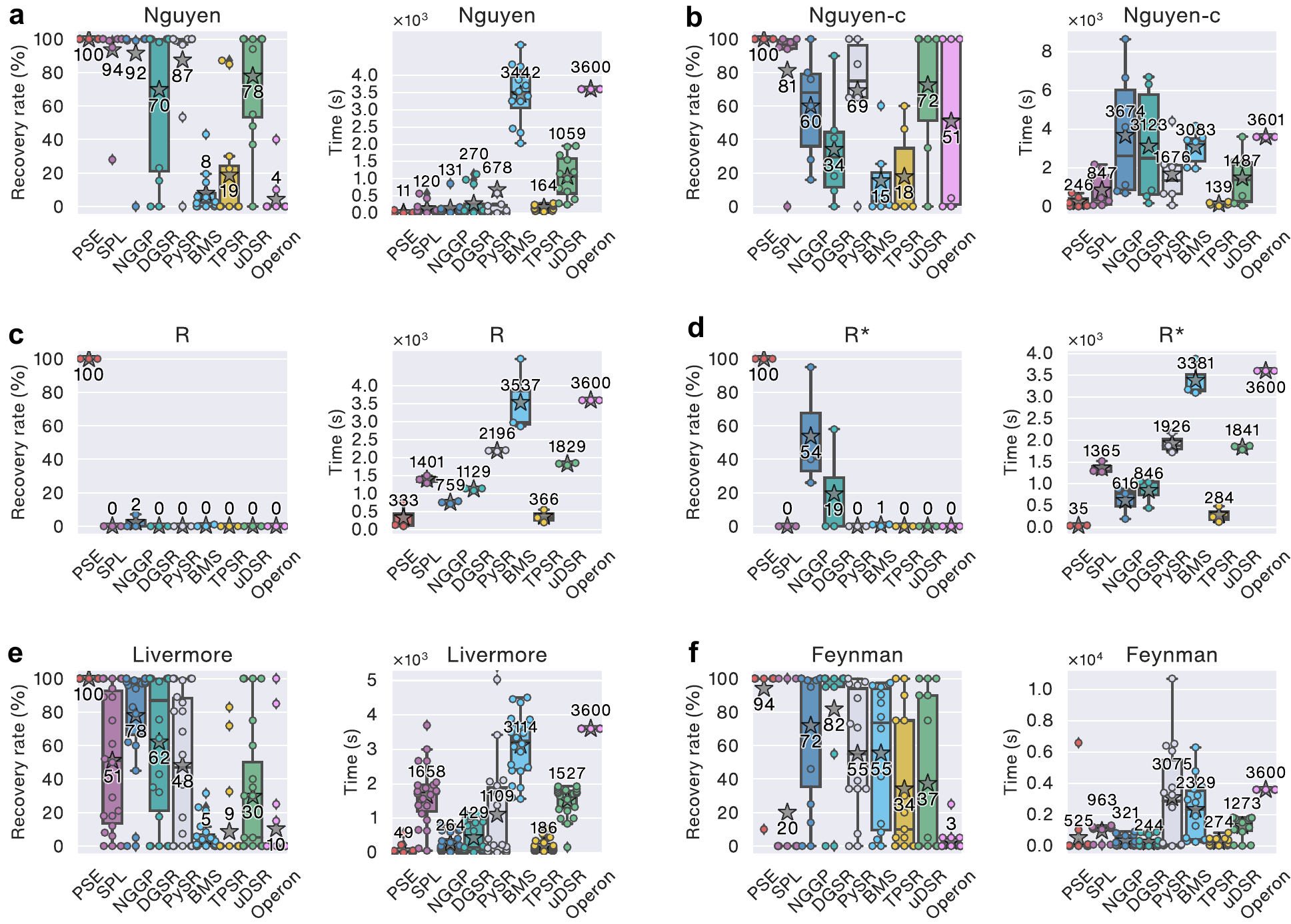}
  \caption{\textbf{Performance of various models on symbolic regression benchmarks.} We employ two evaluation metrics, namely, symbolic recovery rate and average runtime, to assess the performance of each algorithm. Our PSE approach achieves the highest recovery rate across various benchmark problem sets, meanwhile expending the minimal computation time, compared to the baseline methods (e.g., SPL \cite{sun2023symbolic_SPL}, NGGP \cite{mundhenk2021symbolic_NGGP}, DGSR \cite{holt2023deep_DGSR}, PySR \cite{cranmer2023interpretable_PySR}, BMS \cite{guimera2020bayesian}, uDSR \cite{uDSR_landajuela2022unified}, TPSR \cite{Shojaee_reviewer_NIPS2023_TPSR} and Operon \cite{Operon}). This highlights the substantial superiority of PSE. Note that the star marks the mean value over each problem set. Since BMS \cite{guimera2020bayesian} using priors from Wikipedia, it performs reasonably well on subsets of expressions like Feynman, but has very low recovery rates for other expression subsets as it's not suited for finding artificially designed expressions without constants. The Feynman dataset shown here follows the definition in DGSR's paper, which is a subset of SRBench's Feynman problems under data-limited conditions (less than 50 data points). For evaluation on the complete SRBench dataset, see \textcolor{blue}{Results:} \nameref{subsec:robust}. We strictly maintain consistent runtime budgets. The variations in the algorithms' runtime occur because different algorithms trigger their respective early stopping conditions (e.g., reaching an MSE below a very small threshold, or discovering symbolically equivalent expressions, etc.).}
  \label{fig_SRbenchmark_6plot}
\end{figure}

The SR results of different models, in terms of the symbolic recovery rate and computational time, are depicted in Fig. \ref{fig_SRbenchmark_6plot}. It can be seen that the proposed PSE method outperforms the baseline methods for all the benchmark problem sets in terms of recovery rate, meanwhile maintaining the high efficiency (e.g., expending the minimal computation time) that achieves up to two orders of magnitude speedup. The detailed results for each SR problem set are listed in \textcolor{blue}{Supplementary Tables S.\kredit{10}}--\textcolor{blue}{S.\kredit{15}}. An intriguing finding is that PSE achieves an impressive symbolic recovery rate of 100\% on the R benchmark expressions, while the baseline models almost fail (e.g., recovery rate < 2\%). We attribute this to the mathematical nature of the R benchmark expressions, which are all rational fractions like $(x+1)^3/(x^2-x+1)$ (R-1). Confined to the sampling interval $x\in[-1, 1]$, the properties of the R expressions bear a high resemblance to polynomial functions, resulting in intractable local minima that essentially lead to the failure of NGGP and DGSR. This issue can be alleviated given a larger interval, e.g., $x\in[-10, 10]$, as illustrated in the R$^*$ dataset (see \textcolor{blue}{Supplementary Table S.\kredit{12}}). Notably, the performance of DGSR based on pre-trained language models stems from the prevalence of polynomial expressions in the pre-training corpora, while NGGP collapses on account of its limited search capacity in an enormously large search space. In contrast, owing to the direct and parallel evaluation of multi-million expression structures, PSE possesses the capability of accurately and efficiently recovering complex expressions.

\subsection*{Discovery of chaotic dynamics}

\begin{figure}[t!]
  \hspace*{-0.0in}
  \centering
  \includegraphics[width=0.995\textwidth]{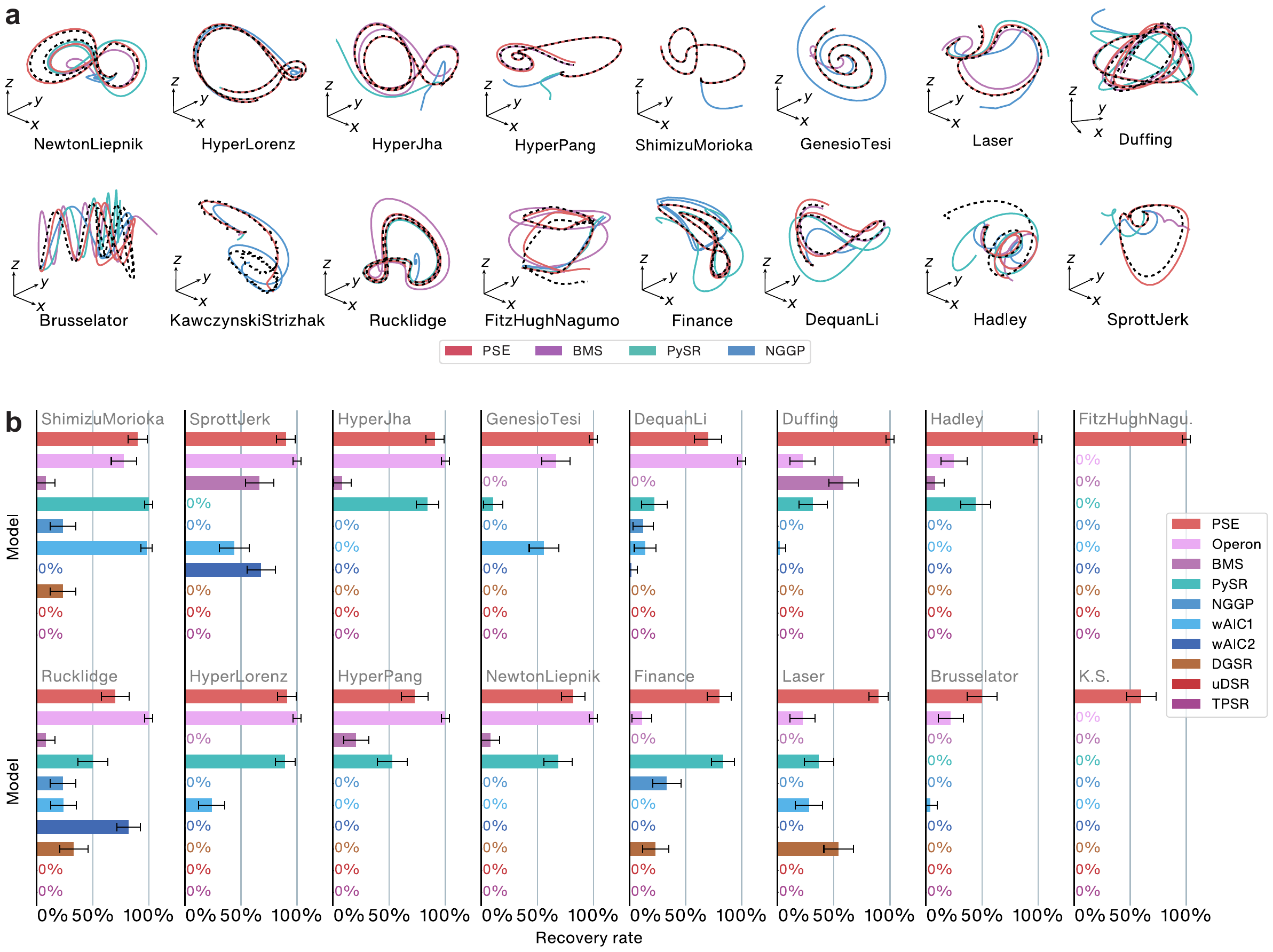}
  \caption{\textbf{Data-driven discovery of nonlinear chaotic dynamics by different models including PSE, BMS, PySR, NGGP, DGSR, wAIC, uDSR, TPSR, and Operon.} A large number of operators (e.g., $+, \times, -, \div, \sin, \cos, \exp, \cosh, \tanh, \mathrm{abs}, \mathrm{sign}$, etc.) are used to simulate the situation in which scientists explore unknown systems in the real world with less prior knowledge about operators. Given the same budget of computational time, PSE has a higher probability of finding the true governing equations among a nearly infinite number of possible expression structures. Each model's runtime is capped at around 10 minutes by limiting the number of iterations. \textbf{a}, Examples of predicted trajectories (e.g., solid lines) by different models compared with the ground truth (e.g., dashed lines). \textbf{b}, Average recovery rates of SR algorithms on 16 nonlinear chaotic dynamics datasets. Each bars with the same color represents the recovery rates under a typical case of 2\% Gaussian noise. More results under different noise levels (1\%, 5\%, and 10\% of the data's standard deviation) can be found in Extended Data Fig. \ref{fig_16_dysts_log_plot_bar} and \textcolor{blue}{Supplementary Tables S.\kredit{16}--S.\kredit{19}}.}
  \label{fig_16_dysts_log_plot}
\end{figure}

Nonlinear dynamics is ubiquitous in nature and typically governed by a set of differential equations. Distilling such governing equations from limited observed data plays a crucial role in better understanding the fundamental mechanism of dynamics. Here, we test the proposed PSE model to discover a series of multi-dimensional autonomous chaotic dynamics (e.g., Lorenz attractor \cite{lorenz1963deterministic} and its variants \cite{gilpin2021chaos_dysts}). The synthetic datasets of these chaotic dynamical systems are described in \textcolor{blue}{Supplementary Note 2.2}. It is noted that we only \kredit{sample} the noisy trajectories while determining the velocity states via smoothed numerical differentiation. We compare PSE with eight pivotal baseline models (namely, Bayesian machine scientist (BMS) \cite{guimera2020bayesian}, PySR \cite{cranmer2023interpretable_PySR}, NGGP \cite{mundhenk2021symbolic_NGGP}, DGSR \cite{holt2023deep_DGSR}, wAIC \cite{gao2022autonomous_NCS}, TPSR \cite{Shojaee_reviewer_NIPS2023_TPSR}, uDSR \cite{uDSR_landajuela2022unified}, and Operon \cite{Operon}) and run each model on 50 different random seeds to calculate the average recovery rate for each dataset. For wAIC, we employed two distinct basis function configurations: one utilizing polynomial basis functions (wAIC1), and another combining polynomial basis functions with the same unary operator basis used by other algorithms (wAIC2). Considering the noise effect, the criterion for successful equation recovery in this experiment is defined as follows: the discovered Pareto front covers the structure of the ground truth equation (allowing for a constant bias term). Since the dynamics of each system is governed by multiple coupled differential equations (e.g., 3$\sim$4 as shown in \textcolor{blue}{Supplementary Table S.3}--\textcolor{blue}{S.6}), the discovery is conducted independently to compute the average recovery rate for each equation, among which the minimum rate is taken to represent each model's capability.

Our main focus herein is to investigate whether SR methods can successfully recover the underlying differential equations given a specific set of token operators under the limit of a short period of computational time (e.g., a few minutes). In our experiments, we set the candidate binary operators as $+$, $-$, $\times$, and $\div$, while the candidate unary operators include $\sin$, $\cos$, $\exp$, $\log$, $\tanh$, $\cosh$, $\mathrm{abs}$, and $\mathrm{sign}$. Fig. \ref{fig_16_dysts_log_plot}\textbf{a}--\textbf{b} depicts the results of discovering the closed-form governing equations for 16 chaotic dynamical systems. The experiments demonstrate that under four different levels of Gaussian noise (1\%, 2\%, 5\%, and 10\% of the data's standard deviation), the proposed PSE approach can achieve a much higher symbolic recovery rate (see Fig. \ref{fig_16_dysts_log_plot}\textbf{b}), enabling to identify more accurately the underlying governing equations, even under noise effect, to better describe the chaotic behaviors (see Fig. \ref{fig_16_dysts_log_plot}\textbf{a}). This substantiates the capability and efficiency of our method on data-driven discovery of governing laws for more complex chaotic dynamical systems beyond the Lorenz attractor. More detailed results are listed in \textcolor{blue}{Supplementary Tables S.\kredit{16}--S.\kredit{19}}.

\subsection*{Electro-mechanical positioning system}
\label{subsubsec:EMPS}

\begin{figure}[t!]
  \centering
  \includegraphics[width=0.995\textwidth]{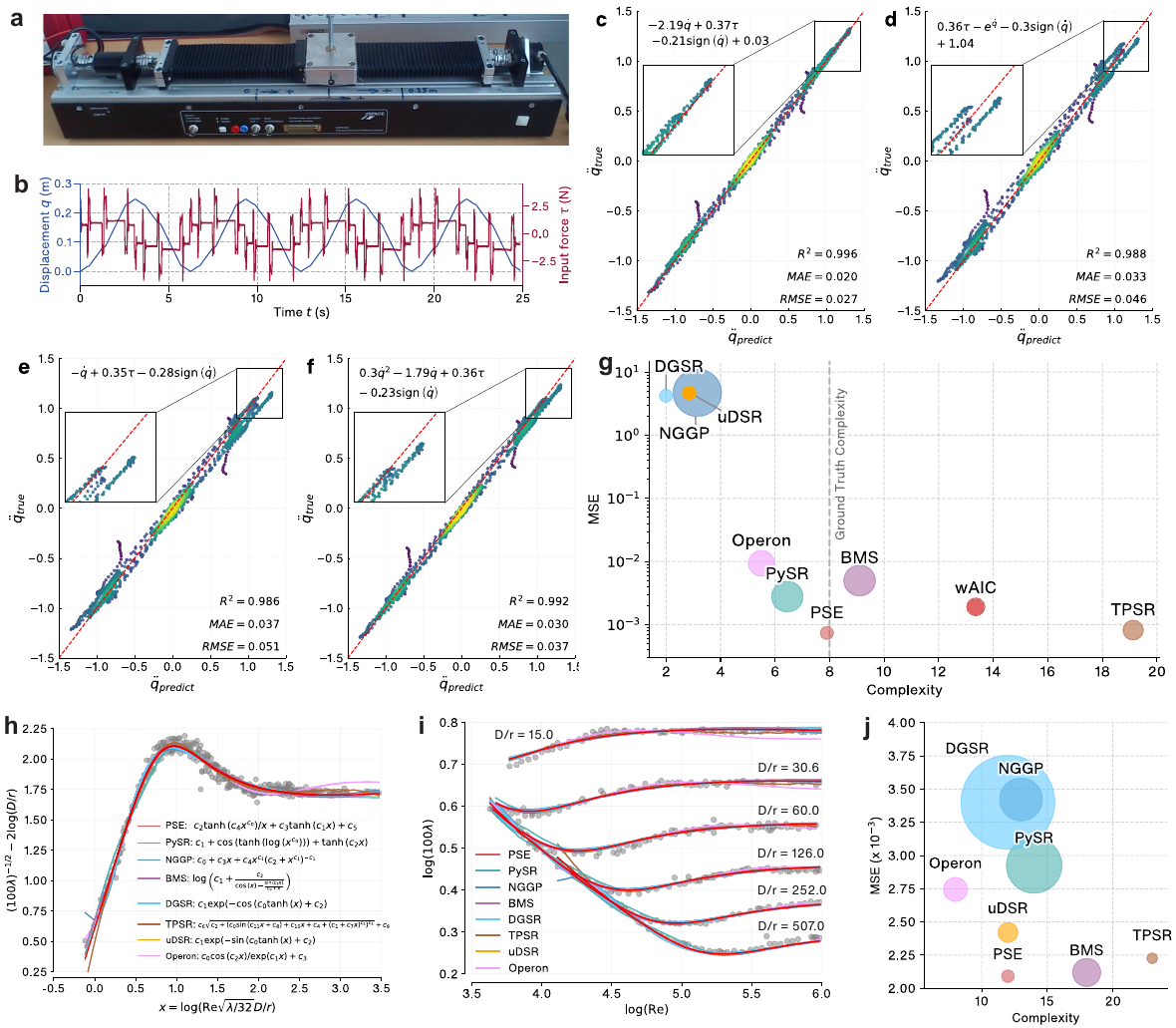}
  \caption{\textbf{Discovering the underlying physical laws with experimental data.} \textbf{a}, The setup of the EMPS experiment. \textbf{b}, Collected displacement and input force data. \textbf{c}, The prediction performance along with a typical governing equation discovered by PSE. \textbf{d}, The prediction performance along with a typical governing equation discovered by PySR. \textbf{e}, The prediction performance along with a typical governing equation discovered by BMS. \textbf{f}, The prediction performance along with a typical governing equation discovered by wAIC. Additional results for other methods on the EMPS experiment can be found in \textcolor{blue}{Supplementary Fig. S.3}. \textbf{g}, The prediction error versus model complexity for different symbolic regression methods on the EMPS dataset, averaged over 20 independent runs. The circle size indicates MSE uncertainty (interquartile range). \textbf{h}, The transformed (or collapsed) Nikuradse's dataset and discovered equations of turbulent friction by different SR methods. The legend shows the median reward models (see Eq. \eref{eq:reward}), among 20 trials, obtained by PSE (red), PySR (green) \cite{cranmer2023interpretable_PySR}, NGGP (purple) \cite{mundhenk2021symbolic_NGGP}, DGSR (skyblue) \cite{holt2023deep_DGSR}, uDSR (yellow) \cite{uDSR_landajuela2022unified}, TPSR (brown) \cite{Shojaee_reviewer_NIPS2023_TPSR}, BMS (blue) \cite{guimera2020bayesian}, and Operon (pink) \cite{Operon}. \textbf{i}, The fitting performance of each SR method on the original Nikuradse's data. \textbf{j}, The prediction mean square error (MSE) versus model complexity for different symbolic regression methods, averaged over 20 independent runs. The circle size indicates MSE uncertainty (interquartile range). Notably, our PSE method excels at fitting turbulent friction data rapidly (e.g., within only 1.5 min), meanwhile discovering a more parsimonious and accurate equation. }
  \label{fig_emps}
\end{figure}

Real-world data, replete with intricate noise and nonlinearity, may hinder the efficacy of SR algorithms. To further validate the capability of our PSE model in uncovering the governing equations for real-world dynamical systems (e.g., mechanical devices), we test its performance on a set of lab experimental data of an electro-mechanical positioning system (EMPS) \cite{janot2019data_EMPS}, as shown in Fig. \ref{fig_emps}\textbf{a}--\textbf{b}. The EMPS setup is a standard configuration of a drive system used for prismatic joints in robots or machine tools. Finding the governing equation of such a system is crucial for designing better controllers and optimizing system parameters. The dataset was bifurcated into two even parts, serving as the training and testing sets, respectively. The reference governing equation is given by $M\ddot{q}=-F_v\dot{q}-F_c\mathrm{sign}(\dot{q})+\tau-c$ \cite{janot2019data_EMPS}, where $q$, $\dot{q}$, and $\ddot{q}$ represent joint position, velocity, and acceleration. Here, $\tau$ is the joint torque/force; $c$, $M$, $F_v$, and $F_c$ are all constant parameters in the equation. Notably, we leveraged the \textit{a priori} knowledge that the governing equation of such a system follows the Newton's second law, with a general form $\ddot{q} = f(\dot{q}, q, \tau)$. Our objective is to uncover the closed form of $f$ based on a predefined set of token operators. In EMPS, there exists friction that dissipates the system energy. Based on this prior knowledge, we include the $\mathrm{sign}$ operator to model such a mechanism. Hence, the candidate math operators we use to test the SR algorithms read $\{+,-,\times,\div, \sin, \cos, \exp, \log, \mathrm{cosh}, \mathrm{tanh}, \mathrm{abs}, \mathrm{sign}\}$.

We compare our PSE model with eight pivotal baseline models, namely, PySR \cite{cranmer2023interpretable_PySR}, NGGP \cite{mundhenk2021symbolic_NGGP}, DGSR \cite{holt2023deep_DGSR}, BMS \cite{guimera2020bayesian}, uDSR \cite{uDSR_landajuela2022unified}, TPSR \cite{Shojaee_reviewer_NIPS2023_TPSR}, wAIC \cite{gao2022autonomous_NCS} and Operon \cite{Operon}. We execute each SR model 20 trials on the training dataset to ascertain the Pareto fronts. Subsequently, we select the discovered equation from the candidate expression set of each Pareto front based on the test dataset which exhibits the highest reward value delineated in Eq. \eref{eq:reward}. The reward is designed to balance the prediction error and the complexity of the discovered equation, which is crucial to deriving the governing equation that is not only as consistent with the data as possible but also parsimonious and interpretable. Finally, we select among 20 trials the discovered equation with the median reward value to represent each SR model's average performance. The results demonstrate that our PSE model achieves the best performance in successfully discovering the underlying governing equation (see Fig. \ref{fig_emps}\textbf{c}--\textbf{f} and \textcolor{blue}{Supplementary Fig. S.3}). Our model is capable of uncovering the correct governing equation while maintaining low prediction error (see Fig. \ref{fig_emps}\textbf{g}).

\subsection*{Governing equation of turbulent friction}
\label{subsubsec:turbulent}

Uncovering the intrinsic relationship between fluid dynamics and frictional resistance has been an enduring pursuit in the field of fluid mechanics, with implications spanning engineering, physics, and industrial applications. In particular, one fundamental challenge lies in finding a unified formula to quantitatively connect the Reynolds number ($Re$), the relative roughness $r/D$, and the friction factor $\lambda$, based on experimental data. The Reynolds number, a dimensionless quantity, captures the balance between inertial and viscous forces within a flowing fluid, which is a key parameter in determining the flow regime, transitioning between laminar and turbulent behaviors. The relative roughness, a ratio between the size of the irregularities and the radius of the pipe, characterizes the interaction between the fluid and the surface it flows over. Such a parameter has a substantial influence on the flow's energy loss and transition to turbulence. The frictional force, arising from the interaction between the fluid and the surface, governs the dissipation of energy and is crucial in determining the efficiency of fluid transport systems. The groundbreaking work \cite{nikuradse1950laws} dated in the 1930s stepped out the first attempt by meticulously cataloging in the lab the flow behavior under friction effect. The experimental data, commonly referred to as the Nikuradse dataset, offers insights into the complex interplay between these parameters ($Re$ and $r/D$) and the resultant friction factor ($\lambda$) in turbulent flows. We herein test the performance of the proposed PSE model in uncovering the underlying law that governs the relationship between fluid dynamics and frictional resistance based on the Nikuradse dataset.

We firstly transform the data by a data collapse approach \cite{prandtl1933recent}, a common practice used in previous studies \cite{goldenfeld2006roughness,turbulent_tao2009}. Our objective is to find a parsimonious closed-form equation given by $\bar{\lambda} = \bar{h}(x)$, where $\bar{\lambda} = \lambda^{-1/2} + 2\log{(r/D)}$ denotes the transformed friction factor, $x = Re\sqrt{\lambda/32}(D/r)$ an intermediate variable, and $\bar{h}$ the target function to be discovered. We consider seven baseline models for comparison, namely, PySR \cite{cranmer2023interpretable_PySR}, NGGP \cite{mundhenk2021symbolic_NGGP}, DGSR \cite{holt2023deep_DGSR}, uDSR \cite{uDSR_landajuela2022unified}, TPSR \cite{Shojaee_reviewer_NIPS2023_TPSR}, BMS \cite{guimera2020bayesian}, and Operon \cite{Operon}. The candidate operators used in these models read $\{+,\times, -,\div, \sin, \cos, \exp, \log, \allowbreak \tanh,\allowbreak \cosh, \allowbreak \square^2, \square^3\}$. We run each SR model for 20 independent trials with different random seeds and choose the identified expression with the median reward as the representative result. For each trial, we report the expression with the highest reward (see Eq. \eref{eq:reward}) on the discovered Pareto front. Fig. \ref{fig_emps}\textbf{h} illustrates discovered governing equations for turbulent friction. The equations discovered by each SR method and the corresponding prediction error versus model complexity are shown in Fig. \ref{fig_emps}\textbf{h}--\textbf{j}. While several methods achieve comparable mean square error (MSE) performance, our PSE model stands out by producing the best balance between fitting accuracy and expression simplicity (see Fig. \ref{fig_emps}\textbf{j}). It is evident from the results that our method not only matches or surpasses the fitting performance of other approaches but also generates significantly more concise symbolic expressions.

\subsection*{Scalability in high-dimensional space}
\label{subsec:high_dimension}

To further assess the scalability of PSE and its capability for implicit feature selection, we introduced a challenging high-dimensional synthetic benchmark. This benchmark comprises 20 problems where the ground-truth equations, each involving 12 active variables, are embedded in a 50-dimensional input space, with the remaining 38 variables acting as distractors. In this high-dimensional and noisy environment, PSE demonstrated remarkable robustness by successfully recovering 40\% of the true expressions. In stark contrast, leading baselines like PySR and Operon failed to identify any correct solutions (0\% recovery rate) within the same computational time budget. This result highlights PSE's distinct advantage in navigating vast search spaces while effectively discerning relevant variables, a critical capability for tackling complex real-world problems. The detailed description of the methodology, equations, and full results can be found in \textcolor{blue}{Supplementary Note 7}.

\subsection*{Model performance analysis}
\label{subsec:model_performance_analysis}

The performance of the proposed PSE model depends on several factors, including the noise level of the given data, model hyper-parameters, and whether certain modules are used. Herein, we present an analysis of these factors as well as the model efficiency and the memory footprint.

\begin{figure}[t!]
  \hspace*{-0.1in}
  \centering
  \includegraphics[width=6.6in]{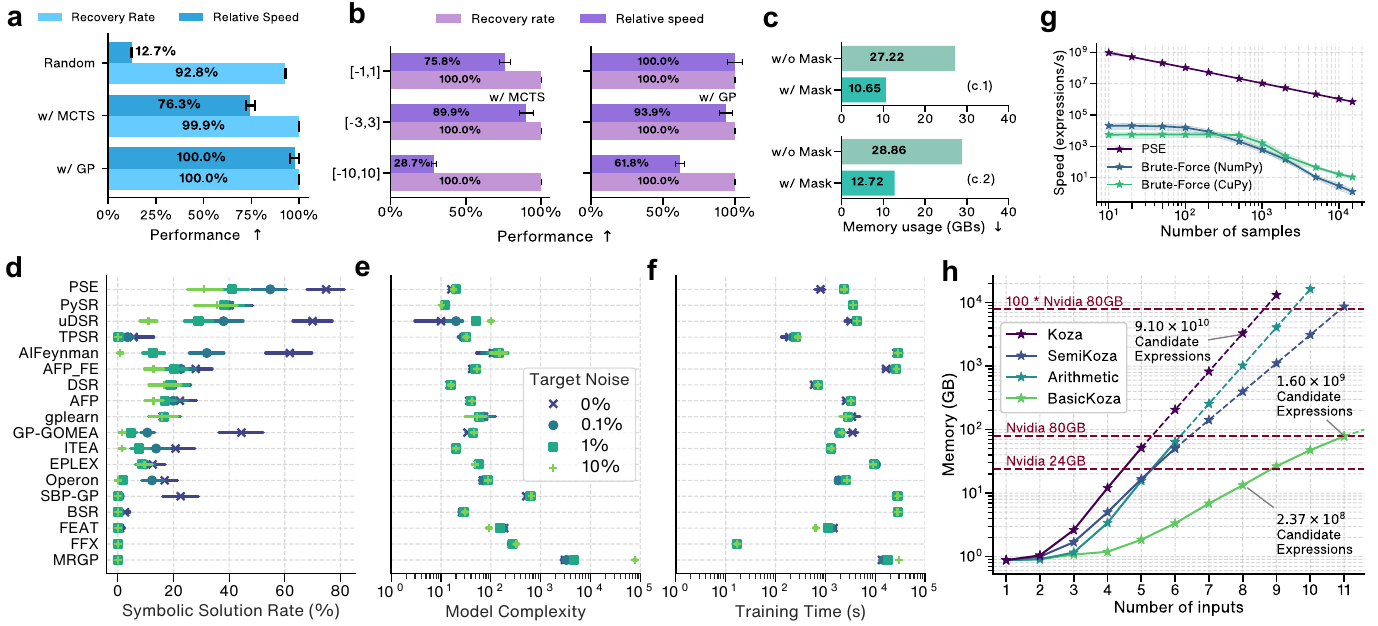}
  \caption{\textbf{Ablation study of the proposed PSE method.} \textbf{a}, Comparison of token generators: Random, MCTS, and GP. \textbf{b}, Ablation of token constants range. \textbf{c}, Ablation of DR Mask on two different PSE configurations: (c.1) a 4-input, 3-layer PSRN with $\mathcal{O}_{\mathrm{Koza}}$  library (e.g., $\{+$, $ \times$, $-$, $\div$, identity, $\sin$, $\cos$, $\exp$, $\log\}$), and (c.2) a 5-input, 3-layer PSRN with $\mathcal{O}_{\mathrm{SemiKoza}}$  library (e.g., $\{+$, $\times$, SemiSub, SemiDiv, identity, neg, inv, $\sin$, $\cos$, $\exp$, $\log\}$). \textbf{d-f}, SRBench \cite{SRBench_la2021contemporary} results with GP as the token generator. Our method achieves the highest symbolic recovery rate across all noise levels (aka, 0, 0.1\%, 1\%, and 10\% of the data's standard deviation) while maintaining competitive performance in model complexity and training time.
  \textbf{g}, Expression search efficiency of the PSRN module. \textbf{h}, Space complexity of PSRN with three Symbol Layers, with respect to the number of input slots and memory footprints. We incorporate four operator sets: $\mathcal{O}_{\mathrm{Koza}}$, $\mathcal{O}_{\mathrm{SemiKoza}}$, $\mathcal{O}_{\mathrm{Arithmetic}}$ (e.g., $\{+$, $\times$, $-$, $\div$, identity$\}$), and $\mathcal{O}_{\mathrm{BasicKoza}}$  (e.g., $\{+$, $\times$, identity, $\mathrm{neg}$, $\mathrm{inv}$, $\sin$, $\cos$, $\exp$, $\log\}$) for comparison. With the expansion of the memory footprint, our model demonstrates scalability by evaluating a greater number of candidate expressions in a single forward pass, leading to enhanced performance. We strictly maintain consistent runtime budgets. The variations in the algorithms' runtime occur because different algorithms trigger their respective early stopping conditions (e.g., reaching an MSE below a very small threshold, or discovering symbolically equivalent expressions, etc.).}
  \label{fig_ablation_merge}
\end{figure}

\paragraph{Model ablation study.}
\label{subsec:ablation}

We conducted three cases of model ablation study to evaluate the role of certain modules. First, we investigate how much improvement the use of GP or MCTS as the token generator for automatic discovery of admissible tokenized input brings. Second, we conduct a sensitivity analysis of the token constants range. Third, we investigate the extent of the benefit of using DR Mask. The results of the ablation experiments are shown in Fig. \ref{fig_ablation_merge}\textbf{a}--\textbf{c}.

Firstly, we evaluate the symbolic recovery rates to compare different token generation methods for PSRN, including random generation, MCTS, and GP. The tests are conducted on the entire Nguyen, Nguyen-c, R, R$^*$, Livermore and Feynman problem sets. It can be observed in Fig. \ref{fig_ablation_merge}\textbf{a} that the recovery rate and relative speed diminishes if the token generator is removed from PSE, indicating its vital role in admissible token search that signals the way forward for expression exploration. Additionally, we observe that using GP as the token generator leads to faster discovery of ground truth expressions compared to MCTS, despite their similar symbolic recovery rates. Secondly, to investigate the sensitivity of our model to the randomly sampled token constants, we set the range to $[-1,1]$, $[-3,3]$, and $[-10,10]$ respectively, and performed the experiments on the Nguyen-c benchmark expressions. The result in Fig. \ref{fig_ablation_merge}\textbf{b} shows that when the range of constant sampling is altered, it significantly affects the time required to find the ground truth expression, but has little impact on the symbolic recovery rate. Last but not least, we test the efficacy of DR Mask for memory saving. The result in Fig. \ref{fig_ablation_merge}\textbf{c} illustrates that DR Mask is able to save the graphic memory up to 50\%, thus improving the capacity of the operator set (e.g., more operators could be included to represent complex expressions).

\paragraph{SRBench evaluation and performance.}
\label{subsec:robust}

We further tested the performance of our model on the SRBench problem sets \cite{SRBench_la2021contemporary}, evaluating 133 mathematical expressions on datasets with four levels of Gaussian-type noise (aka, 0, 0.1\%, 1\%, and 10\% of the data's standard deviation). The results of symbolic recovery rate, model complexity, and computational efficiency for our PSE model in comparison with other 17 baseline models are depicted in Fig. \ref{fig_ablation_merge}\textbf{d}--\textbf{f}. It can be seen that our method significantly outperforms all these existing algorithms in terms of symbolic recovery rate across all noise conditions, while maintaining competitive performance in the context of model complexity and computational efficiency. Note that the results were validated using 10 different random seeds following the standard evaluation protocol in \cite{SRBench_la2021contemporary}, with 95\% confidence intervals calculated, which demonstrated PSE's robustness to handle real-world noise interference while producing concise expressions with relatively short training time required. These findings underscore the effectiveness of our PSE method.

\paragraph{Expression search efficiency.}
\label{expr_search_effi}

We compare the efficiency of PSE in the context of evaluation of large-scale candidate expressions, in comparison with two brute-force methods (e.g., NumPy \cite{NumPy} that performs CPU-based evaluation serially, and CuPy \cite{CuPy} that operates on GPUs in parallel based on batches). Note that PSE possesses the capability of automatic identification and evaluation of common subtrees, while NumPy and CuPy do not have such a function. Assuming independent input variables $\{x_1, \ldots,x_5\}$ with operators $\{+$, $\times$, identity, neg, inv, sin, cos, exp, log$\}$ for a maximum tree depth of 3, the complete set of generated expressions is denoted by $\hat{\mathcal{F}}$. We consider the computational time required to evaluate the loss values of all the expressions, e.g., $||\mathbf{y}-\hat{f}(\mathbf{X})||_2^2$ where $\hat{f}\in \hat{\mathcal{F}}$, under different sample sizes (e.g., $10^{1}\sim 10^{4}$ data points).

The result shows that PSE can quickly evaluate all the corresponding expressions, clearly surpassing the brute-force methods (see Fig. \ref{fig_ablation_merge}\textbf{g}). When the number of samples is less than $10^4$, the search speed of PSE is about 4 orders of magnitude faster, exhibiting an unprecedented increase in efficiency, thanks to the reuse of common subtree evaluation in parallel. Notably, when the number of samples is big, downsampling of the data is suggested in the process of uncovering the equation structure in order to take the speed advantage of PSE during forward propagation. In the coefficient estimation stage, all samples should be used. This could further increase the efficiency of PSE while maintaining accuracy.

\paragraph{Space complexity.}
\label{subsec:space_complexity}

We categorize the operators used in PSE into three types: unary, binary-squared, and binary-triangled, which are represented by $u$, $b_S$, and $b_T$, respectively. Binary-squared operators represent non-commutative operators (e.g., $-$ and $\div$) depending on the order of operands, which requires $\omega_{i-1}^2$ space on GPU during PSRN forward propagation (here, $\omega_{i-1}$ represents the input size of the previous symbol layer). Binary-triangled operators represent commutative operators (e.g., $+$ and $\times$) or the memory-saving version of non-commutative operators that only take up $\omega_{i-1}(\omega_{i-1}+1)/2$ space (e.g., the SemiSub and SemiDiv symbols, described in \nameref{sec:methods}: \nameref{subsec:symbol_layer}, in the Feynman benchmark which are specifically designed to save memory and support operations in only one direction).  

With the number of operators in each category denoted by $N_u$, $N_{b_S}$ and $N_{b_T}$, and the number of independent variables and layers of PSE denoted by $m$ and $l$, respectively, the number of floating-point values required to be stored in PSE can be analyzed. Ignoring the impact of DR Mask, there is a recursive relationship between the tensor dimension of the $(i-1)$-th layer (e.g., $\omega_{i-1}$) and that of the $i$-th layer (e.g., $\omega_i$), namely,

\begin{equation*}
    \omega_{i} = N_u  \omega_{i-1} + N_{b_S} \omega_{i-1}^2 +  N_{b_T}\omega_{i-1}  \frac{\omega_{i-1} + 1}{2} \le \kappa\omega_{i-1}^2,  
\end{equation*}
where $\kappa = N_u + N_{b_T} + N_{b_S}$. Thus the complexity of the number of floating-point values required to be stored by an 
 $l$-layer PSRN can be calculated as $O(\kappa ^ {2 ^ l -1}\omega_0 ^ {2 ^ l})$, where $\omega_0$ represents the number of input slots.

Clearly, the memory consumption of PSRN increases dramatically with the number of layers. If each subtree value is a single-precision floating-point number (e.g., 32-bit), with the input dimension of 20 and operator set $\{+,-,\times,\div,\mathrm{identity},\sin,\cos,\exp,\log\}$, the required memory will reach over $10^{5}$ GBs when the number of layers is 3, which requires a large compute set. Hence, finding a new strategy to relax the space complexity and alleviate the memory requirement is needed to further scale up the proposed model. Fig. \ref{fig_ablation_merge}\textbf{h} illustrates the graphic-memory footprint of various three-layered PSRN architectures, each characterized by a different operator set and the number of input slots. While the rise in memory demands serves as a constraint, this escalation is directly tied to the scalable model’s ability to evaluate a greater number of candidate expressions within a single forward pass. This result also shows that PSRN follows the scaling law (e.g., the model capacity and size scale with the number of token inputs, given the fixed number of layers). The detailed hardware settings are found in \textcolor{blue}{Supplementary Note 2.5}.

\section*{Discussion}
\label{sec:discussion}

This paper introduces a new SR method, called PSE, for the automatic and efficient discovery of parsimonious equations from data. In particular, we propose a PSRN architecture as the core search engine, which (1) automatically captures common subtrees of different symbolic expression trees for shared evaluation to expedite the computation, and (2) capitalizes on a parallel computing paradigm for a notable performance boost. By recognizing and exploiting common subtrees across a vast number of candidate expressions, PSRN bypasses the redundant evaluations inherent in methods that assess each candidate expression independently. This distinction is particularly evident when compared to explicit enumeration methods like Exhaustive Symbolic Regression (ESR) \cite{ESR_Bartlett_2022}, which, while complete, face a significant computational bottleneck by constructing every possible symbolic expression before evaluation—a process that is both time-consuming and practically limited to single-variable problems (see \textcolor{blue}{Supplementary Note 9}). By shifting the paradigm from sequential evaluation to sub-tree reusing and large-scale parallel enumeration, this approach opens a path toward solving more complex and higher-dimensional discovery problems. When coupled with a token generator (e.g., GP or MCTS), the model's ability to delve into complex expressions is further magnified, showing potential to advance the solving of more complex SR problems.

The efficacy of PSE has been extensively evaluated on a variety of complex mathematical equations from both synthetic and experimental datasets, including multiple benchmark problem sets (e.g., Nguyen, Nguyen-c, R, R$^\ast$, Livermore, and Feynman), SRBench, nonlinear chaotic dynamics, and two datasets from lab experiments (EMPS and turbulent friction). We have demonstrated that PSE possesses a powerful capability in general-purpose SR, exhibiting a higher recovery rate and speed. The performance of PSE exceeds several representative baseline models, achieving up to two orders of magnitude efficiency improvement while maintaining higher accuracy. Moreover, even for a target equation of a very complex form, the PSE model is capable of swiftly and reliably uncovering the ground truth directly, resisting the local minima that can arise from ambiguous patterns in the data. Consequently, PSE excels in discovering accurate and parsimonious expressions from very limited data within a short time budget.

Despite its performance, the PSE model faces several challenges that need to be addressed in the future. One of the primary challenges is the rapidly increasing demand for memory in the PSRN module as the number of symbol layers increases (see \textcolor{blue}{Results:} \nameref{subsec:model_performance_analysis}: \nameref{subsec:space_complexity}). Currently, a brute-force implementation of PSRN can only directly handle expressions with an expression tree depth of $\leq 3$ under conventional settings. Otherwise, the method relies on the token generator (e.g., GP or MCTS) to extend PSRN's capacity by providing more complex sub-expressions as tokenized inputs. This bottleneck impedes the PSE's exploration of much deeper expressions, especially for problems requiring extremely deep, monolithic expressions that cannot be easily decomposed into smaller tokens. As detailed in our failure analysis (see \textcolor{blue}{Supplementary Note 8}), this is because the token generator itself may struggle to construct a highly complex sub-expression as a single, coherent unit, leading the search to become trapped in local optima of simpler, ``good-enough'' solutions, as the marginal accuracy gain from the true, highly complex structure may not provide a sufficient reward signal to justify the difficult evolutionary leap. Additionally, our current two-stage approach to handling constants—sampling token constants and then refining them via least squares—is effective but not without limitations. If the true coefficients lie far outside the initial sampling range, the least-squares refinement may fail to find the global optimum.

Furthermore, like all data-driven methods, the performance of PSE is inherently linked to the quality and nature of the data. For instance, its performance degrades under high noise levels, converging to simpler, over-regularized expressions that model the noise rather than the underlying signal. This is related to PSRN's strong affinity for fractional forms; while this is a significant advantage on certain benchmarks, under high noise it could lead to overfitting by modeling noise as a complex rational function (see \textcolor{blue}{Supplementary Note 8}). Similarly, the model may struggle with deceptive fitness landscapes, where a simple, incorrect expression has a similar MSE to the true, more complex one over the given data domain. In such cases, the algorithm's success hinges on the data being sufficiently informative to distinguish the true model from plausible alternatives. In terms of practical constraints, while the method can operate on various computing platforms, its full acceleration is best realized on hardware that supports massive parallelism, though the subtree reusing benefit remains hardware-independent. Like most SR methods, PSE is also constrained by its predefined, discrete operator set and cannot discover novel mathematical forms without being explicitly extended.

Looking ahead, there are several promising avenues for further optimization and enhancement. For instance, ensemble symbolic regression, namely, uDSR \cite{uDSR_landajuela2022unified}, integrates a wide array of techniques including large-scale pre-training \cite{holt2023deep_DGSR}, deep symbolic regression \cite{petersen2021deep}, sparse regression \cite{brunton2016discovering_SINDy}, and AIFeynman \cite{udrescu2020ai_AIFeynman}, and has demonstrated improved SR performance. This suggests significant potential for enhancing our approach by synthesizing PSE with other SR techniques. Similarly, integrating explicit feature selection methods as a pre-processing step could further focus the search on the most relevant variables, representing another promising avenue for enhancing scalability on extremely large-scale problems. The PSRN framework itself also offers opportunities for efficiency improvements, such as adopting more advanced computational backends that demonstrate superior performance in speed and memory utilization. This includes developing more sophisticated token generation strategies, as suggested by our failure analysis (see \textcolor{blue}{Supplementary Note 8}), such as incorporating domain-specific structural priors or employing multi-objective rewards that explicitly value structural novelty. Finally, ongoing work is exploring the integration of prior knowledge, such as the dimensional constraints of physical quantities, to guide the search. This approach promises not only to conserve computational resources but also to expedite the discovery process. We intend to continue addressing these challenges methodically in our forthcoming research.

\section*{Data availability} 
All the datasets used to test the methods in this study are available on GitHub at \url{https://github.com/intell-sci-comput/PSE}.

\section*{Code availability} 
All the source codes used to reproduce the results in this study are available on GitHub at \url{https://github.com/intell-sci-comput/PSE}.

\bibliographystyle{unsrt}
\bibliography{references}

\newpage
\vspace{36pt}
\noindent\textbf{Acknowledgement:}
The work is supported by the National Natural Science Foundation of China (No. 92270118, No. 62276269, and No. 62206299), the Beijing Natural Science Foundation (No. 1232009), and the Strategic Priority Research Program of the Chinese Academy of Sciences (No. XDB0620103). In addition, H.S and Y.L. would like to acknowledge the support from the Fundamental Research Funds for the Central Universities (No. 202230265 and No. E2EG2202X2). \\

\noindent\textbf{Author contributions:} K.R., H.S., Y.L. contributed to the ideation and design of the research; K.R. performed the research; H.S. and J.R.W. supervised the project; all authors contributed to the research discussions, writing, and editing of the paper. \\

\noindent\textbf{Correspondence to:} Hao Sun (\url{haosun@ruc.edu.cn}).\\

\noindent\textbf{Competing interests:}
The authors declare no competing interests.\\

\noindent\textbf{Supplementary information:}
The supplementary information is attached.

\clearpage
\setcounter{figure}{0}
\renewcommand{\figurename}{Extended Data Fig.}
\setcounter{table}{0}
\renewcommand{\tablename}{Extended Data Table}

\begin{figure}[htbp]
  \hspace*{-0.0in}
  \centering
  \includegraphics[width=0.97\textwidth]{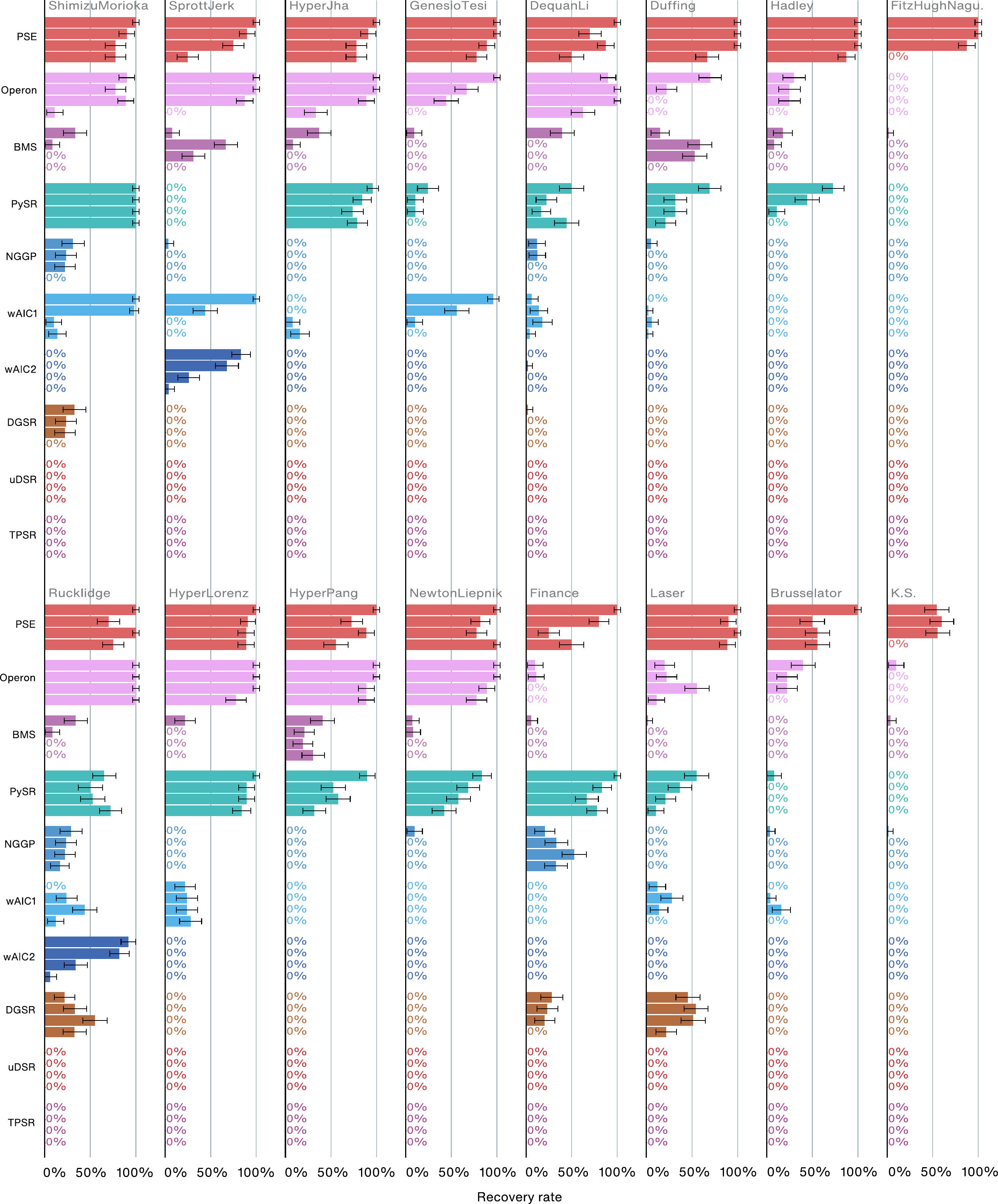}
    \caption{\textbf{Data-driven discovery of nonlinear chaotic dynamics by different models.} The average recovery rates of SR algorithms, including PSE, BMS, PySR, NGGP, DGSR, wAIC, uDSR, TPSR, and Operon, are compared on 16 nonlinear chaotic dynamics datasets. Given the same budget of computational time, PSE has a higher probability of finding the true governing equations among a nearly infinite number of possible expression structures. Each group of bars with the same color, from top to bottom, represents the recovery rates under four different levels of Gaussian noise (1\%, 2\%, 5\%, and 10\% of the data's standard deviation). Detailed results can be found in \textcolor{blue}{Supplementary Tables S.\kredit{16}--S.\kredit{19}}.}
  \label{fig_16_dysts_log_plot_bar}
\end{figure}

\end{document}


\title{\textbf{Supplementary Information} for: \\ Discovering physical laws with parallel symbolic enumeration}

\author[1]{Kai Ruan}
\author[1]{Yilong Xu}
\author[1]{Ze-Feng Gao}
\author[2,3]{Yang Liu}
\author[4]{Yike Guo} 
\author[1]{Ji-Rong Wen} 
\author[1,$^*$]{Hao Sun}

\affil[1]{\footnotesize Gaoling School of Artificial Intelligence, Renmin University of China, Beijing, China}
\affil[2]{\footnotesize School of Engineering Science, University of Chinese Academy of Sciences, Beijing, China}
\affil[3]{\footnotesize State Key Laboratory of Nonlinear Mechanics, Institute of Mechanics, Chinese Academy of Sciences, Beijing, China}
\affil[4]{\footnotesize Department of Computer Science and Engineering, HKUST, Hong Kong, China \vspace{18pt}} 

\affil[*]{Corresponding author.\vspace{12pt}}

\date{}

\maketitle

\vspace{-36pt}

{\footnotesize
\tableofcontents
}

\vspace{24pt}

\noindent This supplementary document provides a detailed description of the proposed algorithm, examples, and discussion of technical challenges for parallel symbolic enumeration (PSE).

\section{Background}
\begin{figure}[htbp]
  \hspace*{-0.0in}
  \centering
  \includegraphics[width=6.5in]{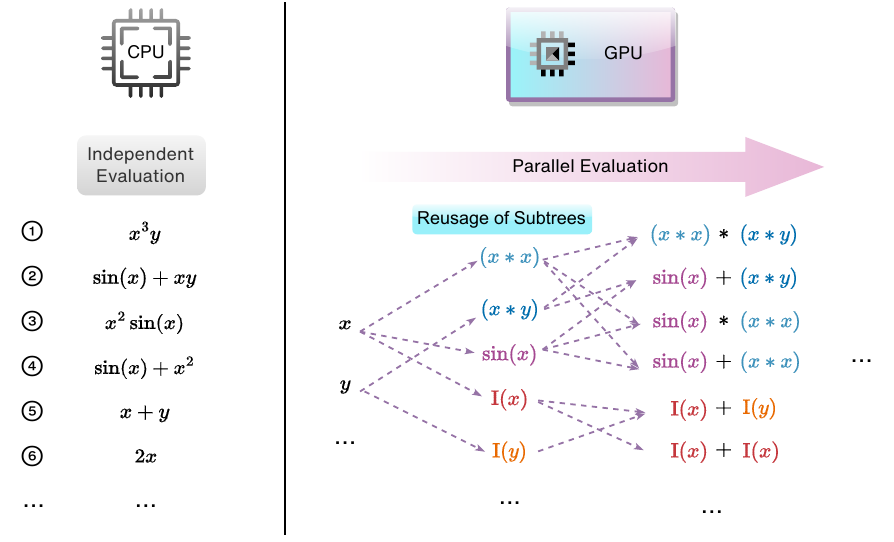}
  \caption{Comparison between CPU independent evaluation and GPU parallel evaluation. The PSRN significantly reduces unnecessary computational workload through the reuse of common subtree values when operating on GPUs. Subtrees bearing identical values are indicated by the same color, while the dashed lines trace the propagation of these subtree values throughout the PSRN architecture.}
  \label{fig_cpu_and_gpu}
\end{figure}

As far as we know, existing SR algorithms inevitably rely on a necessary step, namely evaluating large-scale candidate expressions $\hat{\mathcal{F}}=\{ \hat{f_1}, \hat{f_2}, \ldots \}$ on the given data $\mathcal{D}=(\mathbf{X},\mathbf{y})$ to obtain the error. While some methods involving the use of neural networks can generate candidate expressions using GPUs, the process of obtaining the error for each mathematical expression still depends on sequential and independent evaluations using CPUs. When the number of expressions to be evaluated becomes sufficiently large, there will be a significant amount of redundant evaluations of common subtrees inevitably. In Figure \ref{fig_cpu_and_gpu}, we illustrate the fundamental difference between the PSRN module of PSE and existing methods in terms of symbolic expression evaluation.

\clearpage
\section{Methodology}

\subsection{Benchmark Symbolic Regression Problems}
We employed the Nguyen, Nguyen-c \cite{uy2011semantically_Nguyen,mcdermott2012genetic}, R, R* \cite{krawiec2013approximating_Rbenchmark}, Livermore \cite{mundhenk2021symbolic_NGGP}, and Feynman \cite{udrescu2020ai_AIFeynman} benchmark problem sets to evaluate the symbolic recovery rate and recovery speed of the algorithms. The specific benchmark problem settings were consistent with those described in the NGGP \cite{mundhenk2021symbolic_NGGP} and DGSR \cite{holt2023deep_DGSR}. Table \ref{table:bm_1} and \ref{table:bm_2} shows these SR benchmark problems.

\begin{table}[b!]
\centering
\caption{Nguyen, Nguyen-c, R and R* Benchmark Problems. $U$ denotes uniform sampling over the interval, while $E$ denotes equidistant sampling over the interval. The three parameters $(a, b, c)$ represent the lower bound, upper bound, and the number of sampling points within the interval. The operator set used is $\{+,\times, -,\div,\sin, \cos,\exp, \log\}$.}
\begin{tabular}{ccccHc}
\toprule
\textbf{Benchmark}    & \textbf{Expression}                      & \textbf{Dataset} & \textbf{Tokens}  \\
\midrule
Nguyen-1     & $x_1^3+x_1^2+x_1$                     & $U(-1,1,20)$ & $\{x_1\}$\\
Nguyen-2     & $x_1^4+x_1^3+x_1^2+x_1$                 & $U(-1,1,20)$ & $\{x_1\}$\\
Nguyen-3     & $x_1^5+x_1^4+x_1^3+x_1^2+x_1$             & $U(-1,1,20)$ & $\{x_1\}$\\
Nguyen-4     & $x_1^6+x_1^5+x_1^4+x_1^3+x_1^2+x_1$         & $U(-1,1,20)$ & $\{x_1\}$\\
Nguyen-5     & $\sin(x_1^2)\cos(x_1)-1$              & $U(-1,1,20)$ & $\{x_1\}$\\
Nguyen-6     & $\sin(x_1)+\sin(x_1+x_1^2)$           & $U(-1,1,20)$ & $\{x_1\}$\\
Nguyen-7     & $\log(x_1+1)+\log(x_1^2+1)$          & $U(0,2,20)$ & $\{x_1\}$\\
Nguyen-8     & $\sqrt{x_1}$                      & $U(0,4,20)$ & $\{x_1\}$\\
Nguyen-9     & $\sin(x_1)+\sin(x_2^2)$             & $U(0,1,20)$ & $\{x_1, x_2\}$\\
Nguyen-10    & $2\sin(x_1)\cos(x_2)$               & $U(0,1,20)$ & $\{x_1, x_2\}$\\
Nguyen-11    & $x_1^{x_2}$                           & $U(0,1,20)$ & $\{x_1, x_2\}$\\
Nguyen-12    & $x_1^4-x_1^2+\frac{1}{2}x_2^2-x_2$      & $U(0,1,20)$ & $\{x_1, x_2\}$\\
\midrule
Nguyen-1c    & $3.39x_1^3+2.12x_1^2+1.78x_1$        & $U(-1,1,20)$ & $\{x_1,  \mathrm{const}\}$\\
Nguyen-2c    & $0.48x_1^4+3.39x_1^3+2.12x_1^2+1.78x_1$ & $U(-1,1,20)$ & $\{x_1,  \mathrm{const}\}$\\
Nguyen-5c    & $\sin(x_1^2)\cos(x_1)-0.75$           & $U(0,2,20)$ & $\{x_1,  \mathrm{const}\}$\\
Nguyen-8c    & $\sqrt{1.23x_1}$                   & $U(0,4,20)$ & $\{x_1,  \mathrm{const}\}$\\
Nguyen-9c    & $\sin(1.5x_1)+\sin(0.5x_2^2)$         & $U(0,1,20)$ & $\{x_1, x_2, \mathrm{const}\}$\\
Nguyen-10c    & $\sin(1.5x_1)\cos(0.5x_2)$         & $U(0,1,20)$ & $\{x_1, x_2, \mathrm{const}\}$\\
\midrule
R-1           & $(x_1+1)^3/(x_1^2-x_1+1)$             & $E(-1,1,20)$ & $\{x_1\}$\\
R-2           & $(x_1^5-3x^3+1)/(x_1^2+1)$          & $E(-1,1,20)$ & $\{x_1\}$\\
R-3           & $(x_1^6+x_1^5)/(x_1^4+x_1^3+x_1^2+x_1+1)$   & $E(-1,1,20)$ & $\{x_1\}$\\
\midrule
R-1*           & $(x_1+1)^3/(x_1^2-x_1+1)$             & $E(-10,10,20)$ & $\{x_1\}$\\
R-2*           & $(x_1^5-3x^3+1)/(x_1^2+1)$          & $E(-10,10,20)$ & $\{x_1\}$\\
R-3*           & $(x_1^6+x_1^5)/(x_1^4+x_1^3+x_1^2+x_1+1)$   & $E(-10,10,20)$ & $\{x_1\}$\\
\bottomrule

\end{tabular}
\label{table:bm_1}

\end{table}

\begin{table}[htbp]
\centering
\caption{Livermore and Feynman Benchmark Problems. $U$ denotes uniform sampling over the interval, while $E$ denotes equidistant sampling over the interval. The three parameters $(a, b, c)$ represent the lower bound, upper bound, and the number of sampling points within the interval. The operator set used is $\{+,\times, -,\div,\sin, \cos,\exp, \log\}$.}
\resizebox{0.99\textwidth}{!}{
\begin{tabular}{ccccHc}
\toprule
\textbf{Benchmark}    & \textbf{Expression}                      & \textbf{Dataset} & \textbf{Tokens}  \\
\midrule
Livermore-1  & $1/3+x_1+\sin(x_1^2)$                & $U(-10,10,1000)$ & $\{x_1\}$\\
Livermore-2  & $\sin(x_1^2)\cos(x_1)-2 $            & $U(-1,1,20)$ & $\{x_1\}$\\
Livermore-3  & $\sin(x_1^3)\cos(x_1^2)-1$           & $U(-1,1,20)$ & $\{x_1\}$\\
Livermore-4  & $\log(x_1+1)+\log(x_1^2+1)+\log(x_1)$   & $U(0,2,20)$ & $\{x_1\}$\\
Livermore-5  & $x_1^4-x_1^3+x_1^2-x_2$                 & $U(0,1,20)$ & $\{x_1, x_2\}$\\
Livermore-6  & $4x^4+3x^3+2x^2+x_1$              & $U(-1,1,20)$ & $\{x_1\}$\\
Livermore-7  & $\sinh(x_1)$                      & $U(-1,1,20)$ & $\{x_1\}$\\
Livermore-8  & $\cosh(x_1)$                       & $U(-1,1,20)$ & $\{x_1\}$\\
Livermore-9  & $\scalebox{0.9}{$x_1^9+x_1^8+x_1^7+x_1^6+x_1^5+x_1^4+x_1^3+x_1^2+x_1$}$ & $U(-1,1,20)$ & $\{x_1\}$\\
Livermore-10 & $6\sin(x_1)\cos(x_2)$                 & $U(0,1,20)$ & $\{x_1,x_2\}$\\
Livermore-11 & $x_1^4/(x_1+x_2)$                     & $U(-1,1,50)$ & $\{x_1,x_2\}$\\
Livermore-12 & $x_1^5/x_2^3$                       & $U(-1,1,50)$ & $\{x_1,x_2\}$\\
Livermore-13 & $x_1^{\frac{1}{3}}$                      & $U(0,4,20)$ & $\{x_1\}$\\
Livermore-14 & $x_1^3+x_1^2+x_1+\sin(x_1)+\sin(x_1^2)$    & $U(-1,1,20)$ & $\{x_1\}$\\
Livermore-15 & $x_1^{\frac{1}{5}}$                      & $U(0,4,20)$ & $\{x_1\}$\\
Livermore-16 & $x_1^{\frac{2}{5}}$                      & $U(0,4,20)$ & $\{x_1\}$\\
Livermore-17 & $4\sin(x_1)\cos(x_2)$                & $U(0,1,20)$ & $\{x_1,x_2\}$\\
Livermore-18 & $\sin(x_1^2)\cos(x_1)-5$              & $U(-1,1,20)$ & $\{x_1\}$\\
Livermore-19 & $x_1^5+x_1^4+x_1^2+x_1$                & $U(-1,1,20)$ & $\{x_1\}$\\
Livermore-20 & $\exp(-x_1^2)$                    & $U(-1,1,20)$ & $\{x_1\}$\\
Livermore-21 & $\scalebox{0.9}{$x_1^8+x_1^7+x_1^6+x_1^5+x_1^4+x_1^3+x_1^2+x_1$}$   & $U(-1,1,20)$ & $\{x_1\}$\\
Livermore-22 & $\exp(-\frac{1}{2}x_1^2)$                & $U(-1,1,20)$ & $\{x_1\}$\\
\midrule
Feynman-1          & $x_1 x_2$                  & $U(1,5,20)$ & $\{x_1,x_2\}$\\
Feynman-2          & $\frac{x_1}{2(1+x_2)}$         & $U(1,5,20)$ & $\{x_1,x_2\}$\\
Feynman-3          & $x_1 x_2^2$              & $U(1,5,20)$ & $\{x_1,x_2\}$\\
Feynman-4          & $1+\frac{x_1 x_2}{1-(x_1 x_2/3)}$             & $U(0,1,20)$ & $\{x_1,x_2\}$\\
Feynman-5          & $\frac{x_1}{x_2}$                  & $U(1,5,20)$ & $\{x_1,x_2\}$\\
Feynman-6          & $\frac{1}{2} x_1 x_2^2$           & $U(1,5,20)$ & $\{x_1,x_2\}$\\
Feynman-7          & $\frac{3}{2} x_1 x_2$              & $U(1,5,20)$ & $\{x_1,x_2\}$\\
Feynman-8          & $\frac{x_1}{e^{\frac{x_4 x_5}{x_2 x_3}}+e^{-\frac{x_4 x_5}{x_2 x_3}}}$          & $U(1,3,50)$ & $\{x_1,x_2,x_3,x_4,x_5\}$\\
Feynman-9          & $x_1 x_2 x_3 \log\frac{x_5}{x_4}$     & $U(1,5,50)$ & $\{x_1,x_2,x_3,x_4,x_5\}$\\
Feynman-10          & $x_1 (x_3-x_2) \frac{x_4}{x_5}$        & $U(1,5,50)$ & $\{x_1,x_2,x_3,x_4,x_5\}$\\
Feynman-11          & $\frac{x_1 x_2}{x_5 (x_3^2-x_4^2)}$ & $U(1,3,50)$ & $\{x_1,x_2,x_3,x_4,x_5\}$\\
Feynman-12          & $\frac{x_1 x_2^2 x_3}{3 x_4 x_5}$   & $U(1,5,50)$ & $\{x_1,x_2,x_3,x_4,x_5\}$\\
Feynman-13          & $x_1 (e^{\frac{x_2 x_3}{x_4 x_5}}-1)$  & $U(1,5,50)$ & $\{x_1,x_2,x_3,x_4,x_5\}$\\
Feynman-14          & $x_5 x_1 x_2 (\frac{1}{x_4}-\frac{1}{x_3})$    & $U(1,5,50)$ & $\{x_1,x_2,x_3,x_4,x_5\}$\\
Feynman-15          & $x_1 (x_2+x_3 x_4 \sin(x_5))$   & $U(1,5,50)$ & $\{x_1,x_2,x_3,x_4,x_5\}$\\
\bottomrule

\end{tabular}}
\label{table:bm_2}
\end{table}

\subsection{Chaotic Dynamics Dataset}


We use a nonlinear chaotic dynamics dataset called dysts \cite{gilpin2021chaos_dysts}. Considering the widespread presence of chaotic dynamics in the real world, such as in finance and meteorology, we believe that this dataset act as a robust benchmark for evaluating SR algorithms' ability to uncover underlying physical laws. Using this dataset avoids solely focusing on toy examples like Lorenz Attrator \cite{lorenz1963deterministic}, enabling a more objective and general assessment of SR algorithms' performance. By doing so, we can better understand the algorithm's capabilities in uncovering underlying dynamical equations.

We conducted tests using chaotic dynamics 
NewtonLiepnik \cite{leipnik1981double},
HyperLorenz \cite{gilpin2021chaos_dysts},
HyperJha \cite{gilpin2021chaos_dysts},
HyperPang \cite{pang2011new},
ShimizuMorioka \cite{shimizu1980bifurcation},
GenesioTesi \cite{Roberto1992GenesioTesi},
Laser \cite{abooee2013analysis_Laser},
Duffing \cite{duffing1918erzwungene},
Brusselator \cite{prigogine1980being_brusselator},
KawczynskiStrizhak \cite{strizhak1995complex_Kawczynski},
Rucklidge \cite{rucklidge1992chaos},
FitzHughNagumo \cite{fitzhugh1961impulses},
Finance \cite{cai2007new_Finance},
DequanLi \cite{li2008three_Dequanli},
Hadley \cite{Hadley} and SprottJerk \cite{sprott1997simplest}. Tables \ref{table:cs_1to4}, \ref{table:cs_5to8}, \ref{table:cs_9to12} and \ref{table:cs_13to16} show the governing equations of these chaotic dynamics. For systems with additional external force, we set the coefficients of their external force terms to 0 (i.e., autonomous chaotic dynamics). All other parameters were kept at their default values from the dysts \kredit{problem set}. The trajectory data is generated by solving numerical equations, and during this process, we sampled 1000 points, with every 100 points representing one period, and introduced 1\% Gaussian noise.

In this experiment, a large operator set $\{+, \times, -, \div, \sin, \cos, \exp, \log, \mathrm{abs}, \mathrm{sign}, \mathrm{tanh}, \mathrm{cosh}\}$, are used to simulate the situation in which scientists explore unknown systems in the real world. The models being compared were allocated a time budget of approximately 90 seconds. After the algorithms completed their search, we assessed whether the Pareto front produced by each algorithm contained expressions with the same structure (allowing for a constant bias term) as the ground truth expression. We calculated the average recovery rate of the derivatives (e.g., $\dot{x}$, $\dot{y}$, $\dot{z}$, which were determined using data affected by noise.) based on 50 repeated experiments, and then selected the minimum recovery rate among these derivatives to represent the overall recovery rate of the whole system.

\begin{figure}[htbp]
  \hspace*{-0.0in}
  \centering
  \includegraphics[width=6.5in]{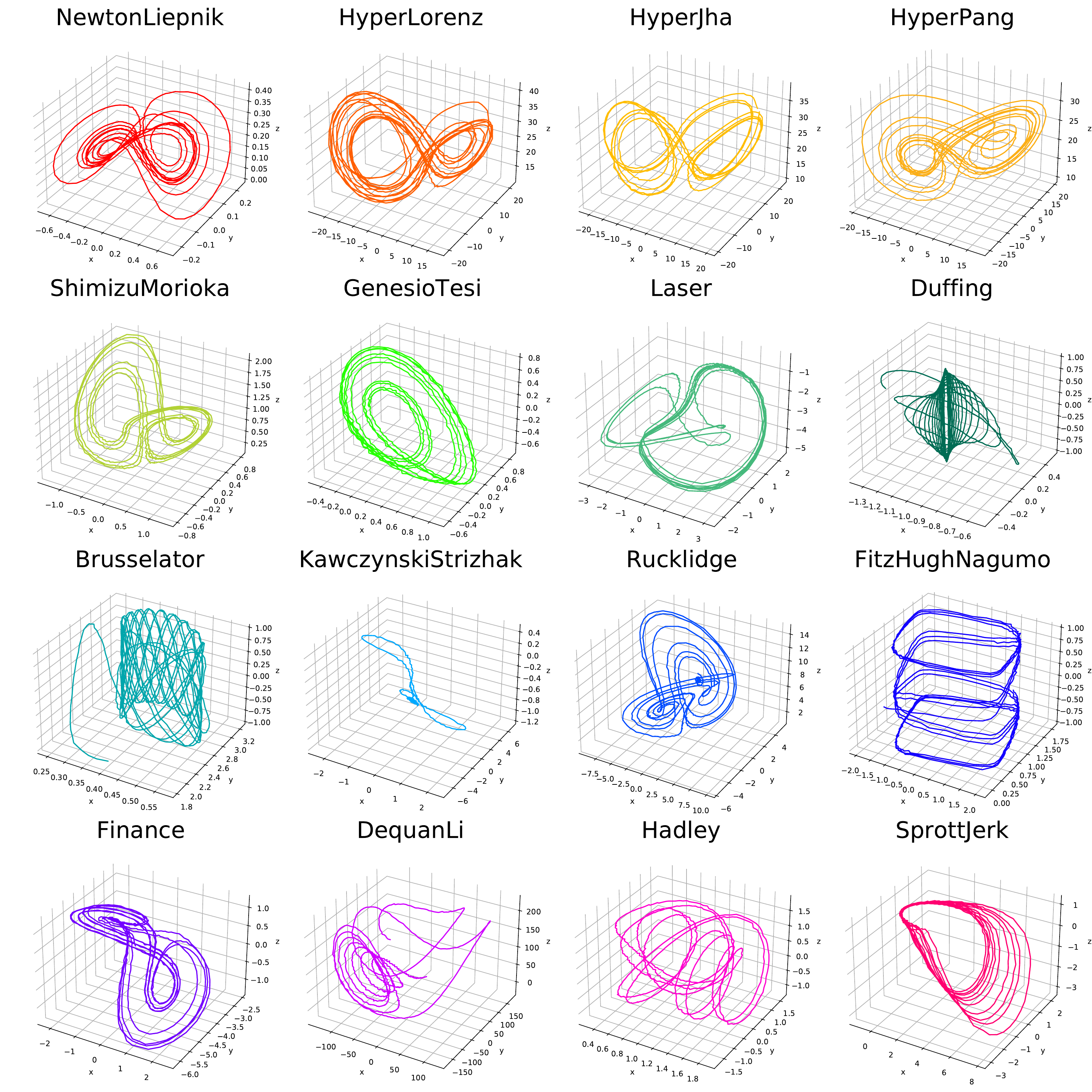}
  
  \caption{Visualization of 16 different nonlinear chaotic systems, each with 1\% Gaussian noise added. For systems with 4 dimensions, only the first 3 dimensions are shown.}
  \label{fig_16chaoticsys}
\end{figure}

\begin{table}[b!]
\centering
\caption{Chaotic dynamics 1 to 4.}
\begin{tabular}{lll}

\toprule
\textbf{Chaotic Dynamic} & \textbf{Governing Equations} & \textbf{Parameters} \\

\midrule
  NewtonLiepnik & $\begin{aligned}
    \dot{x} &= -ax + y +10yz \\
    \dot{y} &= -x-0.4y+5xz\\
    \dot{z} &= bz-5xy
           \end{aligned}$ & $a: 0.4, b: 0.175$ \\
\midrule
  HyperLorenz & $\begin{aligned}
    \dot{x} &= a (y - x) + w \\
    \dot{y} &= -x z + c x - y\\
    \dot{z} &= -b z + x y \\
    \dot{w} &= d w - x z
           \end{aligned}$ & $a: 10, b: 2.667, c: 28, d: 1.1$\\

\midrule
  HyperJha & $\begin{aligned}
    \dot{x} &= a (y - x) + w \\
    \dot{y} &= -x z + b x - y\\
    \dot{z} &= x y - c z \\
    \dot{w} &=  -x z + d w
           \end{aligned}$ & $a: 10, b: 28, c: 2.667, d: 1.3$\\

\midrule
  HyperPang & $\begin{aligned}
    \dot{x} &= a (y - x) \\
    \dot{y} &= -x z + c y + w\\
    \dot{z} &= x y - b z \\
    \dot{w} &=-d (x + y)
           \end{aligned}$ & $a: 36, b: 3, c: 20, d: 2$\\

\bottomrule
\end{tabular}
\label{table:cs_1to4}
\end{table}
\begin{table}[htbp]

\centering
\caption{Chaotic dynamics 5 to 8.}
\begin{tabular}{lll}

\toprule
\textbf{Chaotic Dynamic} & \textbf{Governing Equations} & \textbf{Parameters} \\

\midrule
  ShimizuMorioka & $\begin{aligned}
    \dot{x} &= y \\
    \dot{y} &=  x - a y - x z\\
    \dot{z} &= -b z + x ^ 2
           \end{aligned}$ & $a: 0.85, b: 0.5$ \\
\midrule
  GenesioTesi & $\begin{aligned}
    \dot{x} &= y \\
    \dot{y} &= z\\
    \dot{z} &= -c x - b y - a z + x ^ 2
           \end{aligned}$ & $a: 0.44, b: 1.1, c: 1$\\

\midrule
  Laser & $\begin{aligned}
    \dot{x} &= a (y - x) + b y z ^ 2 \\
    \dot{y} &= c x + d x z ^ 2\\
    \dot{z} &= h z + k x ^ 2
           \end{aligned}$ & $\begin{aligned}
               &a: 10.0, b: 1.0, c: 5.0,\\ &d: -1.0, h: -5.0, k: -6.0
           \end{aligned} $\\

\midrule
  Duffing & $\begin{aligned}
    \dot{x} &= y \\
    \dot{y} &= -\delta y - \beta x - \alpha x ^ 3\\
    \dot{z} &= \omega
           \end{aligned}$ & $\alpha: 1.0, \beta: -1.0, \delta: 0.1,  \omega: 1.4$\\

\bottomrule
\end{tabular}
\label{table:cs_5to8}
\end{table}
\begin{table}[htbp]

\centering
\caption{Chaotic dynamics 9 to 12.}
\begin{tabular}{lll}

\toprule
\textbf{Chaotic Dynamic} & \textbf{Governing Equations} & \textbf{Parameters} \\

\midrule
  Brusselator & $\begin{aligned}
    \dot{x} &= a + x ^ 2 y - (b + 1) x \\
    \dot{y} &= b x - x ^ 2 y\\
    \dot{z} &= w
           \end{aligned}$ & $a: 0.4, b: 1.2, w: 0.81$ \\
\midrule
  KawczynskiStrizhak & $\begin{aligned}
    \dot{x} &= \gamma (y - x ^ 3 + 3 \mu x) \\
    \dot{y} &= -2 \mu x - y - z + \beta\\
    \dot{z} &= \kappa (x - z)
           \end{aligned}$ & $\beta: -0.4, \gamma: 0.49, \kappa: 0.2, \mu: 2.1$\\

\midrule
  Rucklidge & $\begin{aligned}
    \dot{x} &= -a x + b y - y z \\
    \dot{y} &= x\\
    \dot{z} &= -z + y ^ 2
           \end{aligned}$ & $a: 2.0, b: 6.7$\\

\midrule
  FitzHughNagumo & $\begin{aligned}
    \dot{x} &= x - x ^ 3 / 3 - y + curr \\
    \dot{y} &= \gamma (x + a - b y)\\
    \dot{z} &= \omega
           \end{aligned}$ & $\begin{aligned}&a: 0.7, b: 0.8, curr: 0.965,\\& \gamma: 0.08, \omega: 0.04365\end{aligned}$\\

\bottomrule
\end{tabular}
\label{table:cs_9to12}
\end{table}
\begin{table}[htbp]

\centering
\caption{Chaotic dynamics 13 to 16.}
\begin{tabular}{lll}

\toprule
\textbf{Chaotic Dynamic} & \textbf{Governing Equations} & \textbf{Parameters} \\

\midrule
  Finance & $\begin{aligned}
    \dot{x} &= (1 / b - a) x + z + x y \\
    \dot{y} &= -b y - x ^ 2\\
    \dot{z} &= -x - c z
           \end{aligned}$ & $a: 0.001, b: 0.2, c: 1.1$ \\
\midrule
  DequanLi & $\begin{aligned}
    \dot{x} &= a (y - x) + d x z \\
    \dot{y} &= k x + f y - x z\\
    \dot{z} &= c z + x y - \epsilon x ^ 2
           \end{aligned}$ & $a: 40, c: 1.833, d: 0.16, \epsilon: 0.65, f: 20, k: 55$\\

\midrule
  Hadley & $\begin{aligned}
    \dot{x} &= -y ^ 2 - z ^ 2 - a x + a f \\
    \dot{y} &= x y - b x z - y + g\\
    \dot{z} &= b x y + x z - z
           \end{aligned}$ & $a: 0.2, b: 4, f: 9, g: 1$\\

\midrule
  SprottJerk & $\begin{aligned}
    \dot{x} &= y \\
    \dot{y} &= z\\
    \dot{z} &= -x + y ^ 2 - \mu z
           \end{aligned}$ & $\mu: 2.017$\\

\bottomrule
\end{tabular}
\label{table:cs_13to16}
\end{table}

\subsection{Configurations of Models}

\paragraph{PSE:} \kredit{In our experiment, we use GP and MCTS as our token generator. For GP, we use a tournament size of 10, maximum tree height of 10, maximum constant value of 5, crossover probability of 0.1, mutation probability of 0.5, population size of 50, number of generations of 20, hall of fame size of 20, and a token discount factor of 0.99. For MCTS, we set the number of constant candidates to 2, the number of attempts for each constant to 3, and uses a sampling interval of 0.1 for token constants.} The \textcolor{black}{PCTS} model for the symbolic regression benchmark employs operators $\{$$+$, $\times$, $-$, $\div$, $\mathrm{identity}$, $\sin$, $\cos$, $\exp$, $\log$$\}$. It has a structure of 3 layers, accepts 7 input tokens, and performs random down-sampling when the number of samples exceeds 20. It retrieves the top 10 expressions with minimum error in each PSRN forward pass. For chaotic dynamics, the operators expand to include $\{+,\times, -,\div, \sin, \cos, \exp, \log ,\tanh, \cosh, \mathrm{sign}, \mathrm{abs}\}$, and down-sampling is set to 100. For EMPS experiment, the operators used are the same as those used in chaotic dynamics experiments, and down-sampling is set to 200. For the turbulence experiment, the operators used are $\{+,\times, -,\div, \sin, \cos, \exp, \log, \tanh, \cosh,$$ \mathrm{pow2}, \mathrm{pow3}\}$, and down-sampling is set to 100. In the experiments described, when there is not enough graphic memory, $-$ and $\div$ operators are replaced with SemiSub and SemiDiv to reduce the graphic memory needed.

\paragraph{DGSR:} DGSR is designed for symbolic regression with a two-step process \cite{holt2023deep_DGSR}. In the first step, it pre-trains an encoder-decoder model on a vast set of mathematical expressions, aiming to capture the posterior distribution. The second step involves refining this distribution during inference to generate symbolic expressions. The framework incorporates a permutation-invariant encoding mechanism and an autoregressive decoding process, leveraging transformer architecture to handle the complexities of SR. We employed the code from \url{https://github.com/samholt/DeepGenerativeSymbolicRegression}, specifically the version as of September 12, 2023, with default hyperparameters. Since the authors did not provide a pre-trained model for searching expressions containing coefficients, we retrained a constant expression search model with hardcoded constants of 1, 5, and 10, using the operator library $\mathcal{O}_{\mathrm{Koza}}$, following the settings in the repository's codebase. All other hyperparameters remain at their default values. The operators used are the same as those in PSE.

\paragraph{NGGP:} The NGGP \cite{mundhenk2021symbolic_NGGP} model proposes a novel methodology in the realm of symbolic regression by integrating neural-guided search with genetic programming. This hybrid technique leverages a neural network to seed the initial population for the genetic algorithm, aiming to capitalize on the neural network's pattern recognition capabilities to enhance the search process. The model operates by cycling between generating candidate expressions through a neural sequence generator and refining these candidates using a stateless genetic programming component. We utilized the code from \url{https://github.com/dso-org/deep-symbolic-optimization}. For symbolic regression benchmark tasks, we employed default hyperparameters, and use operator library $\mathcal{O}_{\mathrm{Koza}}$. In physics data tasks with a specified time budget, we used the same operator sets as PSE and 5 GP iterations, 20000 training samples in each training epoch to avoid excessively long training sessions, and batch size is selected to be 50. The operators used are the same as those in PSE.

\paragraph{PySR:} PySR \cite{cranmer2023interpretable_PySR} is an open-source symbolic regression library that employs a multi-population genetic algorithm for equation discovery. It features an evolutionary approach through an evolve-simplify-optimize loop, focusing on the enhancement of constant optimization in empirical expressions. PySR utilizes tournament selection and incorporates elements of simulated annealing within its algorithmic structure, aiming to improve the search process. The library operates on a highly optimized Julia backend, SymbolicRegression.jl, which is integral to its performance and efficiency in handling SR tasks. We use the code available at \url{https://github.com/MilesCranmer/PySR}. For different experiments,  we used the same operator sets as PSE and restrict the approximate time budget by setting the number of iterations. In benchmark symbolic regression problems, the number of iterations is selected to be 10000. In chaotic dynamics experiments and EMPS experiment, the number of iterations is selected to be 30. In turbulence experiment, the number of iterations is selected to be 135. The operators used are the same as those in PSE.

\paragraph{BMS:} The BMS model \cite{guimera2020bayesian} introduces a novel Bayesian approach to automated model discovery, aiming to find interpretable mathematical expressions from data. It employs a Markov chain Monte Carlo (MCMC) algorithm to navigate the space of possible models, effectively sampling from the posterior distribution over mathematical expressions. BMS uses a maximum entropy principle to learn a prior over expressions from a large empirical corpus, typically compiled from scientific literature, such as Wikipedia. We use the code available at \url{https://bitbucket.org/rguimera/machine-scientist/}. The prior file used in turbulent experiments is \verb|nv1.np5.2017-10-18 18_07_| \verb|35.227360.dat| given in the repository. For EMPS, we used prior file with \verb|nv3.np3|. For chaotic dynamics, we utilized \verb|nv3.np3|, \verb|nv4.np8| and \verb|nv5.np7|  corresponding to the number of variables involved. The model parameters \verb|nsample|, \verb|thin|, and \verb|burnin| are selected to be 100, 10, and 500 respectively for limiting the time. The operators used are the same as those in PSE.

\paragraph{SPL:} The SPL model \cite{sun2023symbolic_SPL} employs a Monte Carlo tree search (MCTS) agent for traversing expression trees in pursuit of the most suitable expressions that align with the fundamental physics. This model adeptly incorporates existing knowledge and specific constraints within its grammatical structure, showcasing a strong capability to identify intricate expressions effectively. We utilized the code from \url{https://github.com/isds-neu/SymbolicPhysicsLearner}. And we used the result from Sun \textit{et al.} \cite{sun2023symbolic_SPL} and Xu \textit{et al.} \cite{xu2024reinforcement_RSRM}.

\kredit{

\paragraph{wAIC:} As a comparative baseline, we consider the wAIC method \cite{gao2022autonomous_NCS}. To comprehensively evaluate wAIC's performance, we examine two basis function settings: (1) wAIC1 uses polynomial basis functions, including constant, linear, quadratic, and cubic terms; (2) wAIC2 extends wAIC1 by adding the same unary operator basis functions used by other algorithms, such as sine, cosine, and exponential functions. These two settings represent simple and complex basis function spaces, respectively, allowing us to test wAIC's performance under different basis function configurations. In our experiments, we employ the commonly used hyperparameters lambda $= [0.1, 0.1, 0.1]$ from the codebase. This method demonstrates the ability to handle complex system modeling while maintaining computational efficiency, particularly in its performance on short time series data. We utilized the code from \url{https://codeocean.com/capsule/0776097/tree/v1}.
}

\kredit{
\paragraph{uDSR:} Unified Deep Symbolic Regression (uDSR) \cite{uDSR_landajuela2022unified} is a comprehensive framework that integrates five distinct symbolic regression strategies: recursive problem simplification, neural-guided search, large-scale pre-training, genetic programming, and linear models. Built upon the Deep Symbolic Regression (DSR) algorithm, uDSR employs a neural network controller to learn a distribution over mathematical expressions. It incorporates AI Feynman's recursive problem simplification \cite{udrescu2020ai_AIFeynman} as an algorithmic wrapper, a set transformer encoder-decoder architecture for large-scale pre-training, genetic programming as an inner optimization loop, and a novel LINEAR token to abstract linear models. The framework uses both supervised and reinforcement learning objectives for pre-training and employs various in situ priors and constraints to bias and prune the search space. We utilized the code from \url{https://github.com/dso-org/deep-symbolic-optimization}}

\kredit{
\paragraph{TPSR:} Transformer-based Planning for Symbolic Regression (TPSR)  \cite{Shojaee_reviewer_NIPS2023_TPSR} combines pre-trained transformer models with MCTS for symbolic regression. TPSR uses MCTS as a decoding strategy to guide equation sequence generation, incorporating equation verification reward such as fitting accuracy and complexity. In our experiments, we set $\lambda=0.1$, \texttt{horizon }$=200$, \texttt{width }$=5$, \texttt{num\_beams }$=2$, \texttt{rollout }$=5$, and \texttt{max\_number\_bags }$=10$. The framework includes a reward function balancing equation fitting accuracy and complexity. TPSR integrates next-token prediction probabilities from pre-trained transformer SR models into the MCTS planning process. This approach aims to optimize the generation of equation sequences while considering non-differentiable performance reward, potentially improving the balance between fitting accuracy and model complexity in symbolic regression tasks.
In the EMPS dataset, we reflected the token outside the original Koza set and $\mathrm{sign}$, modifying the token of the E2E model to identity transformation. We then fine-tuned the model using the training module of the E2E model and generated equations using the default configuration.
For the chaotic dataset, we retained the original model without modifications and generated equations following the default TPSR model settings, but we adjusted the width and rollout parameters to 5.
In the roughpipe dataset, we reduced tokens outside the Koza set structure, added additional tokens ($\cosh$ and $\tanh$), and retrained the model until the loss converged. We then proceeded to generate equations using the default configuration. We utilized
 the code from \url{https://github.com/deep-symbolic-mathematics/TPSR}}

\paragraph{Operon:} The Operon framework \cite{Operon} is an efficient C++ genetic programming library focused on performance, modularity, and usability, which we accessed through its Python wrapper, PyOperon. It introduces an efficient linear tree encoding and a scalable concurrency model, where each logical thread is responsible for generating a new individual. The evolutionary process is modeled around the concept of an ``offspring generator'', a streaming operator that defines how new individuals are obtained, allowing different algorithmic variants to be expressed within the same basic generational loop. A notable feature of Operon is its native support for automatic differentiation, which enables the calculation of model derivatives and allows for seamless integration with gradient-based local search approaches for parameter optimization. We utilized the code from \url{https://github.com/heal-research/pyoperon}. For our experiments, we used the default hyperparameters provided by the library and set a consistent time limit to ensure a fair comparison with other methods.

\paragraph{ESR:} Exhaustive Symbolic Regression (ESR) \cite{ESR_Bartlett_2022} is a method that systematically considers all possible equations up to a specified maximum complexity from a given basis set of operators. By explicitly evaluating every possible combination, ESR is guaranteed to find the true Pareto front of optimal solutions, provided that parameters are perfectly optimized. A key aspect of ESR is its use of the Minimum Description Length (MDL) principle to provide a rigorous and objective method for model selection, combining accuracy and complexity into a single objective. ESR is limited to discovering functions of a single variable. The fundamental difference between ESR and our approach is that ESR performs an explicit enumeration—constructing every possible expression as a symbolic object and evaluating them one by one, which is computationally intensive. In contrast, our PSRN performs an implicit enumeration, leveraging subtree reuse to first identify the expression(s) with the minimum loss before recursively deriving their symbolic forms. We utilized the code from \url{https://github.com/DeaglanBartlett/ESR}. For our experiments, we used the default hyperparameters and set a consistent time limit for a fair comparison.

\subsection{Evaluation Metrics}
Here, we will introduce the definitions of some key evaluation metrics:

\textbf{Complexity}: The number of operators in an expression. A higher complexity often suggests a greater risk of overfitting when the expression is used in a predictive model. This concept is consistent with the principle of Occam's razor, which posits that, all else being equal, simpler explanations are generally preferable to more complex ones.

\textbf{Recovery Rate}: The proportion of instances where SR algorithms identify symbolically equivalent expressions compared to the ground truth, measured across numerous independent trials with different random seeds. In the symbolic regression benchmark test, the found expressions must be symbolically equivalent to the ground truth expression (validated using SymPy \cite{sympy}). For symbolic regression benchmark expressions containing constants, the constants are trimmed to two decimal places before comparison. In chaotic dynamic discovery task, due to the presence of Gaussian noise, an expression is considered successfully recovered if it has the same structural form as the ground truth expression (allowing for a constant bias term).

\subsection{Hardware and Environment}
The experiments were performed on servers with Nvidia A100 (80GB) and Intel(R) Xeon(R) Platinum 8380 cpus @ 2.30GHz.
Note that for environments running the PSE model, a minimum Pytorch version of 2.0.0 is required. Versions below this threshold may be incompatible with the \verb|torch.topk| operation and may result in computational issues.

\clearpage
\section{Results}
The following table shows the detailed recovery rates and time cost on the symbolic regression \kredit{benchmark problem sets} and 16 chaotic dynamics.

\subsection{Symbolic Regression Benchmark Problem}

\hsedit{Table \ref{tab:recovery_performance} shows the overall recovery rates for different methods, while Table \ref{tab:time_performance} summarizes the corresponding time performance.} Tables \ref{table:sr_Nguyen}, \ref{table:sr_Nguyen-c}, \ref{table:sr_R}, \ref{table:sr_Rast}, \ref{table:sr_Livermore}, and \ref{table:sr_Feynman} show the recovery rate and time cost on each symbolic regression \kredit{benchmark problem set}.

\begin{table}[htbp]
\centering
\caption{Recovery rates of different methods for six benchmark problem sets \kredit{(with 95\% confidence intervals)}.}
\label{tab:recovery_performance}
\resizebox{\textwidth}{!}{\begin{tabular}{lccccccccc}
\toprule
Dataset & PSE & SPL & NGGP & DGSR & PySR & BMS & uDSR & TPSR & Operon \\
\midrule
Feynman & \textbf{94.00 ± 1.18} & 20.00 ± 2.10 & 71.87 ± 2.10 & 77.00 ± 2.02 & 53.57 ± 1.88 & 51.67 ± 2.16 & 37.33 ± 2.38 & 34.00 ± 2.07 & 2.67 ± 0.33 \\
Livermore & \textbf{100.00 ± 0.00} & 51.00 ± 1.69 & 77.86 ± 1.45 & 62.08 ± 1.80 & 48.07 ± 1.78 & 5.30 ± 0.37 & 29.32 ± 1.68 & 8.51 ± 0.98 & 10.23 ± 1.15 \\
Nguyen & \textbf{100.00 ± 0.00} & 93.50 ± 1.17 & 91.50 ± 1.63 & 69.70 ± 2.54 & 87.41 ± 1.73 & 7.82 ± 0.72 & 77.83 ± 1.93 & 18.72 ± 1.87 & 4.17 ± 0.66 \\
Nguyen-c & \textbf{100.00 ± 0.00} & 81.17 ± 3.19 & 60.00 ± 2.59 & 33.99 ± 2.62 & 69.17 ± 2.99 & 15.28 ± 1.93 & 72.50 ± 3.53 & 17.72 ± 2.23 & 50.83 ± 4.32 \\
R & \textbf{100.00 ± 0.00} & 0.00 ± 0.00 & 2.33 ± 0.46 & 0.00 ± 0.00 & 0.00 ± 0.00 & 0.34 ± 0.07 & 0.00 ± 0.00 & 0.00 ± 0.00 & 0.00 ± 0.00 \\
R$^*$ & \textbf{100.00 ± 0.00} & 0.00 ± 0.00 & 53.67 ± 4.14 & 19.30 ± 3.80 & 0.00 ± 0.00 & 0.69 ± 0.07 & 0.00 ± 0.00 & 0.00 ± 0.00 & 0.00 ± 0.00 \\
\midrule
Average 95\% CI & {\bm{$\pm0.2$}} & {\bm{$\pm1.4$}} & {\bm{$\pm2.1$}} & {\bm{$\pm2.1$}} & {\bm{$\pm1.4$}} & {\bm{$\pm0.9$}} & {\bm{$\pm1.6$}} & {\bm{$\pm1.2$}} & {\bm{$\pm1.1$}} \\
\bottomrule
\end{tabular}}
\end{table}


\begin{table}[htbp]
\centering
\caption{Ratio of $R^2>0.99$ of different methods for six benchmark problem sets \kredit{(with 95\% confidence intervals)}.}
\label{tab:r2_gt_099_ratio}
\resizebox{\textwidth}{!}{
\kreditNCS{
\begin{tabular}{lccccccccc}

\toprule
Dataset & PSE & SPL & NGGP & DGSR & PySR & BMS & uDSR & TPSR & Operon \\
\midrule
Feynman & \textbf{0.97 ± 0.07} & 0.40 ± 0.06 & 0.90 ± 0.02 & 0.89 ± 0.08 & 0.86 ± 0.03 & 0.90 ± 0.01 & 0.55 ± 0.07 & 0.96 ± 0.03 & 0.35 ± 0.02 \\
Livermore & \textbf{1.00 ± 0.00} & 0.67 ± 0.03 & \textbf{1.00 ± 0.00} & 0.84 ± 0.08 & 0.92 ± 0.01 & 0.45 ± 0.01 & 0.62 ± 0.10 & 0.88 ± 0.02 & 0.52 ± 0.06 \\
Nguyen & \textbf{1.00 ± 0.00} & 0.96 ± 0.02 & 0.93 ± 0.02 & 0.75 ± 0.10 & \textbf{1.00 ± 0.00} & 0.61 ± 0.03 & 0.84 ± 0.13 & 0.88 ± 0.03 & 0.09 ± 0.03 \\
Nguyen-c & \textbf{1.00 ± 0.00} & 0.91 ± 0.07 & \textbf{1.00 ± 0.00}  & 0.86 ± 0.03 & 0.88 ± 0.05 & 0.70 ± 0.02 & 0.94 ± 0.08 & \textbf{1.00 ± 0.00} & 0.97 ± 0.03 \\
R & \textbf{1.00 ± 0.00} & 0.67 ± 0.12 & 0.82 ± 0.04 & \textbf{1.00 ± 0.00} & \textbf{1.00 ± 0.00} & 0.17 ± 0.05 & 0.08 ± 0.16 & 0.82 ± 0.07 & 0.00 ± 0.00 \\
R$^*$ & \textbf{1.00 ± 0.00} & 0.96 ± 0.07 & 0.63 ± 0.04 & 0.57 ± 0.26 & 0.41 ± 0.08 & 0.76 ± 0.03 & 0.67 ± 0.27 & 0.85 ± 0.06 & 0.00 ± 0.00 \\
\midrule
Average 95\% CI & {\bm{$\pm0.01$}} & {\bm{$\pm0.06$}} & {\bm{$\pm0.02$}} & {\bm{$\pm0.09$}} & {\bm{$\pm0.03$}} & {\bm{$\pm0.02$}} & {\bm{$\pm0.14$}} & {\bm{$\pm0.04$}} & {\bm{$\pm0.01$}} \\
\bottomrule
\end{tabular}
}
}
\end{table}


\begin{table}[htbp]
\centering
\caption{Time performance of different methods for six benchmark problem sets \kredit{(with 95\% confidence intervals)}.}
\label{tab:time_performance}
\resizebox{\textwidth}{!}{\begin{tabular}{lccccccccc}
\toprule
Dataset & PSE & SPL & NGGP & DGSR & PySR & BMS & uDSR & TPSR & Operon \\
\midrule
Feynman & 525.4 ± 86.1 & 962.7 ± 23.3 & 320.8 ± 17.9 & \textbf{262.3 ± 17.0} & 3094.2 ± 163.0 & 2475.8 ± 97.7 & 1272.8 ± 35.8 & 273.6 ± 12.3 & 3600.3 ± 0.0 \\
Livermore & \textbf{48.7 ± 5.1} & 1658.4 ± 36.9 & 264.4 ± 10.5 & 428.5 ± 19.0 & 1108.7 ± 68.7 & 3113.7 ± 38.0 & 1541.7 ± 18.5 & 185.7 ± 7.5 & 3600.2 ± 0.0 \\
Nguyen & \textbf{11.0 ± 1.4} & 119.8 ± 10.6 & 130.8 ± 13.3 & 269.7 ± 24.6 & 677.9 ± 112.0 & 3442.1 ± 48.5 & 1059.5 ± 35.0 & 164.5 ± 5.1 & 3600.2 ± 0.0 \\
Nguyen-c & 246.1 ± 22.9 & 1446.5 ± 113.3 & 6489.7 ± 631.4 & 3181.1 ± 246.1 & 1319.8 ± 134.8 & 3066.0 ± 69.7 & 1487.1 ± 115.2 & \textbf{138.8 ± 6.2} & 3600.6 ± 0.0 \\
R & \textbf{333.1 ± 42.8} & 1393.4 ± 13.2 & 767.6 ± 5.6 & 1129.0 ± 3.7 & 2180.4 ± 4.8 & 3826.9 ± 106.7 & 1829.2 ± 3.6 & 365.6 ± 20.2 & 3600.2 ± 0.0 \\
R$^*$ & \textbf{35.0 ± 1.2} & 1364.6 ± 15.5 & 616.3 ± 42.0 & 846.0 ± 39.6 & 1925.9 ± 26.3 & 3381.3 ± 49.0 & 1841.1 ± 5.3 & 283.6 ± 21.2 & 3600.1 ± 0.0 \\
\midrule
Average 95\% CI & {\bm{$\pm26.6$}} & {\bm{$\pm35.4$}} & {\bm{$\pm120.1$}} & {\bm{$\pm58.3$}} & {\bm{$\pm84.9$}} & {\bm{$\pm68.3$}} & {\bm{$\pm35.6$}} & {\bm{$\pm12.1$}} & {\bm{$\pm0.0$}} \\
\bottomrule
\end{tabular}}
\end{table}

\begin{table}[htbp]
\centering
\caption{Symbolic regression results on Nguyen dataset.}
\resizebox{0.99\textwidth}{!}{
\begin{tabular}{cHcccccccccccccccccc}
\toprule
{\textbf{Benchmark}} & {\textbf{Expression}} & \multicolumn{9}{c}{\textbf{Recovery rate}} & \multicolumn{9}{c}{\textbf{Time cost (s)}} \\
{} & {} & {PSE} & {SPL} & {NGGP} & {DGSR} & {PySR} & {BMS} & {uDSR} & {TPSR} & {Operon} & {PSE} & {SPL} & {NGGP} & {DGSR} & {PySR} & {BMS} & {uDSR} & {TPSR} & {Operon} \\
\midrule
Nguyen-1 & $x_{1}^{3} + x_{1}^{2} + x_{1}$ & \cellcolor[HTML]{d3a9ce} \bfseries 100\% & \cellcolor[HTML]{d3a9ce} \bfseries 100\% & \cellcolor[HTML]{d3a9ce} \bfseries 100\% & \cellcolor[HTML]{d3a9ce} \bfseries 100\% & \cellcolor[HTML]{d3a9ce} \bfseries 100\% & 43\% & \cellcolor[HTML]{d3a9ce} \bfseries 100\% & 87\% & 0\% & \cellcolor[HTML]{9bcae9} \bfseries 1.26 & 8.78 & 13.45 & 10.01 & 2.10 & 2335.26 & 246.28 & 107.52 & 3600.15 \\
Nguyen-2 & $x_{1}^{4} + x_{1}^{3} + x_{1}^{2} + x_{1}$ & \cellcolor[HTML]{d3a9ce} \bfseries 100\% & \cellcolor[HTML]{d3a9ce} \bfseries 100\% & \cellcolor[HTML]{d3a9ce} \bfseries 100\% & \cellcolor[HTML]{d3a9ce} \bfseries 100\% & 99\% & 14\% & \cellcolor[HTML]{d3a9ce} \bfseries 100\% & 85\% & 0\% & \cellcolor[HTML]{9bcae9} \bfseries 1.32 & 7.30 & 18.20 & 8.25 & 2.17 & 3398.75 & 714.30 & 131.00 & 3600.17 \\
Nguyen-3 & $x_{1}^{5} + x_{1}^{4} + x_{1}^{3} + x_{1}^{2} + x_{1}$ & \cellcolor[HTML]{d3a9ce} \bfseries 100\% & \cellcolor[HTML]{d3a9ce} \bfseries 100\% & \cellcolor[HTML]{d3a9ce} \bfseries 100\% & \cellcolor[HTML]{d3a9ce} \bfseries 100\% & \cellcolor[HTML]{d3a9ce} \bfseries 100\% & 2\% & \cellcolor[HTML]{d3a9ce} \bfseries 100\% & 30\% & 0\% & \cellcolor[HTML]{9bcae9} \bfseries 2.93 & 81.29 & 34.86 & 8.45 & 4.15 & 3874.32 & 1142.35 & 237.70 & 3600.16 \\
Nguyen-4 & $x_{1}^{6} + x_{1}^{5} + x_{1}^{4} + x_{1}^{3} + x_{1}^{2} + x_{1}$ & \cellcolor[HTML]{d3a9ce} \bfseries 100\% & 99\% & \cellcolor[HTML]{d3a9ce} \bfseries 100\% & \cellcolor[HTML]{d3a9ce} \bfseries 100\% & 97\% & 3\% & \cellcolor[HTML]{d3a9ce} \bfseries 100\% & 22\% & 0\% & \cellcolor[HTML]{9bcae9} \bfseries 3.22 & 567.06 & 71.65 & 27.16 & 258.13 & 4257.95 & 1190.55 & 199.00 & 3600.14 \\
Nguyen-5 & $\sin{\left(x_{1}^{2} \right)} \cos{\left(x_{1} \right)} - 1$ & \cellcolor[HTML]{d3a9ce} \bfseries 100\% & 95\% & 99\% & 98\% & \cellcolor[HTML]{d3a9ce} \bfseries 100\% & 0\% & 48\% & 0\% & 0\% & \cellcolor[HTML]{9bcae9} \bfseries 3.25 & 431.23 & 175.45 & 69.51 & 577.04 & 2030.73 & 1927.26 & 61.56 & 3600.15 \\
Nguyen-6 & $\sin{\left(x_{1} \right)} + \sin{\left(x_{1}^{2} + x_{1} \right)}$ & \cellcolor[HTML]{d3a9ce} \bfseries 100\% & \cellcolor[HTML]{d3a9ce} \bfseries 100\% & 99\% & 0\% & \cellcolor[HTML]{d3a9ce} \bfseries 100\% & 0\% & \cellcolor[HTML]{d3a9ce} \bfseries 100\% & 0\% & 0\% & 2.78 & 64.65 & 122.97 & 7.29 & \cellcolor[HTML]{9bcae9} \bfseries 1.96 & 4319.68 & 401.27 & 111.03 & 3600.16 \\
Nguyen-7 & $\log{\left(x_{1} + 1 \right)} + \log{\left(x_{1}^{2} + 1 \right)}$ & \cellcolor[HTML]{d3a9ce} \bfseries 100\% & \cellcolor[HTML]{d3a9ce} \bfseries 100\% & \cellcolor[HTML]{d3a9ce} \bfseries 100\% & 15\% & \cellcolor[HTML]{d3a9ce} \bfseries 100\% & 0\% & 55\% & 0\% & 10\% & 17.97 & \cellcolor[HTML]{9bcae9} \bfseries 14.99 & 126.60 & 970.35 & 78.67 & 3708.15 & 1684.55 & 53.64 & 3600.16 \\
Nguyen-8 & $\sqrt{x_{1}}$ & \cellcolor[HTML]{d3a9ce} \bfseries 100\% & \cellcolor[HTML]{d3a9ce} \bfseries 100\% & \cellcolor[HTML]{d3a9ce} \bfseries 100\% & \cellcolor[HTML]{d3a9ce} \bfseries 100\% & 53\% & 6\% & 94\% & 0\% & 40\% & \cellcolor[HTML]{9bcae9} \bfseries 3.26 & 5.59 & 63.27 & 68.88 & 212.03 & 3245.07 & 999.42 & 43.06 & 3600.14 \\
Nguyen-9 & $\sin{\left(x_{1} \right)} + \sin{\left(x_{2}^{2} \right)}$ & \cellcolor[HTML]{d3a9ce} \bfseries 100\% & \cellcolor[HTML]{d3a9ce} \bfseries 100\% & \cellcolor[HTML]{d3a9ce} \bfseries 100\% & \cellcolor[HTML]{d3a9ce} \bfseries 100\% & \cellcolor[HTML]{d3a9ce} \bfseries 100\% & 0\% & \cellcolor[HTML]{d3a9ce} \bfseries 100\% & 0\% & 0\% & \cellcolor[HTML]{9bcae9} \bfseries 2.29 & 5.74 & 17.81 & 8.19 & 2.91 & 4887.11 & 624.95 & 285.22 & 3600.12 \\
Nguyen-10 & $2 \sin{\left(x_{1} \right)} \cos{\left(x_{2} \right)}$ & \cellcolor[HTML]{d3a9ce} \bfseries 100\% & \cellcolor[HTML]{d3a9ce} \bfseries 100\% & \cellcolor[HTML]{d3a9ce} \bfseries 100\% & \cellcolor[HTML]{d3a9ce} \bfseries 100\% & \cellcolor[HTML]{d3a9ce} \bfseries 100\% & 19\% & 37\% & 0\% & 0\% & \cellcolor[HTML]{9bcae9} \bfseries 2.37 & 53.24 & 45.06 & 79.70 & 4.43 & 3255.49 & 1547.74 & 242.87 & 3600.24 \\
Nguyen-11 & $x_{1}^{x_{2}}$ & \cellcolor[HTML]{d3a9ce} \bfseries 100\% & \cellcolor[HTML]{d3a9ce} \bfseries 100\% & \cellcolor[HTML]{d3a9ce} \bfseries 100\% & 23\% & \cellcolor[HTML]{d3a9ce} \bfseries 100\% & 5\% & \cellcolor[HTML]{d3a9ce} \bfseries 100\% & 0\% & 0\% & \cellcolor[HTML]{9bcae9} \bfseries 2.42 & 10.16 & 22.55 & 847.79 & 57.95 & 3537.27 & 282.90 & 220.37 & 3600.21 \\
Nguyen-12 & $x_{1}^{4} - x_{1}^{3} + \frac{1}{2} x_{2}^{2} - x_{2}$ & \cellcolor[HTML]{d3a9ce} \bfseries 100\% & 28\% & 0\% & 0\% & 0\% & 1\% & 0\% & 0\% & 0\% & \cellcolor[HTML]{9bcae9} \bfseries 88.98 & 187.90 & 857.18 & 1130.91 & 6933.52 & 2455.88 & 1952.29 & 280.90 & 3600.14 \\
\bottomrule
\end{tabular}}
\label{table:sr_Nguyen}
\end{table}

\begin{table}[htbp]
\centering
\caption{Symbolic regression results on Nguyen-c dataset.}
\resizebox{0.99\textwidth}{!}{
\begin{tabular}{cHcccccccccccccccccc}
\toprule
{\textbf{Benchmark}} & {\textbf{Expression}} & \multicolumn{9}{c}{\textbf{Recovery rate}} & \multicolumn{9}{c}{\textbf{Time cost (s)}} \\
{} & {} & {PSE} & {SPL} & {NGGP} & {DGSR} & {PySR} & {BMS} & {uDSR} & {TPSR} & {Operon} & {PSE} & {SPL} & {NGGP} & {DGSR} & {PySR} & {BMS} & {uDSR} & {TPSR} & {Operon} \\
\midrule
Nguyen-1c & $3.39 x_{1}^{3} + 2.12 x_{1}^{2} + 1.78 x_{1}$ & \cellcolor[HTML]{d3a9ce} \bfseries 100\% & \cellcolor[HTML]{d3a9ce} \bfseries 100\% & 76\% & 45\% & 65\% & 60\% & \cellcolor[HTML]{d3a9ce} \bfseries 100\% & 60\% & 0\% & \cellcolor[HTML]{9bcae9} \bfseries 25.62 & 452.73 & 804.64 & 181.72 & 666.11 & 2017.22 & 558.58 & 125.46 & 3600.23 \\
Nguyen-2c & \scalebox{0.8}{$0.48 x_{1}^{4} + 3.39 x_{1}^{3} + 2.12 x_{1}^{2} + 1.78 x_{1}$} & \cellcolor[HTML]{d3a9ce} \bfseries 100\% & 94\% & 16\% & 18\% & 0\% & 25\% & 0\% & 46\% & 5\% & \cellcolor[HTML]{9bcae9} \bfseries 51.32 & 295.77 & 1122.97 & 541.87 & 2047.44 & 3554.27 & 2208.38 & 107.75 & 3600.54 \\
Nguyen-5c & $\sin{\left(x_{1}^{2} \right)} \cos{\left(x_{1} \right)} - 0.75$ & \cellcolor[HTML]{d3a9ce} \bfseries 100\% & 95\% & 28\% & 41\% & 85\% & 1\% & \cellcolor[HTML]{d3a9ce} \bfseries 100\% & 0\% & 0\% & \cellcolor[HTML]{9bcae9} \bfseries 13.89 & 2178.89 & 8649.82 & 4158.69 & 671.19 & 1965.81 & 2317.40 & 99.68 & 3600.55 \\
Nguyen-8c & $\sqrt{1.23 x_{1}}$ & \cellcolor[HTML]{d3a9ce} \bfseries 100\% & \cellcolor[HTML]{d3a9ce} \bfseries 100\% & \cellcolor[HTML]{d3a9ce} \bfseries 100\% & 90\% & \cellcolor[HTML]{d3a9ce} \bfseries 100\% & 5\% & \cellcolor[HTML]{d3a9ce} \bfseries 100\% & 0\% & \cellcolor[HTML]{d3a9ce} \bfseries 100\% & 704.40 & 77.89 & 701.43 & 845.45 & 73.60 & 3435.69 & 172.30 & \cellcolor[HTML]{9bcae9} \bfseries 41.89 & 3600.58 \\
Nguyen-9c & $\sin{\left(1.5 x_{1} \right)} + \sin{\left(0.5 x_{2}^{2} \right)}$ & \cellcolor[HTML]{d3a9ce} \bfseries 100\% & 98\% & 60\% & 0\% & \cellcolor[HTML]{d3a9ce} \bfseries 100\% & 0\% & \cellcolor[HTML]{d3a9ce} \bfseries 100\% & 0\% & \cellcolor[HTML]{d3a9ce} \bfseries 100\% & 478.73 & 2001.40 & 21005.76 & 7043.42 & \cellcolor[HTML]{9bcae9} \bfseries 46.42 & 4075.21 & 60.40 & 254.11 & 3600.76 \\
Nguyen-10c & $\sin{\left(1.5 x_{1} \right)} \cos{\left(0.5 x_{2} \right)}$ & \cellcolor[HTML]{d3a9ce} \bfseries 100\% & 0\% & 80\% & 9\% & 65\% & 0\% & 35\% & 0\% & \cellcolor[HTML]{d3a9ce} \bfseries 100\% & \cellcolor[HTML]{9bcae9} \bfseries 202.55 & 3672.39 & 6653.39 & 6315.63 & 4413.90 & 3347.57 & 3605.62 & 203.67 & 3600.76 \\
\bottomrule
\end{tabular}}
\label{table:sr_Nguyen-c}
\end{table}

\begin{table}[htbp]
\centering
\caption{Symbolic regression results on R dataset.}
\resizebox{0.99\textwidth}{!}{
\begin{tabular}{cHcccccccccccccccccc}
\toprule
{\textbf{Benchmark}} & {\textbf{Expression}} & \multicolumn{9}{c}{\textbf{Recovery rate}} & \multicolumn{9}{c}{\textbf{Time cost (s)}} \\
{} & {} & {PSE} & {SPL} & {NGGP} & {DGSR} & {PySR} & {BMS} & {uDSR} & {TPSR} & {Operon} & {PSE} & {SPL} & {NGGP} & {DGSR} & {PySR} & {BMS} & {uDSR} & {TPSR} & {Operon} \\
\midrule
R-1 & $\frac{\left(x_{1} + 1\right)^{3}}{x_{1}^{2} - x_{1} + 1}$ & \cellcolor[HTML]{d3a9ce} \bfseries 100\% & 0\% & 0\% & 0\% & 0\% & 1\% & 0\% & 0\% & 0\% & \cellcolor[HTML]{9bcae9} \bfseries 133.00 & 1507.01 & 759.25 & 1148.08 & 2196.87 & 4754.01 & 1856.96 & 334.84 & 3600.15 \\
R-2 & $\frac{x_{1}^{5} - 3 x_{1}^{3} + 1}{x_{1}^{2} + 1}$ & \cellcolor[HTML]{d3a9ce} \bfseries 100\% & 0\% & 0\% & 0\% & 0\% & 0\% & 0\% & 0\% & 0\% & \cellcolor[HTML]{9bcae9} \bfseries 98.22 & 1397.70 & 722.63 & 1091.35 & 2212.13 & 2875.85 & 1794.24 & 205.12 & 3600.15 \\
R-3 & $\frac{x_{1}^{6} + x_{1}^{5}}{x_{1}^{4} + x_{1}^{3} + x_{1}^{2} + x_{1} + 1}$ & \cellcolor[HTML]{d3a9ce} \bfseries 100\% & 0\% & 7\% & 0\% & 0\% & 0\% & 0\% & 0\% & 0\% & 767.96 & 1275.47 & 820.86 & 1147.61 & 2132.06 & 3850.87 & 1836.52 & \cellcolor[HTML]{9bcae9} \bfseries 556.80 & 3600.15 \\
\bottomrule
\end{tabular}}
\label{table:sr_R}
\end{table}

\begin{table}[htbp]
\centering
\caption{Symbolic regression results on R* dataset.}
\resizebox{0.99\textwidth}{!}{
\begin{tabular}{cHcccccccccccccccccc}
\toprule
{\textbf{Benchmark}} & {\textbf{Expression}} & \multicolumn{9}{c}{\textbf{Recovery rate}} & \multicolumn{9}{c}{\textbf{Time cost (s)}} \\
{} & {} & {PSE} & {SPL} & {NGGP} & {DGSR} & {PySR} & {BMS} & {uDSR} & {TPSR} & {Operon} & {PSE} & {SPL} & {NGGP} & {DGSR} & {PySR} & {BMS} & {uDSR} & {TPSR} & {Operon} \\
\midrule
R*-1 & $\frac{\left(x_{1} + 1\right)^{3}}{x_{1}^{2} - x_{1} + 1}$ & \cellcolor[HTML]{d3a9ce} \bfseries 100\% & 0\% & 26\% & 0\% & 0\% & 1\% & 0\% & 0\% & 0\% & \cellcolor[HTML]{9bcae9} \bfseries 44.35 & 1296.93 & 880.24 & 1034.55 & 2178.47 & 3095.16 & 1860.95 & 487.96 & 3600.14 \\
R*-2 & $\frac{x_{1}^{5} - 3 x_{1}^{3} + 1}{x_{1}^{2} + 1}$ & \cellcolor[HTML]{d3a9ce} \bfseries 100\% & 0\% & 40\% & 0\% & 0\% & 1\% & 0\% & 0\% & 0\% & \cellcolor[HTML]{9bcae9} \bfseries 36.42 & 1521.33 & 775.00 & 1059.25 & 1724.86 & 3877.82 & 1874.95 & 123.66 & 3600.15 \\
R*-3 & $\frac{x_{1}^{6} + x_{1}^{5}}{x_{1}^{4} + x_{1}^{3} + x_{1}^{2} + x_{1} + 1}$ & \cellcolor[HTML]{d3a9ce} \bfseries 100\% & 0\% & 95\% & 58\% & 0\% & 0\% & 0\% & 0\% & 0\% & \cellcolor[HTML]{9bcae9} \bfseries 24.09 & 1275.47 & 193.51 & 444.13 & 1874.46 & 3170.96 & 1787.53 & 239.32 & 3600.15 \\
\bottomrule
\end{tabular}}
\label{table:sr_Rast}
\end{table}

\begin{table}[htbp]
\centering
\caption{Symbolic regression results on Livermore dataset.}
\resizebox{0.99\textwidth}{!}{
\begin{tabular}{cHcccccccccccccccccc}
\toprule
{\textbf{Benchmark}} & {\textbf{Expression}} & \multicolumn{9}{c}{\textbf{Recovery rate}} & \multicolumn{9}{c}{\textbf{Time cost (s)}} \\
{} & {} & {PSE} & {SPL} & {NGGP} & {DGSR} & {PySR} & {BMS} & {uDSR} & {TPSR} & {Operon} & {PSE} & {SPL} & {NGGP} & {DGSR} & {PySR} & {BMS} & {uDSR} & {TPSR} & {Operon} \\
\midrule
Livermore-1 & $x_{1} + \sin{\left(x_{1}^{2} \right)} + \frac{1}{3}$ & \cellcolor[HTML]{d3a9ce} \bfseries 100\% & 94\% & \cellcolor[HTML]{d3a9ce} \bfseries 100\% & 76\% & 89\% & 0\% & 0\% & 32\% & 0\% & \cellcolor[HTML]{9bcae9} \bfseries 6.19 & 56.39 & 77.48 & 895.84 & 29.98 & 1884.62 & 1879.05 & 202.74 & 3600.14 \\
Livermore-2 & $\sin{\left(x_{1}^{2} \right)} \cos{\left(x_{1} \right)} - 2$ & \cellcolor[HTML]{d3a9ce} \bfseries 100\% & 29\% & \cellcolor[HTML]{d3a9ce} \bfseries 100\% & \cellcolor[HTML]{d3a9ce} \bfseries 100\% & 55\% & 1\% & 0\% & 0\% & 0\% & \cellcolor[HTML]{9bcae9} \bfseries 2.81 & 1568.58 & 100.95 & 58.38 & 1378.44 & 1600.78 & 1744.70 & 64.20 & 3600.15 \\
Livermore-3 & $\sin{\left(x_{1}^{3} \right)} \cos{\left(x_{1}^{2} \right)} - 1$ & \cellcolor[HTML]{d3a9ce} \bfseries 100\% & 50\% & 96\% & \cellcolor[HTML]{d3a9ce} \bfseries 100\% & 86\% & 0\% & 0\% & 0\% & 0\% & \cellcolor[HTML]{9bcae9} \bfseries 65.57 & 121.77 & 205.71 & 84.68 & 398.74 & 1932.20 & 1712.40 & 74.85 & 3600.22 \\
Livermore-4 & \scalebox{0.8}{$\log{\left(x_{1} \right)} + \log{\left(x_{1} + 1 \right)} + \log{\left(x_{1}^{2} + 1 \right)}$} & \cellcolor[HTML]{d3a9ce} \bfseries 100\% & 61\% & \cellcolor[HTML]{d3a9ce} \bfseries 100\% & \cellcolor[HTML]{d3a9ce} \bfseries 100\% & \cellcolor[HTML]{d3a9ce} \bfseries 100\% & 0\% & \cellcolor[HTML]{d3a9ce} \bfseries 100\% & 0\% & 0\% & 21.71 & 661.39 & 29.00 & 42.79 & \cellcolor[HTML]{9bcae9} \bfseries 6.56 & 3867.33 & 840.65 & 165.99 & 3600.14 \\
Livermore-5 & $x_{1}^{4} - x_{1}^{3} + x_{1}^{2} - x_{2}$ & \cellcolor[HTML]{d3a9ce} \bfseries 100\% & \cellcolor[HTML]{d3a9ce} \bfseries 100\% & \cellcolor[HTML]{d3a9ce} \bfseries 100\% & 0\% & 99\% & 4\% & 0\% & 0\% & 0\% & \cellcolor[HTML]{9bcae9} \bfseries 7.20 & 1714.67 & 65.75 & 1060.11 & 131.11 & 3338.33 & 1732.95 & 150.61 & 3600.16 \\
Livermore-6 & $4 x_{1}^{4} + 3 x_{1}^{3} + 2 x_{1}^{2} + x_{1}$ & \cellcolor[HTML]{d3a9ce} \bfseries 100\% & 8\% & 98\% & \cellcolor[HTML]{d3a9ce} \bfseries 100\% & 0\% & 21\% & \cellcolor[HTML]{d3a9ce} \bfseries 100\% & 83\% & 0\% & \cellcolor[HTML]{9bcae9} \bfseries 15.13 & 1958.34 & 189.15 & 152.36 & 2063.86 & 3770.60 & 999.05 & 110.79 & 3600.15 \\
Livermore-7 & $\sinh{\left(x_{1} \right)}$ & \cellcolor[HTML]{d3a9ce} \bfseries 100\% & 18\% & 0\% & 0\% & 0\% & 1\% & 0\% & 0\% & 0\% & \cellcolor[HTML]{9bcae9} \bfseries 2.29 & 1362.29 & 643.53 & 1140.42 & 36.12 & 3161.92 & 1731.10 & 55.68 & 3600.13 \\
Livermore-8 & $\cosh{\left(x_{1} \right)}$ & \cellcolor[HTML]{d3a9ce} \bfseries 100\% & 6\% & 0\% & 0\% & 0\% & 1\% & 0\% & 0\% & 0\% & \cellcolor[HTML]{9bcae9} \bfseries 2.19 & 946.12 & 672.93 & 1140.30 & 89.14 & 2477.80 & 1730.20 & 57.07 & 3600.16 \\
Livermore-9 & \scalebox{1.0}{$x_{1}^{9} + x_{1}^{8} + \cdots + x_{1}^{2} + x_{1}$} & \cellcolor[HTML]{d3a9ce} \bfseries 100\% & 21\% & 79\% & 43\% & 0\% & 0\% & 0\% & 0\% & 0\% & \cellcolor[HTML]{9bcae9} \bfseries 51.88 & 1375.24 & 405.70 & 924.97 & 2103.65 & 4459.96 & 1923.50 & 447.45 & 3600.13 \\
Livermore-10 & $6 \sin{\left(x_{1} \right)} \cos{\left(x_{2} \right)}$ & \cellcolor[HTML]{d3a9ce} \bfseries 100\% & 75\% & 5\% & 35\% & 68\% & 23\% & 5\% & 0\% & 0\% & \cellcolor[HTML]{9bcae9} \bfseries 10.96 & 1773.06 & 813.33 & 1057.65 & 3418.58 & 3727.09 & 1766.85 & 288.12 & 3600.16 \\
Livermore-11 & $\frac{x_{1}^{2} x_{1}^{2}}{x_{1} + x_{2}}$ & \cellcolor[HTML]{d3a9ce} \bfseries 100\% & 0\% & \cellcolor[HTML]{d3a9ce} \bfseries 100\% & 18\% & 80\% & 1\% & 30\% & 0\% & 0\% & \cellcolor[HTML]{9bcae9} \bfseries 1.94 & 1684.00 & 66.84 & 12.43 & 5028.28 & 2366.54 & 1779.25 & 702.87 & 3600.16 \\
Livermore-12 & $\frac{x_{1}^{5}}{x_{2}^{3}}$ & \cellcolor[HTML]{d3a9ce} \bfseries 100\% & \cellcolor[HTML]{d3a9ce} \bfseries 100\% & 62\% & \cellcolor[HTML]{d3a9ce} \bfseries 100\% & 39\% & 0\% & 0\% & 0\% & 0\% & \cellcolor[HTML]{9bcae9} \bfseries 2.19 & 1795.80 & 117.98 & 9.24 & 5396.28 & 3080.95 & 1926.50 & 551.48 & 3600.21 \\
Livermore-13 & $x_{1}^{\frac{1}{3}}$ & \cellcolor[HTML]{d3a9ce} \bfseries 100\% & 12\% & 97\% & 98\% & 80\% & 0\% & 75\% & 0\% & 0\% & 56.52 & 2665.69 & \cellcolor[HTML]{9bcae9} \bfseries 53.60 & 93.35 & 172.28 & 2554.68 & 1324.55 & 100.18 & 3600.15 \\
Livermore-14 & \scalebox{0.8}{$x_{1}^{3} + x_{1}^{2} + x_{1} + \sin{\left(x_{1} \right)} + \sin{\left(x_{1}^{2} \right)}$} & \cellcolor[HTML]{d3a9ce} \bfseries 100\% & \cellcolor[HTML]{d3a9ce} \bfseries 100\% & 99\% & 0\% & \cellcolor[HTML]{d3a9ce} \bfseries 100\% & 0\% & 0\% & 0\% & 0\% & \cellcolor[HTML]{9bcae9} \bfseries 3.05 & 1875.82 & 143.00 & 52.13 & 9.92 & 4506.13 & 1818.30 & 127.23 & 3600.14 \\
Livermore-15 & $x_{1}^{\frac{1}{5}}$ & \cellcolor[HTML]{d3a9ce} \bfseries 100\% & 0\% & 98\% & \cellcolor[HTML]{d3a9ce} \bfseries 100\% & 37\% & 5\% & 5\% & 0\% & 25\% & 225.37 & 2662.18 & 133.97 & 187.01 & \cellcolor[HTML]{9bcae9} \bfseries 8.18 & 2384.41 & 1730.60 & 50.91 & 3600.13 \\
Livermore-16 & $x_{1}^{\frac{2}{5}}$ & \cellcolor[HTML]{d3a9ce} \bfseries 100\% & 0\% & 97\% & 0\% & 7\% & 12\% & 40\% & 0\% & 15\% & 550.75 & 3700.33 & 184.83 & 786.57 & 216.37 & 2933.22 & 1548.00 & \cellcolor[HTML]{9bcae9} \bfseries 66.14 & 3600.15 \\
Livermore-17 & $4 \sin{\left(x_{1} \right)} \cos{\left(x_{2} \right)}$ & \cellcolor[HTML]{d3a9ce} \bfseries 100\% & 89\% & 45\% & 64\% & \cellcolor[HTML]{d3a9ce} \bfseries 100\% & 31\% & 0\% & 0\% & 0\% & \cellcolor[HTML]{9bcae9} \bfseries 2.75 & 637.18 & 642.14 & 629.29 & 227.37 & 3472.83 & 1822.95 & 195.16 & 3600.13 \\
Livermore-18 & $\sin{\left(x_{1}^{2} \right)} \cos{\left(x_{1} \right)} - 5$ & \cellcolor[HTML]{d3a9ce} \bfseries 100\% & 18\% & 60\% & \cellcolor[HTML]{d3a9ce} \bfseries 100\% & 0\% & 0\% & 0\% & 0\% & 0\% & \cellcolor[HTML]{9bcae9} \bfseries 15.93 & 1132.71 & 600.62 & 146.83 & 1706.93 & 1561.24 & 1851.75 & 44.93 & 3600.22 \\
Livermore-19 & $x_{1}^{5} + x_{1}^{4} + x_{1}^{2} + x_{1}$ & \cellcolor[HTML]{d3a9ce} \bfseries 100\% & 89\% & \cellcolor[HTML]{d3a9ce} \bfseries 100\% & \cellcolor[HTML]{d3a9ce} \bfseries 100\% & \cellcolor[HTML]{d3a9ce} \bfseries 100\% & 6\% & \cellcolor[HTML]{d3a9ce} \bfseries 100\% & 72\% & 0\% & \cellcolor[HTML]{9bcae9} \bfseries 4.51 & 2043.58 & 23.67 & 7.36 & 4.55 & 3674.00 & 986.15 & 200.00 & 3600.23 \\
Livermore-20 & $e^{- x_{1}^{2}}$ & \cellcolor[HTML]{d3a9ce} \bfseries 100\% & \cellcolor[HTML]{d3a9ce} \bfseries 100\% & \cellcolor[HTML]{d3a9ce} \bfseries 100\% & \cellcolor[HTML]{d3a9ce} \bfseries 100\% & 17\% & 8\% & 60\% & 0\% & \cellcolor[HTML]{d3a9ce} \bfseries 100\% & \cellcolor[HTML]{9bcae9} \bfseries 2.09 & 3003.36 & 14.77 & 14.73 & 16.52 & 3493.23 & 1261.10 & 45.90 & 3600.22 \\
Livermore-21 & \scalebox{1.0}{$x_{1}^{8} + x_{1}^{7} + \cdots + x_{1}^{2} + x_{1}$} & \cellcolor[HTML]{d3a9ce} \bfseries 100\% & 52\% & 94\% & \cellcolor[HTML]{d3a9ce} \bfseries 100\% & 0\% & 0\% & \cellcolor[HTML]{d3a9ce} \bfseries 100\% & 0\% & 0\% & \cellcolor[HTML]{9bcae9} \bfseries 15.71 & 1970.23 & 294.23 & 58.87 & 1944.50 & 4362.53 & 157.50 & 345.79 & 3600.23 \\
Livermore-22 & $e^{- 0.5 x_{1}^{2}}$ & \cellcolor[HTML]{d3a9ce} \bfseries 100\% & \cellcolor[HTML]{d3a9ce} \bfseries 100\% & 83\% & 32\% & 0\% & 3\% & 30\% & 0\% & 85\% & 4.53 & 1776.83 & 337.37 & 872.14 & \cellcolor[HTML]{9bcae9} \bfseries 4.25 & 3891.06 & 1651.20 & 37.19 & 3600.14 \\
\bottomrule
\end{tabular}}
\label{table:sr_Livermore}
\end{table}

\begin{table}[htbp]
\centering
\caption{Symbolic regression results on Feynman dataset.}
\resizebox{0.99\textwidth}{!}{
\begin{tabular}{cHcccccccccccccccccc}
\toprule
{\textbf{Benchmark}} & {\textbf{Expression}} & \multicolumn{9}{c}{\textbf{Recovery rate}} & \multicolumn{9}{c}{\textbf{Time cost (s)}} \\
{} & {} & {PSE} & {SPL} & {NGGP} & {DGSR} & {PySR} & {BMS} & {uDSR} & {TPSR} & {Operon} & {PSE} & {SPL} & {NGGP} & {DGSR} & {PySR} & {BMS} & {uDSR} & {TPSR} & {Operon} \\
\midrule
Feynman-1 & $x_{1} x_{2}$ & \cellcolor[HTML]{d3a9ce} \bfseries 100\% & \cellcolor[HTML]{d3a9ce} \bfseries 100\% & \cellcolor[HTML]{d3a9ce} \bfseries 100\% & \cellcolor[HTML]{d3a9ce} \bfseries 100\% & \cellcolor[HTML]{d3a9ce} \bfseries 100\% & 97\% & \cellcolor[HTML]{d3a9ce} \bfseries 100\% & \cellcolor[HTML]{d3a9ce} \bfseries 100\% & 5\% & \cellcolor[HTML]{9bcae9} \bfseries 1.72 & 85.68 & 12.22 & 6.71 & 2.30 & 311.17 & 14.70 & 20.40 & 3600.50 \\
Feynman-2 & $\frac{x_{1}}{2 \left(x_{2} + 1\right)}$ & \cellcolor[HTML]{d3a9ce} \bfseries 100\% & 0\% & 99\% & \cellcolor[HTML]{d3a9ce} \bfseries 100\% & 88\% & 70\% & 90\% & 30\% & 0\% & \cellcolor[HTML]{9bcae9} \bfseries 2.71 & 1181.64 & 45.09 & 41.25 & 30.54 & 1291.45 & 1046.90 & 36.45 & 3600.23 \\
Feynman-3 & $x_{1} x_{2}^{2}$ & \cellcolor[HTML]{d3a9ce} \bfseries 100\% & \cellcolor[HTML]{d3a9ce} \bfseries 100\% & \cellcolor[HTML]{d3a9ce} \bfseries 100\% & \cellcolor[HTML]{d3a9ce} \bfseries 100\% & 98\% & 90\% & \cellcolor[HTML]{d3a9ce} \bfseries 100\% & 75\% & 0\% & \cellcolor[HTML]{9bcae9} \bfseries 1.86 & 88.65 & 11.84 & 2.88 & 40.27 & 517.39 & 43.45 & 53.70 & 3600.21 \\
Feynman-4 & $\frac{x_{1} x_{2}}{1 - \frac{x_{1} x_{2}}{3}} + 1$ & \cellcolor[HTML]{d3a9ce} \bfseries 100\% & 0\% & 0\% & 0\% & 0\% & 8\% & 0\% & 0\% & 0\% & \cellcolor[HTML]{9bcae9} \bfseries 41.91 & 1061.49 & 711.07 & 725.39 & 6074.75 & 2960.55 & 1505.85 & 153.97 & 3600.24 \\
Feynman-5 & $\frac{x_{1}}{x_{2}}$ & \cellcolor[HTML]{d3a9ce} \bfseries 100\% & \cellcolor[HTML]{d3a9ce} \bfseries 100\% & \cellcolor[HTML]{d3a9ce} \bfseries 100\% & 95\% & 97\% & 95\% & \cellcolor[HTML]{d3a9ce} \bfseries 100\% & 75\% & 25\% & \cellcolor[HTML]{9bcae9} \bfseries 2.08 & 78.55 & 12.44 & 3.08 & 109.72 & 362.67 & 42.25 & 65.98 & 3600.23 \\
Feynman-6 & $\frac{1}{2} x_{1} x_{2}^{2}$ & \cellcolor[HTML]{d3a9ce} \bfseries 100\% & 0\% & \cellcolor[HTML]{d3a9ce} \bfseries 100\% & \cellcolor[HTML]{d3a9ce} \bfseries 100\% & 96\% & 97\% & 90\% & 90\% & 0\% & \cellcolor[HTML]{9bcae9} \bfseries 2.16 & 1316.97 & 122.17 & 35.03 & 190.46 & 410.80 & 914.05 & 60.54 & 3600.25 \\
Feynman-7 & $\frac{3 x_{1} x_{2}}{2}$ & \cellcolor[HTML]{d3a9ce} \bfseries 100\% & 0\% & \cellcolor[HTML]{d3a9ce} \bfseries 100\% & \cellcolor[HTML]{d3a9ce} \bfseries 100\% & 73\% & 96\% & 75\% & \cellcolor[HTML]{d3a9ce} \bfseries 100\% & 0\% & \cellcolor[HTML]{9bcae9} \bfseries 2.20 & 1301.57 & 69.20 & 15.54 & 93.19 & 261.59 & 1192.05 & 20.46 & 3600.22 \\
Feynman-8 & $\frac{x_{1}}{e^{\frac{- x_{4} x_{5}}{x_{2} x_{3}}} + e^{\frac{x_{4} x_{5}}{x_{2} x_{3}}}}$ & \cellcolor[HTML]{d3a9ce} \bfseries 10\% & 0\% & 0\% & 0\% & 0\% & 0\% & 0\% & 0\% & 0\% & 6593.13 & 1224.00 & 920.51 & 898.20 & 10694.80 & 6302.95 & 1761.00 & \cellcolor[HTML]{9bcae9} \bfseries 457.10 & 3600.58 \\
Feynman-9 & $x_{1} x_{2} x_{3} \log{\left(\frac{x_{5}}{x_{4}} \right)}$ & \cellcolor[HTML]{d3a9ce} \bfseries 100\% & 0\% & 95\% & 55\% & 34\% & 13\% & 0\% & 0\% & 5\% & \cellcolor[HTML]{9bcae9} \bfseries 10.20 & 1084.67 & 295.75 & 517.40 & 3359.48 & 4533.53 & 1820.25 & 368.10 & 3600.60 \\
Feynman-10 & $\frac{x_{1} x_{4} \left(- x_{2} + x_{3}\right)}{x_{5}}$ & \cellcolor[HTML]{d3a9ce} \bfseries 100\% & 0\% & \cellcolor[HTML]{d3a9ce} \bfseries 100\% & \cellcolor[HTML]{d3a9ce} \bfseries 100\% & 39\% & 85\% & 0\% & 20\% & 5\% & \cellcolor[HTML]{9bcae9} \bfseries 3.30 & 1192.10 & 45.36 & 23.52 & 3267.69 & 1925.93 & 1796.45 & 482.20 & 3600.30 \\
Feynman-11 & $\frac{x_{1} x_{2}}{x_{5} \left(x_{3}^{2} - x_{4}^{2}\right)}$ & \cellcolor[HTML]{d3a9ce} \bfseries 100\% & 0\% & 25\% & \cellcolor[HTML]{d3a9ce} \bfseries 100\% & 71\% & 0\% & 0\% & 0\% & 0\% & \cellcolor[HTML]{9bcae9} \bfseries 21.10 & 1136.84 & 800.94 & 342.79 & 6531.40 & 3233.19 & 1794.45 & 270.30 & 3600.22 \\
Feynman-12 & $\frac{x_{1} x_{2}^{2} x_{3}}{3 x_{4} x_{5}}$ & \cellcolor[HTML]{d3a9ce} \bfseries 100\% & 0\% & 46\% & 95\% & 34\% & 4\% & 0\% & 5\% & 0\% & \cellcolor[HTML]{9bcae9} \bfseries 114.68 & 1181.74 & 661.46 & 331.04 & 2475.89 & 3811.45 & 1802.50 & 410.95 & 3600.45 \\
Feynman-13 & $x_{1} \left(e^{\frac{x_{2} x_{3}}{x_{4} x_{5}}} - 1\right)$ & \cellcolor[HTML]{d3a9ce} \bfseries 100\% & 0\% & 14\% & 10\% & 32\% & 0\% & 0\% & 0\% & 0\% & 1042.28 & 1164.30 & \cellcolor[HTML]{9bcae9} \bfseries 832.85 & 880.70 & 6402.28 & 4779.89 & 1795.00 & 846.88 & 3600.47 \\
Feynman-14 & $x_{1} x_{2} x_{5} \left(\frac{1}{x_{4}} -\frac{1}{x_{3}}\right)$ & \cellcolor[HTML]{d3a9ce} \bfseries 100\% & 0\% & 99\% & \cellcolor[HTML]{d3a9ce} \bfseries 100\% & 34\% & 77\% & 0\% & 10\% & 0\% & \cellcolor[HTML]{9bcae9} \bfseries 2.13 & 1189.21 & 225.26 & 100.04 & 3404.15 & 2785.54 & 1798.35 & 474.80 & 3600.14 \\
Feynman-15 & $x_{1} \left(x_{2} + x_{3} x_{4} \sin{\left(x_{5} \right)}\right)$ & \cellcolor[HTML]{d3a9ce} \bfseries 100\% & 0\% & \cellcolor[HTML]{d3a9ce} \bfseries 100\% & \cellcolor[HTML]{d3a9ce} \bfseries 100\% & 7\% & 42\% & 5\% & 5\% & 0\% & 40.22 & 1152.71 & 46.11 & \cellcolor[HTML]{9bcae9} \bfseries 11.41 & 3736.32 & 3649.05 & 1764.35 & 382.10 & 3600.17 \\
\bottomrule
\end{tabular}}
\label{table:sr_Feynman}
\end{table}

\subsection{Chaotic Dynamics}

\kredit{Tables \ref{table:cs_result_0_01}--\ref{table:cs_result_0_1} show the average recovery rate and time cost on 16 chaotic dynamics  under four different levels of Gaussian noise (aka, 1\%, 2\%, 5\%, and 10\% standard deviation), with time budget 10 minute.} Our PSE model is capable of identifying the underlying control equations within chaotic dynamics data more rapidly and accurately under a given time budget. Generally speaking, the ability to precisely recover the most complex symbolic expressions within a system of equations determines the overall effectiveness of the system's reconstruction. We have observed that these challenging expressions can be discovered by the PSE with a significantly higher recovery rate, demonstrating the superiority of our method.

\begin{table}[b!]
\caption{Recovery rate and time cost on chaotic dynamics task (1.0\% noise)}

\centering
\resizebox{\textwidth}{!}{
\begin{tabular}{llcccccccccccccccccccc}
\toprule
\multirow{2}{*}{Chaotic System} & \multirow{2}{*}{Variant} & \multicolumn{10}{c}{Recovery Rate (\%)} & \multicolumn{10}{c}{Avg. Time Cost (s)}\\
\cmidrule(lr){3-12} \cmidrule(lr){13-22}
& & PSE & Operon & BMS & PySR & NGGP & wAIC1 & wAIC2 & DGSR & uDSR & TPSR & PSE & Operon & BMS & PySR & NGGP & wAIC1 & wAIC2 & DGSR & uDSR & TPSR \\
\midrule
\multirow{3}{*}{ShimizuMorioka} & $\dot{x}$ & \textbf{100.0} & \textbf{100.0} & 98.1 & \textbf{100.0} & \textbf{100.0} & \textbf{100.0} & 84.6 & \textbf{100.0} & 58.3 & 0.0 & 611.6 & 600.0 & 602.6 & 685.6 & 607.7 & 679.8 & 760.0 & 613.1 & 721.9 & 585.4 \\
 & $\dot{y}$ & \textbf{100.0} & \textbf{100.0} & 33.3 & \textbf{100.0} & 31.1 & \textbf{100.0} & 0.0 & 32.6 & 0.0 & 0.0 & 608.9 & 600.0 & 718.1 & 873.3 & 613.3 & 710.1 & 777.4 & 611.7 & 744.0 & 641.5 \\
 & $\dot{z}$ & \textbf{100.0} & 90.0 & 37.0 & \textbf{100.0} & 60.7 & \textbf{100.0} & 0.0 & 71.7 & 0.0 & 0.0 & 611.4 & 600.1 & 752.0 & 848.1 & 611.0 & 692.4 & 793.8 & 610.0 & 746.7 & 614.2 \\
\midrule
\multirow{3}{*}{Rucklidge} & $\dot{x}$ & \textbf{100.0} & \textbf{100.0} & 34.0 & 65.3 & 28.8 & 0.0 & \textbf{100.0} & 21.7 & 0.0 & 0.0 & 608.4 & 600.0 & 664.3 & 846.4 & 543.9 & 679.5 & 798.1 & 610.6 & 651.7 & 632.5 \\
 & $\dot{y}$ & \textbf{100.0} & \textbf{100.0} & \textbf{100.0} & \textbf{100.0} & \textbf{100.0} & \textbf{100.0} & \textbf{100.0} & \textbf{100.0} & 0.0 & 0.0 & 611.2 & 600.1 & 602.6 & 655.9 & 530.7 & 683.0 & 770.2 & 607.3 & 671.8 & 599.9 \\
 & $\dot{z}$ & \textbf{100.0} & \textbf{100.0} & 35.8 & \textbf{100.0} & 71.2 & \textbf{100.0} & 91.7 & \textbf{100.0} & 0.0 & 0.0 & 608.5 & 600.1 & 723.3 & 689.1 & 551.0 & 701.4 & 797.4 & 610.0 & 670.0 & 681.1 \\
\midrule
\multirow{3}{*}{SprottJerk} & $\dot{x}$ & \textbf{100.0} & \textbf{100.0} & 96.1 & \textbf{100.0} & \textbf{100.0} & \textbf{100.0} & \textbf{100.0} & \textbf{100.0} & \textbf{100.0} & 0.0 & 609.8 & 600.1 & 603.3 & 646.0 & 549.8 & 664.0 & 772.4 & 615.0 & 675.2 & 665.5 \\
 & $\dot{y}$ & \textbf{100.0} & \textbf{100.0} & 98.0 & \textbf{100.0} & \textbf{100.0} & \textbf{100.0} & 91.7 & \textbf{100.0} & 35.7 & 0.0 & 610.1 & 600.1 & 602.6 & 637.7 & 590.3 & 686.6 & 802.9 & 602.8 & 708.7 & 629.6 \\
 & $\dot{z}$ & \textbf{100.0} & \textbf{100.0} & 7.8 & 0.0 & 3.4 & \textbf{100.0} & 83.3 & 0.0 & 0.0 & 0.0 & 610.1 & 600.0 & 736.0 & 1095.2 & 606.3 & 694.2 & 816.4 & 609.8 & 746.1 & 602.8 \\
\midrule
\multirow{4}{*}{HyperLorenz} & $\dot{w}$ & \textbf{100.0} & \textbf{100.0} & 31.5 & \textbf{100.0} & 0.0 & 26.0 & 53.8 & 0.0 & 0.0 & 0.0 & 610.3 & 600.1 & 729.4 & 734.3 & 506.1 & 666.4 & 758.7 & 609.6 & 514.5 & 591.8 \\
 & $\dot{x}$ & \textbf{100.0} & \textbf{100.0} & 38.2 & \textbf{100.0} & 0.0 & 96.0 & \textbf{100.0} & 0.0 & 0.0 & 0.0 & 608.0 & 600.1 & 744.5 & 664.3 & 500.3 & 678.3 & 780.4 & 600.6 & 527.5 & 651.8 \\
 & $\dot{y}$ & \textbf{100.0} & \textbf{100.0} & 21.8 & \textbf{100.0} & 27.9 & \textbf{100.0} & 0.0 & 17.4 & 0.0 & 0.0 & 612.3 & 600.0 & 798.1 & 723.3 & 548.3 & 677.4 & 761.8 & 589.6 & 522.9 & 642.5 \\
 & $\dot{z}$ & \textbf{100.0} & \textbf{100.0} & 42.6 & \textbf{100.0} & 59.0 & 22.0 & \textbf{100.0} & \textbf{100.0} & 0.0 & 40.0 & 610.0 & 600.0 & 797.0 & 664.0 & 515.8 & 639.1 & 773.7 & 612.0 & 491.7 & 711.3 \\
\midrule
\multirow{4}{*}{HyperJha} & $\dot{w}$ & \textbf{100.0} & \textbf{100.0} & 50.0 & \textbf{100.0} & 0.0 & 16.0 & 53.8 & 0.0 & 0.0 & 0.0 & 608.9 & 600.1 & 650.3 & 730.6 & 427.3 & 663.4 & 772.6 & 612.4 & 476.0 & 676.1 \\
 & $\dot{x}$ & \textbf{100.0} & \textbf{100.0} & 46.3 & \textbf{100.0} & 0.0 & 64.0 & \textbf{100.0} & 0.0 & 0.0 & 0.0 & 610.4 & 600.1 & 658.4 & 672.1 & 432.1 & 681.6 & 791.4 & 598.1 & 534.5 & 684.6 \\
 & $\dot{y}$ & \textbf{100.0} & \textbf{100.0} & 37.0 & 95.9 & 27.9 & 0.0 & 0.0 & 43.5 & 0.0 & 0.0 & 612.1 & 600.1 & 713.4 & 772.0 & 478.2 & 692.0 & 782.1 & 613.2 & 521.4 & 620.3 \\
 & $\dot{z}$ & \textbf{100.0} & \textbf{100.0} & 50.0 & \textbf{100.0} & 57.4 & 44.0 & 0.0 & \textbf{100.0} & 0.0 & 0.0 & 606.6 & 600.0 & 745.3 & 666.6 & 440.1 & 662.7 & 762.4 & 611.1 & 468.0 & 740.8 \\
\midrule
\multirow{4}{*}{HyperPang} & $\dot{w}$ & \textbf{100.0} & \textbf{100.0} & 48.1 & \textbf{100.0} & \textbf{100.0} & 46.0 & \textbf{100.0} & \textbf{100.0} & 0.0 & 0.0 & 610.0 & 600.0 & 609.5 & 728.0 & 563.4 & 688.4 & 752.8 & 608.8 & 559.8 & 708.8 \\
 & $\dot{x}$ & \textbf{100.0} & \textbf{100.0} & 44.4 & \textbf{100.0} & 90.2 & 80.0 & \textbf{100.0} & 87.0 & 0.0 & 0.0 & 611.3 & 600.0 & 714.5 & 718.2 & 501.9 & 698.7 & 807.2 & 614.3 & 576.6 & 656.2 \\
 & $\dot{y}$ & \textbf{100.0} & \textbf{100.0} & 40.7 & 89.8 & 0.0 & 56.0 & \textbf{100.0} & 0.0 & 0.0 & 0.0 & 607.5 & 600.1 & 793.0 & 705.0 & 532.8 & 686.2 & 775.3 & 610.7 & 542.2 & 665.1 \\
 & $\dot{z}$ & \textbf{100.0} & \textbf{100.0} & 44.4 & \textbf{100.0} & 67.2 & 0.0 & 0.0 & \textbf{100.0} & 0.0 & 0.0 & 607.5 & 600.1 & 797.1 & 641.5 & 508.7 & 678.8 & 771.9 & 612.8 & 544.5 & 614.5 \\
\midrule
\multirow{3}{*}{GenesioTesi} & $\dot{x}$ & \textbf{100.0} & \textbf{100.0} & 94.4 & \textbf{100.0} & \textbf{100.0} & \textbf{100.0} & \textbf{100.0} & \textbf{100.0} & 87.5 & 0.0 & 610.8 & 600.0 & 602.9 & 665.3 & 605.5 & 649.1 & 733.3 & 612.3 & 786.9 & 663.3 \\
 & $\dot{y}$ & \textbf{100.0} & \textbf{100.0} & \textbf{100.0} & \textbf{100.0} & \textbf{100.0} & \textbf{100.0} & 61.5 & \textbf{100.0} & 0.0 & 0.0 & 608.3 & 600.0 & 602.9 & 666.7 & 601.6 & 648.1 & 746.9 & 611.3 & 776.2 & 709.8 \\
 & $\dot{z}$ & \textbf{100.0} & \textbf{100.0} & 9.4 & 24.5 & 0.0 & 96.0 & 0.0 & 0.0 & 0.0 & 0.0 & 609.7 & 600.1 & 671.1 & 1098.9 & 612.0 & 693.0 & 775.7 & 612.7 & 804.8 & 641.8 \\
\midrule
\multirow{3}{*}{NewtonLiepnik} & $\dot{x}$ & \textbf{100.0} & \textbf{100.0} & 49.1 & 83.7 & 23.0 & \textbf{100.0} & \textbf{100.0} & 10.6 & 0.0 & 0.0 & 612.5 & 600.1 & 743.3 & 837.5 & 596.9 & 691.2 & 805.4 & 615.0 & 874.2 & 581.7 \\
 & $\dot{y}$ & \textbf{100.0} & \textbf{100.0} & 38.2 & \textbf{100.0} & 9.8 & 0.0 & 0.0 & 0.0 & 0.0 & 0.0 & 608.0 & 600.1 & 785.3 & 759.6 & 608.1 & 685.6 & 792.7 & 605.3 & 856.6 & 728.0 \\
 & $\dot{z}$ & \textbf{100.0} & \textbf{100.0} & 7.3 & \textbf{100.0} & 18.0 & 0.0 & 0.0 & 37.0 & 0.0 & 0.0 & 609.6 & 600.1 & 718.8 & 741.6 & 604.7 & 659.3 & 761.0 & 620.1 & 873.5 & 629.6 \\
\midrule
\multirow{3}{*}{DequanLi} & $\dot{x}$ & \textbf{100.0} & 90.0 & 39.6 & 50.0 & 25.9 & 6.0 & 0.0 & 2.2 & 0.0 & 0.0 & 611.6 & 600.0 & 746.0 & 715.7 & 529.8 & 672.9 & 772.5 & 605.0 & 550.4 & 720.6 \\
 & $\dot{y}$ & \textbf{100.0} & \textbf{100.0} & 41.2 & \textbf{100.0} & 72.4 & 90.0 & 41.7 & \textbf{100.0} & 0.0 & 0.0 & 610.6 & 600.1 & 770.9 & 650.5 & 554.1 & 656.6 & 743.7 & 614.3 & 577.4 & 666.0 \\
 & $\dot{z}$ & \textbf{100.0} & 90.0 & 43.1 & \textbf{100.0} & 12.1 & 94.0 & 33.3 & 84.8 & 0.0 & 0.0 & 609.9 & 600.0 & 702.6 & 661.8 & 494.3 & 681.3 & 759.5 & 610.4 & 577.8 & 689.1 \\
\midrule
\multirow{3}{*}{Finance} & $\dot{x}$ & \textbf{100.0} & \textbf{100.0} & 22.6 & \textbf{100.0} & 65.5 & \textbf{100.0} & 0.0 & 69.6 & 0.0 & 0.0 & 611.1 & 600.1 & 764.7 & 894.7 & 612.0 & 683.7 & 794.2 & 610.5 & 692.2 & 688.0 \\
 & $\dot{y}$ & \textbf{100.0} & 10.0 & 5.7 & \textbf{100.0} & 20.7 & 0.0 & 0.0 & 28.3 & 0.0 & 0.0 & 607.9 & 600.1 & 786.3 & 925.8 & 608.0 & 645.4 & 732.7 & 607.2 & 711.3 & 643.9 \\
 & $\dot{z}$ & \textbf{100.0} & \textbf{100.0} & 24.5 & \textbf{100.0} & \textbf{100.0} & 70.0 & 25.0 & 95.7 & 0.0 & 0.0 & 613.8 & 600.1 & 654.5 & 768.1 & 609.3 & 693.3 & 808.4 & 577.0 & 684.9 & 645.9 \\
\midrule
\multirow{3}{*}{Duffing} & $\dot{x}$ & \textbf{100.0} & \textbf{100.0} & 96.2 & \textbf{100.0} & \textbf{100.0} & \textbf{100.0} & \textbf{100.0} & \textbf{100.0} & 75.0 & 0.0 & 609.2 & 600.1 & 602.6 & 729.7 & 610.2 & 663.2 & 785.7 & 610.3 & 653.2 & 739.6 \\
 & $\dot{y}$ & \textbf{100.0} & 70.0 & 15.1 & 69.4 & 5.0 & 92.0 & 0.0 & 0.0 & 0.0 & 0.0 & 606.5 & 600.1 & 795.8 & 837.8 & 617.5 & 677.5 & 746.3 & 616.6 & 693.8 & 724.7 \\
 & $\dot{z}$ & \textbf{100.0} & 70.0 & 54.7 & \textbf{100.0} & 70.0 & 0.0 & 0.0 & 47.8 & 40.0 & 0.0 & 610.2 & 600.1 & 607.4 & 942.7 & 580.5 & 624.6 & 739.4 & 582.6 & 650.8 & 610.9 \\
\midrule
\multirow{3}{*}{Laser} & $\dot{x}$ & \textbf{100.0} & 20.0 & 18.9 & 55.1 & 0.0 & 12.0 & 7.7 & 45.7 & 0.0 & 0.0 & 609.8 & 600.1 & 796.0 & 792.2 & 608.8 & 688.3 & 783.5 & 609.6 & 699.3 & 699.2 \\
 & $\dot{y}$ & \textbf{100.0} & 20.0 & 1.9 & \textbf{100.0} & 48.3 & 94.0 & \textbf{100.0} & \textbf{100.0} & 0.0 & 0.0 & 610.5 & 600.0 & 795.7 & 698.9 & 611.1 & 673.1 & 767.9 & 614.0 & 710.1 & 698.1 \\
 & $\dot{z}$ & \textbf{100.0} & \textbf{100.0} & 26.4 & \textbf{100.0} & 53.3 & 18.0 & 0.0 & 84.8 & 0.0 & 0.0 & 609.7 & 600.0 & 782.3 & 744.3 & 599.3 & 693.3 & 744.9 & 612.2 & 720.5 & 634.5 \\
\midrule
\multirow{3}{*}{Hadley} & $\dot{x}$ & \textbf{100.0} & 30.0 & 17.6 & 72.9 & 0.0 & 0.0 & 0.0 & 0.0 & 0.0 & 0.0 & 610.3 & 600.1 & 795.6 & 947.2 & 611.8 & 637.5 & 749.1 & 610.6 & 627.8 & 662.5 \\
 & $\dot{y}$ & \textbf{100.0} & \textbf{100.0} & 19.6 & \textbf{100.0} & 0.0 & 0.0 & 0.0 & 0.0 & 0.0 & 0.0 & 611.5 & 600.0 & 766.5 & 797.7 & 607.3 & 644.1 & 721.6 & 614.2 & 643.4 & 615.8 \\
 & $\dot{z}$ & \textbf{100.0} & \textbf{100.0} & 39.2 & \textbf{100.0} & 13.8 & 0.0 & 0.0 & 6.5 & 0.0 & 0.0 & 608.8 & 600.0 & 734.8 & 780.2 & 613.0 & 674.5 & 792.6 & 615.6 & 658.9 & 547.9 \\
\midrule
\multirow{3}{*}{Brusselator} & $\dot{x}$ & \textbf{100.0} & 90.0 & 0.0 & 8.2 & 3.3 & 0.0 & 0.0 & 0.0 & 0.0 & 0.0 & 613.6 & 600.0 & 800.0 & 1117.7 & 620.6 & 635.8 & 775.0 & 594.1 & 802.2 & 666.4 \\
 & $\dot{y}$ & \textbf{100.0} & \textbf{100.0} & 5.7 & 98.0 & 16.7 & 8.0 & 0.0 & 4.3 & 0.0 & 0.0 & 610.1 & 600.1 & 785.3 & 865.4 & 618.0 & 623.7 & 762.5 & 587.8 & 808.5 & 655.8 \\
 & $\dot{z}$ & \textbf{100.0} & 40.0 & 32.1 & \textbf{100.0} & 62.7 & 72.0 & 23.1 & 19.6 & 43.3 & 0.0 & 614.2 & 600.1 & 611.0 & 1061.2 & 600.2 & 650.7 & 764.4 & 612.7 & 823.0 & 708.2 \\
\midrule
\multirow{3}{*}{FitzHughNagumo} & $\dot{x}$ & \textbf{100.0} & 0.0 & 1.9 & 0.0 & 0.0 & 0.0 & 0.0 & 0.0 & 0.0 & 0.0 & 611.8 & 600.1 & 744.9 & 1146.6 & 604.2 & 647.6 & 785.3 & 1701.8 & 580.5 & 680.4 \\
 & $\dot{y}$ & \textbf{100.0} & \textbf{100.0} & 37.7 & \textbf{100.0} & 93.1 & 0.0 & 0.0 & 37.0 & 0.0 & 0.0 & 610.8 & 600.1 & 753.1 & 854.2 & 612.2 & 653.7 & 792.1 & 587.3 & 575.2 & 654.9 \\
 & $\dot{z}$ & \textbf{100.0} & \textbf{100.0} & 58.5 & \textbf{100.0} & 94.8 & 44.0 & 58.3 & 97.8 & 31.0 & 35.0 & 609.7 & 600.1 & 604.9 & 635.6 & 593.2 & 688.3 & 780.3 & 604.1 & 591.3 & 667.8 \\
\midrule
\multirow{3}{*}{KawczynskiStrizhak} & $\dot{x}$ & 54.5 & 10.0 & 3.8 & 0.0 & 1.7 & 16.0 & 0.0 & 0.0 & 0.0 & 0.0 & 610.6 & 600.0 & 797.4 & 1073.1 & 610.1 & 672.4 & 773.6 & 611.6 & 747.1 & 570.8 \\
 & $\dot{y}$ & \textbf{100.0} & \textbf{100.0} & 79.2 & \textbf{100.0} & 76.3 & 0.0 & 0.0 & 41.3 & 0.0 & 0.0 & 607.1 & 600.0 & 790.5 & 801.3 & 601.3 & 681.6 & 791.2 & 610.0 & 765.3 & 636.1 \\
 & $\dot{z}$ & 90.9 & \textbf{100.0} & 58.5 & \textbf{100.0} & \textbf{100.0} & 0.0 & 0.0 & \textbf{100.0} & 0.0 & 0.0 & 610.1 & 600.1 & 677.1 & 783.8 & 610.9 & 671.5 & 763.8 & 616.6 & 758.0 & 656.3 \\
\midrule
\bfseries Average 95\% CI & & \cellcolor[RGB]{211,169,206}\makecell{\bfseries 97.2 \\ \bfseries $\pm$ 6.1} & \cellcolor[RGB]{211,169,206}\makecell{\bfseries 66.2 \\ \bfseries $\pm$ 21.3} & \cellcolor[RGB]{211,169,206}\makecell{\bfseries 17.3 \\ \bfseries $\pm$ 8.0} & \cellcolor[RGB]{211,169,206}\makecell{\bfseries 57.2 \\ \bfseries $\pm$ 20.7} & \cellcolor[RGB]{211,169,206}\makecell{\bfseries 7.3 \\ \bfseries $\pm$ 5.6} & \cellcolor[RGB]{211,169,206}\makecell{\bfseries 21.0 \\ \bfseries $\pm$ 20.8} & \cellcolor[RGB]{211,169,206}\makecell{\bfseries 10.9 \\ \bfseries $\pm$ 15.9} & \cellcolor[RGB]{211,169,206}\makecell{\bfseries 8.2 \\ \bfseries $\pm$ 8.0} & \cellcolor[RGB]{211,169,206}\makecell{\bfseries 0.0 \\ \bfseries $\pm$ 0.0} & \cellcolor[RGB]{211,169,206}\makecell{\bfseries 0.0 \\ \bfseries $\pm$ 0.0} & \cellcolor[RGB]{102,204,255}\makecell{\bfseries 610.07 \\ \bfseries $\pm$ 0.49} & \cellcolor[RGB]{102,204,255}\makecell{\bfseries 600.05 \\ \bfseries $\pm$ 0.00} & \cellcolor[RGB]{102,204,255}\makecell{\bfseries 716.57 \\ \bfseries $\pm$ 20.25} & \cellcolor[RGB]{102,204,255}\makecell{\bfseries 792.03 \\ \bfseries $\pm$ 39.56} & \cellcolor[RGB]{102,204,255}\makecell{\bfseries 571.09 \\ \bfseries $\pm$ 15.03} & \cellcolor[RGB]{102,204,255}\makecell{\bfseries 671.86 \\ \bfseries $\pm$ 5.78} & \cellcolor[RGB]{102,204,255}\makecell{\bfseries 773.02 \\ \bfseries $\pm$ 6.04} & \cellcolor[RGB]{102,204,255}\makecell{\bfseries 629.34 \\ \bfseries $\pm$ 43.16} & \cellcolor[RGB]{102,204,255}\makecell{\bfseries 664.52 \\ \bfseries $\pm$ 30.80} & \cellcolor[RGB]{102,204,255}\makecell{\bfseries 655.96 \\ \bfseries $\pm$ 12.59} \\
\bottomrule
\end{tabular}}
\label{table:cs_result_0_01}
\end{table}


\begin{table}[b!]
\caption{Recovery rate and time cost on chaotic dynamics task (2.0\% noise)}
\centering
\resizebox{\textwidth}{!}{
\begin{tabular}{llcccccccccccccccccccc}
\toprule
\multirow{2}{*}{Chaotic System} & \multirow{2}{*}{Variant} & \multicolumn{10}{c}{Recovery Rate (\%)} & \multicolumn{10}{c}{Avg. Time Cost (s)}\\
\cmidrule(lr){3-12} \cmidrule(lr){13-22}
& & PSE & Operon & BMS & PySR & NGGP & wAIC1 & wAIC2 & DGSR & uDSR & TPSR & PSE & Operon & BMS & PySR & NGGP & wAIC1 & wAIC2 & DGSR & uDSR & TPSR \\
\midrule
\multirow{3}{*}{ShimizuMorioka} & $\dot{x}$ & 90.0 & \textbf{100.0} & 94.9 & \textbf{100.0} & 96.2 & 98.0 & 88.0 & 96.2 & 70.7 & 0.0 & 613.3 & 600.1 & 622.6 & 704.7 & 622.0 & 699.8 & 783.6 & 640.1 & 737.9 & 695.1 \\
 & $\dot{y}$ & \textbf{100.0} & \textbf{100.0} & 8.3 & \textbf{100.0} & 23.4 & \textbf{100.0} & 12.0 & 23.4 & 0.0 & 0.0 & 614.7 & 600.0 & 739.2 & 899.5 & 642.7 & 739.6 & 809.4 & 629.6 & 781.1 & 606.8 \\
 & $\dot{z}$ & 90.0 & 77.8 & 8.3 & \textbf{100.0} & 70.7 & \textbf{100.0} & 0.0 & 46.1 & 0.0 & 0.0 & 609.4 & 600.1 & 774.0 & 883.6 & 636.8 & 720.2 & 826.4 & 626.1 & 765.7 & 604.1 \\
\midrule
\multirow{3}{*}{Rucklidge} & $\dot{x}$ & \textbf{100.0} & \textbf{100.0} & 8.3 & 50.0 & 23.4 & 24.0 & 92.0 & 33.2 & 0.0 & 0.0 & 624.9 & 600.0 & 687.9 & 870.5 & 556.0 & 713.3 & 836.3 & 637.4 & 677.3 & 616.9 \\
 & $\dot{y}$ & 70.0 & \textbf{100.0} & 96.2 & \textbf{100.0} & 96.2 & 88.0 & 82.0 & 96.2 & 0.0 & 0.0 & 610.1 & 600.0 & 624.8 & 687.4 & 550.0 & 713.8 & 806.9 & 620.5 & 703.2 & 646.1 \\
 & $\dot{z}$ & \textbf{100.0} & \textbf{100.0} & 8.3 & \textbf{100.0} & 46.1 & \textbf{100.0} & \textbf{100.0} & 96.2 & 0.0 & 0.0 & 610.1 & 600.1 & 751.8 & 703.7 & 567.8 & 724.5 & 825.0 & 633.2 & 683.4 & 623.5 \\
\midrule
\multirow{3}{*}{SprottJerk} & $\dot{x}$ & \textbf{100.0} & \textbf{100.0} & 66.6 & \textbf{100.0} & 96.2 & \textbf{100.0} & 80.0 & 96.2 & 96.2 & 0.0 & 616.1 & 600.1 & 632.9 & 673.8 & 572.4 & 688.4 & 806.0 & 634.6 & 700.4 & 632.5 \\
 & $\dot{y}$ & \textbf{100.0} & \textbf{100.0} & 94.9 & \textbf{100.0} & 96.2 & \textbf{100.0} & 92.0 & 96.2 & 8.3 & 0.0 & 615.3 & 600.1 & 626.4 & 662.4 & 604.7 & 704.9 & 823.8 & 618.3 & 727.6 & 630.1 \\
 & $\dot{z}$ & 90.0 & \textbf{100.0} & 88.0 & 0.0 & 0.0 & 44.0 & 68.0 & 0.0 & 0.0 & 0.0 & 618.0 & 600.0 & 755.3 & 1144.9 & 618.8 & 723.8 & 846.4 & 623.6 & 768.3 & 727.2 \\
\midrule
\multirow{4}{*}{HyperLorenz} & $\dot{w}$ & \textbf{100.0} & \textbf{100.0} & 23.4 & \textbf{100.0} & 0.0 & 24.0 & 58.0 & 0.0 & 0.0 & 0.0 & 614.5 & 600.1 & 748.6 & 758.6 & 524.0 & 686.0 & 786.4 & 622.6 & 540.2 & 651.3 \\
 & $\dot{x}$ & 90.9 & \textbf{100.0} & 8.3 & \textbf{100.0} & 0.0 & 76.0 & \textbf{100.0} & 0.0 & 0.0 & 0.0 & 613.2 & 600.1 & 780.6 & 693.6 & 521.6 & 706.2 & 798.4 & 626.5 & 548.5 & 624.6 \\
 & $\dot{y}$ & \textbf{100.0} & \textbf{100.0} & 33.2 & 89.5 & 23.4 & \textbf{100.0} & 0.0 & 18.6 & 0.0 & 0.0 & 608.7 & 600.0 & 831.9 & 743.3 & 561.3 & 695.4 & 789.7 & 605.6 & 541.3 & 677.1 \\
 & $\dot{z}$ & \textbf{100.0} & \textbf{100.0} & 0.0 & \textbf{100.0} & 70.7 & 34.0 & 84.0 & 96.2 & 0.0 & 40.0 & 615.6 & 600.1 & 825.2 & 686.6 & 530.0 & 656.5 & 806.3 & 642.6 & 507.6 & 673.1 \\
\midrule
\multirow{4}{*}{HyperJha} & $\dot{w}$ & 90.9 & \textbf{100.0} & 23.9 & \textbf{100.0} & 0.0 & 34.0 & 14.0 & 0.0 & 0.0 & 0.0 & 615.6 & 600.1 & 675.8 & 762.4 & 439.9 & 694.6 & 803.5 & 638.7 & 498.2 & 666.3 \\
 & $\dot{x}$ & \textbf{100.0} & \textbf{100.0} & 39.8 & \textbf{100.0} & 0.0 & 78.0 & 68.0 & 0.0 & 0.0 & 0.0 & 621.9 & 600.0 & 683.7 & 695.2 & 443.5 & 705.8 & 812.3 & 622.3 & 556.3 & 619.2 \\
 & $\dot{y}$ & \textbf{100.0} & \textbf{100.0} & 8.3 & 84.2 & 23.4 & 0.0 & 0.0 & 0.0 & 0.0 & 0.0 & 614.0 & 600.1 & 730.9 & 790.9 & 498.9 & 719.9 & 805.1 & 632.9 & 539.2 & 671.6 \\
 & $\dot{z}$ & \textbf{100.0} & \textbf{100.0} & 23.9 & \textbf{100.0} & 50.3 & 48.0 & 0.0 & 96.2 & 0.0 & 0.0 & 612.0 & 600.1 & 767.8 & 685.7 & 458.8 & 691.6 & 796.3 & 635.0 & 488.4 & 623.0 \\
\midrule
\multirow{4}{*}{HyperPang} & $\dot{w}$ & \textbf{100.0} & \textbf{100.0} & 39.8 & \textbf{100.0} & 96.2 & 40.0 & 98.0 & 96.2 & 0.0 & 0.0 & 611.8 & 600.1 & 627.5 & 764.2 & 589.4 & 720.5 & 775.1 & 637.1 & 576.2 & 641.7 \\
 & $\dot{x}$ & \textbf{100.0} & \textbf{100.0} & 54.5 & \textbf{100.0} & 90.7 & 68.0 & 76.0 & 86.3 & 0.0 & 0.0 & 625.9 & 600.1 & 743.4 & 748.6 & 517.4 & 724.9 & 836.8 & 644.4 & 599.2 & 665.8 \\
 & $\dot{y}$ & 72.7 & \textbf{100.0} & 20.7 & 52.6 & 0.0 & 50.0 & \textbf{100.0} & 0.0 & 0.0 & 0.0 & 614.8 & 600.0 & 828.3 & 729.7 & 554.3 & 701.7 & 811.0 & 635.4 & 566.1 & 606.3 \\
 & $\dot{z}$ & \textbf{100.0} & \textbf{100.0} & 54.5 & \textbf{100.0} & 50.0 & 0.0 & 0.0 & 96.2 & 0.0 & 0.0 & 613.5 & 600.1 & 823.1 & 659.6 & 521.0 & 700.5 & 803.6 & 634.8 & 562.0 & 590.3 \\
\midrule
\multirow{3}{*}{GenesioTesi} & $\dot{x}$ & \textbf{100.0} & \textbf{100.0} & 61.5 & \textbf{100.0} & 96.2 & \textbf{100.0} & 92.0 & 96.2 & 86.3 & 0.0 & 614.9 & 600.1 & 625.8 & 689.7 & 622.2 & 664.0 & 753.5 & 640.0 & 809.1 & 703.1 \\
 & $\dot{y}$ & \textbf{100.0} & \textbf{100.0} & 96.2 & \textbf{100.0} & 96.2 & \textbf{100.0} & 62.0 & 96.2 & 0.0 & 0.0 & 613.8 & 600.1 & 624.7 & 689.3 & 624.0 & 661.9 & 769.2 & 627.1 & 800.9 & 622.2 \\
 & $\dot{z}$ & \textbf{100.0} & 66.7 & 0.0 & 10.5 & 0.0 & 56.0 & 0.0 & 0.0 & 0.0 & 0.0 & 615.4 & 600.1 & 686.0 & 1129.9 & 631.0 & 707.1 & 811.6 & 630.7 & 831.3 & 644.0 \\
\midrule
\multirow{3}{*}{NewtonLiepnik} & $\dot{x}$ & \textbf{100.0} & \textbf{100.0} & 39.8 & 68.4 & 33.3 & \textbf{100.0} & \textbf{100.0} & 0.0 & 0.0 & 0.0 & 623.9 & 600.1 & 762.8 & 871.7 & 625.1 & 718.3 & 821.6 & 632.7 & 897.3 & 647.3 \\
 & $\dot{y}$ & 81.8 & \textbf{100.0} & 8.3 & \textbf{100.0} & 0.0 & 0.0 & 0.0 & 0.0 & 0.0 & 0.0 & 617.6 & 600.0 & 819.2 & 785.5 & 634.1 & 705.8 & 829.5 & 621.8 & 881.8 & 718.5 \\
 & $\dot{z}$ & 90.9 & \textbf{100.0} & 14.0 & \textbf{100.0} & 28.2 & 0.0 & 0.0 & 8.3 & 0.0 & 0.0 & 616.7 & 600.1 & 749.3 & 773.9 & 619.6 & 681.4 & 781.2 & 640.2 & 912.4 & 620.2 \\
\midrule
\multirow{3}{*}{DequanLi} & $\dot{x}$ & 80.0 & \textbf{100.0} & 8.3 & 22.2 & 12.2 & 14.0 & 2.0 & 0.0 & 0.0 & 0.0 & 615.9 & 600.1 & 779.7 & 731.1 & 552.8 & 688.5 & 795.7 & 618.9 & 562.0 & 726.9 \\
 & $\dot{y}$ & \textbf{100.0} & \textbf{100.0} & 20.7 & 94.4 & 66.3 & 88.0 & 4.0 & 96.2 & 0.0 & 0.0 & 615.6 & 600.1 & 797.8 & 666.7 & 567.2 & 672.9 & 765.9 & 629.9 & 599.2 & 709.2 \\
 & $\dot{z}$ & 70.0 & \textbf{100.0} & 0.0 & \textbf{100.0} & 71.7 & 66.0 & 6.0 & 86.3 & 0.0 & 0.0 & 616.8 & 600.0 & 722.1 & 680.7 & 513.6 & 704.9 & 779.1 & 635.3 & 606.1 & 724.7 \\
\midrule
\multirow{3}{*}{Finance} & $\dot{x}$ & \textbf{100.0} & \textbf{100.0} & 33.3 & 83.3 & 50.0 & \textbf{100.0} & 0.0 & 33.2 & 0.0 & 0.0 & 621.6 & 600.1 & 789.8 & 933.6 & 626.3 & 698.4 & 813.7 & 632.0 & 710.6 & 639.3 \\
 & $\dot{y}$ & 80.0 & 11.1 & 0.0 & \textbf{100.0} & 33.2 & 0.0 & 0.0 & 23.4 & 0.0 & 0.0 & 612.3 & 600.1 & 810.3 & 947.7 & 626.7 & 661.5 & 755.3 & 627.4 & 741.4 & 682.1 \\
 & $\dot{z}$ & 80.0 & \textbf{100.0} & 12.0 & \textbf{100.0} & 96.2 & 60.0 & 12.0 & 61.5 & 0.0 & 0.0 & 614.4 & 600.1 & 669.2 & 798.2 & 624.2 & 713.0 & 835.2 & 591.6 & 700.1 & 687.9 \\
\midrule
\multirow{3}{*}{Duffing} & $\dot{x}$ & \textbf{100.0} & \textbf{100.0} & 66.6 & \textbf{100.0} & 96.2 & \textbf{100.0} & 86.0 & 96.2 & 46.5 & 0.0 & 619.0 & 600.1 & 630.3 & 745.2 & 627.6 & 695.2 & 810.5 & 634.3 & 680.6 & 686.2 \\
 & $\dot{y}$ & \textbf{100.0} & 22.2 & 71.7 & 31.6 & 0.0 & 84.0 & 0.0 & 0.0 & 0.0 & 0.0 & 621.6 & 600.0 & 812.1 & 857.4 & 633.6 & 693.7 & 763.3 & 641.3 & 717.2 & 615.6 \\
 & $\dot{z}$ & \textbf{100.0} & 55.6 & 58.6 & \textbf{100.0} & 46.1 & 2.0 & 48.0 & 39.8 & 20.7 & 0.0 & 623.7 & 600.0 & 628.6 & 981.9 & 595.6 & 652.7 & 770.4 & 596.9 & 668.0 & 726.4 \\
\midrule
\multirow{3}{*}{Laser} & $\dot{x}$ & \textbf{100.0} & 22.2 & 28.2 & 36.8 & 0.0 & 74.0 & 90.0 & 54.5 & 0.0 & 0.0 & 617.6 & 600.1 & 834.8 & 829.0 & 621.4 & 708.4 & 809.9 & 627.9 & 733.1 & 745.3 \\
 & $\dot{y}$ & 90.0 & 44.4 & 0.0 & 94.7 & 39.8 & 56.0 & 94.0 & 96.2 & 0.0 & 0.0 & 611.0 & 600.1 & 825.0 & 729.0 & 638.5 & 694.6 & 789.9 & 638.3 & 726.1 & 687.8 \\
 & $\dot{z}$ & \textbf{100.0} & \textbf{100.0} & 23.4 & \textbf{100.0} & 23.9 & 28.0 & 0.0 & 86.3 & 0.0 & 0.0 & 617.7 & 600.1 & 799.0 & 777.2 & 620.3 & 725.1 & 764.5 & 635.4 & 745.7 & 613.5 \\
\midrule
\multirow{3}{*}{Hadley} & $\dot{x}$ & \textbf{100.0} & 25.0 & 18.6 & 44.4 & 0.0 & 0.0 & 0.0 & 0.0 & 0.0 & 0.0 & 611.8 & 600.1 & 818.3 & 991.4 & 638.9 & 650.4 & 773.0 & 626.9 & 647.2 & 632.6 \\
 & $\dot{y}$ & \textbf{100.0} & \textbf{100.0} & 28.2 & \textbf{100.0} & 0.0 & 0.0 & 0.0 & 0.0 & 0.0 & 0.0 & 609.6 & 600.0 & 802.4 & 822.1 & 632.5 & 658.2 & 751.3 & 633.8 & 660.6 & 724.0 \\
 & $\dot{z}$ & \textbf{100.0} & \textbf{100.0} & 8.3 & \textbf{100.0} & 71.7 & 0.0 & 0.0 & 14.0 & 0.0 & 0.0 & 615.3 & 600.1 & 754.4 & 798.6 & 640.3 & 691.1 & 825.3 & 629.6 & 682.4 & 676.9 \\
\midrule
\multirow{3}{*}{Brusselator} & $\dot{x}$ & 50.0 & 88.9 & 0.0 & 0.0 & 0.0 & 4.0 & 0.0 & 0.0 & 0.0 & 0.0 & 633.4 & 600.1 & 818.4 & 1171.1 & 648.3 & 666.9 & 791.8 & 610.9 & 836.8 & 643.8 \\
 & $\dot{y}$ & \textbf{100.0} & \textbf{100.0} & 0.0 & 94.7 & 18.6 & 20.0 & 0.0 & 0.0 & 0.0 & 0.0 & 624.6 & 600.1 & 817.4 & 906.7 & 631.3 & 653.0 & 785.1 & 616.0 & 824.8 & 674.6 \\
 & $\dot{z}$ & \textbf{100.0} & 22.2 & 23.4 & \textbf{100.0} & 63.2 & 68.0 & 12.0 & 28.2 & 0.0 & 0.0 & 620.7 & 600.0 & 636.4 & 1097.9 & 620.4 & 678.6 & 793.3 & 625.7 & 839.7 & 686.9 \\
\midrule
\multirow{3}{*}{FitzHughNagumo} & $\dot{x}$ & \textbf{100.0} & 0.0 & 0.0 & 0.0 & 0.0 & 0.0 & 0.0 & 0.0 & 0.0 & 0.0 & 617.4 & 600.0 & 760.2 & 1172.6 & 624.0 & 667.7 & 816.8 & 1741.3 & 603.5 & 665.1 \\
 & $\dot{y}$ & \textbf{100.0} & \textbf{100.0} & 8.3 & \textbf{100.0} & 83.9 & 0.0 & 0.0 & 8.3 & 0.0 & 0.0 & 624.4 & 600.0 & 789.7 & 890.9 & 637.9 & 686.2 & 823.9 & 603.6 & 601.4 & 606.0 \\
 & $\dot{z}$ & \textbf{100.0} & \textbf{100.0} & 70.7 & \textbf{100.0} & 61.5 & 62.0 & 72.0 & 66.6 & 23.4 & 35.0 & 610.6 & 600.0 & 621.5 & 666.5 & 619.6 & 705.3 & 817.5 & 622.3 & 609.6 & 661.9 \\
\midrule
\multirow{3}{*}{KawczynskiStrizhak} & $\dot{x}$ & 60.0 & 0.0 & 0.0 & 0.0 & 0.0 & 2.0 & 0.0 & 0.0 & 0.0 & 0.0 & 625.5 & 600.1 & 827.4 & 1125.9 & 635.4 & 696.3 & 799.5 & 642.1 & 782.8 & 667.6 \\
 & $\dot{y}$ & \textbf{100.0} & \textbf{100.0} & 46.5 & \textbf{100.0} & 46.5 & 0.0 & 0.0 & 20.7 & 0.0 & 0.0 & 612.0 & 600.0 & 822.4 & 831.5 & 630.9 & 707.7 & 820.3 & 640.2 & 794.8 & 656.7 \\
 & $\dot{z}$ & 70.0 & \textbf{100.0} & 70.7 & \textbf{100.0} & 96.2 & 2.0 & 0.0 & 96.2 & 0.0 & 0.0 & 624.9 & 600.0 & 702.7 & 809.8 & 626.9 & 699.8 & 792.7 & 643.8 & 775.0 & 675.4 \\
\midrule
\bfseries Average 95\% CI & & \cellcolor[RGB]{211,169,206}\makecell{\bfseries 83.5 \\ \bfseries $\pm$ 8.1} & \cellcolor[RGB]{211,169,206}\makecell{\bfseries 59.2 \\ \bfseries $\pm$ 22.5} & \cellcolor[RGB]{211,169,206}\makecell{\bfseries 11.7 \\ \bfseries $\pm$ 11.1} & \cellcolor[RGB]{211,169,206}\makecell{\bfseries 42.1 \\ \bfseries $\pm$ 18.7} & \cellcolor[RGB]{211,169,206}\makecell{\bfseries 5.8 \\ \bfseries $\pm$ 5.9} & \cellcolor[RGB]{211,169,206}\makecell{\bfseries 18.4 \\ \bfseries $\pm$ 14.7} & \cellcolor[RGB]{211,169,206}\makecell{\bfseries 9.5 \\ \bfseries $\pm$ 13.7} & \cellcolor[RGB]{211,169,206}\makecell{\bfseries 8.4 \\ \bfseries $\pm$ 8.7} & \cellcolor[RGB]{211,169,206}\makecell{\bfseries 0.0 \\ \bfseries $\pm$ 0.0} & \cellcolor[RGB]{211,169,206}\makecell{\bfseries 0.0 \\ \bfseries $\pm$ 0.0} & \cellcolor[RGB]{102,204,255}\makecell{\bfseries 616.73 \\ \bfseries $\pm$ 1.50} & \cellcolor[RGB]{102,204,255}\makecell{\bfseries 600.05 \\ \bfseries $\pm$ 0.00} & \cellcolor[RGB]{102,204,255}\makecell{\bfseries 741.56 \\ \bfseries $\pm$ 20.95} & \cellcolor[RGB]{102,204,255}\makecell{\bfseries 820.62 \\ \bfseries $\pm$ 41.28} & \cellcolor[RGB]{102,204,255}\makecell{\bfseries 590.62 \\ \bfseries $\pm$ 15.62} & \cellcolor[RGB]{102,204,255}\makecell{\bfseries 694.91 \\ \bfseries $\pm$ 6.23} & \cellcolor[RGB]{102,204,255}\makecell{\bfseries 799.98 \\ \bfseries $\pm$ 6.65} & \cellcolor[RGB]{102,204,255}\makecell{\bfseries 650.57 \\ \bfseries $\pm$ 43.94} & \cellcolor[RGB]{102,204,255}\makecell{\bfseries 687.25 \\ \bfseries $\pm$ 31.43} & \cellcolor[RGB]{102,204,255}\makecell{\bfseries 660.08 \\ \bfseries $\pm$ 11.12} \\
\bottomrule
\end{tabular}}
\label{table:cs_result_0_02}
\end{table}


\begin{table}[b!]
\caption{Recovery rate and time cost on chaotic dynamics task (5.0\% noise)
}
\centering

\resizebox{\textwidth}{!}{
\begin{tabular}{llcccccccccccccccccccc}
\toprule
\multirow{2}{*}{Chaotic System} & \multirow{2}{*}{Variant} & \multicolumn{10}{c}{Recovery Rate (\%)} & \multicolumn{10}{c}{Avg. Time Cost (s)}\\
\cmidrule(lr){3-12} \cmidrule(lr){13-22}
& & PSE & Operon & BMS & PySR & NGGP & wAIC1 & wAIC2 & DGSR & uDSR & TPSR & PSE & Operon & BMS & PySR & NGGP & wAIC1 & wAIC2 & DGSR & uDSR & TPSR \\
\midrule
\multirow{3}{*}{ShimizuMorioka} & $\dot{x}$ & 77.8 & \textbf{100.0} & 93.1 & \textbf{100.0} & 87.6 & 52.0 & 50.0 & 87.6 & 67.4 & 0.0 & 617.4 & 600.0 & 619.6 & 718.9 & 629.0 & 712.3 & 786.7 & 633.4 & 752.0 & 572.5 \\
 & $\dot{y}$ & \textbf{100.0} & \textbf{100.0} & 0.0 & \textbf{100.0} & 22.2 & \textbf{100.0} & 0.0 & 22.2 & 0.0 & 0.0 & 614.3 & 600.1 & 747.0 & 916.2 & 643.3 & 728.7 & 795.0 & 632.2 & 771.3 & 657.6 \\
 & $\dot{z}$ & 77.8 & 88.9 & 0.0 & \textbf{100.0} & 67.4 & 10.0 & 0.0 & 23.1 & 0.0 & 0.0 & 619.8 & 600.0 & 773.4 & 877.4 & 635.5 & 717.5 & 820.1 & 623.7 & 761.8 & 671.0 \\
\midrule
\multirow{3}{*}{Rucklidge} & $\dot{x}$ & \textbf{100.0} & \textbf{100.0} & 0.0 & 52.6 & 22.2 & 98.0 & 64.0 & 55.2 & 0.0 & 0.0 & 615.4 & 600.1 & 680.9 & 867.9 & 568.5 & 708.2 & 826.1 & 638.1 & 669.4 & 629.7 \\
 & $\dot{y}$ & \textbf{100.0} & \textbf{100.0} & 87.6 & \textbf{100.0} & 87.6 & 80.0 & 68.0 & 87.6 & 0.0 & 0.0 & 619.7 & 600.1 & 619.6 & 682.3 & 554.9 & 701.6 & 805.0 & 637.0 & 698.3 & 626.1 \\
 & $\dot{z}$ & \textbf{100.0} & \textbf{100.0} & 0.0 & \textbf{100.0} & 23.1 & 44.0 & 34.0 & 87.6 & 0.0 & 0.0 & 613.6 & 600.1 & 741.8 & 711.6 & 568.3 & 719.8 & 824.1 & 633.3 & 698.0 & 647.9 \\
\midrule
\multirow{3}{*}{SprottJerk} & $\dot{x}$ & 75.0 & \textbf{100.0} & 53.4 & \textbf{100.0} & 87.6 & 78.0 & 32.0 & 87.6 & 87.6 & 0.0 & 618.4 & 600.1 & 619.4 & 677.6 & 565.2 & 692.8 & 801.6 & 641.6 & 702.3 & 653.5 \\
 & $\dot{y}$ & \textbf{100.0} & \textbf{100.0} & 93.1 & \textbf{100.0} & 87.6 & 78.0 & 44.0 & 87.6 & 0.0 & 0.0 & 614.5 & 600.1 & 622.8 & 661.2 & 605.7 & 701.2 & 841.8 & 623.0 & 729.2 & 637.6 \\
 & $\dot{z}$ & 87.5 & 87.5 & 31.3 & 0.0 & 0.0 & 0.0 & 26.0 & 0.0 & 0.0 & 0.0 & 611.8 & 600.0 & 771.5 & 1121.2 & 623.9 & 722.9 & 849.9 & 629.4 & 769.8 & 653.9 \\
\midrule
\multirow{4}{*}{HyperLorenz} & $\dot{w}$ & 88.9 & \textbf{100.0} & 22.2 & \textbf{100.0} & 0.0 & 24.0 & 70.0 & 0.0 & 0.0 & 0.0 & 617.7 & 600.0 & 755.7 & 751.8 & 521.6 & 686.1 & 788.7 & 637.7 & 526.5 & 634.1 \\
 & $\dot{x}$ & 88.9 & \textbf{100.0} & 0.0 & \textbf{100.0} & 0.0 & 96.0 & 78.0 & 0.0 & 0.0 & 0.0 & 622.2 & 600.1 & 771.5 & 696.2 & 517.2 & 692.1 & 800.1 & 615.4 & 549.3 & 644.7 \\
 & $\dot{y}$ & \textbf{100.0} & \textbf{100.0} & 55.2 & 89.5 & 22.2 & 98.0 & 0.0 & 27.6 & 0.0 & 0.0 & 616.0 & 600.1 & 821.5 & 743.5 & 572.0 & 704.1 & 780.8 & 614.7 & 539.8 & 638.1 \\
 & $\dot{z}$ & \textbf{100.0} & \textbf{100.0} & 0.0 & \textbf{100.0} & 67.4 & 72.0 & 10.0 & 87.6 & 0.0 & 40.0 & 616.4 & 600.0 & 822.5 & 681.4 & 533.2 & 668.8 & 794.6 & 632.8 & 506.3 & 637.6 \\
\midrule
\multirow{4}{*}{HyperJha} & $\dot{w}$ & 88.9 & \textbf{100.0} & 19.9 & \textbf{100.0} & 0.0 & 56.0 & 2.0 & 0.0 & 0.0 & 0.0 & 614.0 & 600.1 & 668.9 & 758.3 & 442.3 & 691.3 & 803.1 & 642.3 & 498.8 & 567.2 \\
 & $\dot{x}$ & 88.9 & \textbf{100.0} & 51.7 & \textbf{100.0} & 0.0 & 90.0 & 2.0 & 0.0 & 0.0 & 0.0 & 614.0 & 600.0 & 676.2 & 690.9 & 451.2 & 696.7 & 816.9 & 612.6 & 559.8 & 667.1 \\
 & $\dot{y}$ & 77.8 & 88.9 & 0.0 & 73.7 & 22.2 & 8.0 & 0.0 & 0.0 & 0.0 & 0.0 & 611.0 & 600.1 & 733.4 & 803.1 & 488.4 & 724.5 & 798.3 & 626.6 & 543.3 & 677.1 \\
 & $\dot{z}$ & \textbf{100.0} & \textbf{100.0} & 19.9 & \textbf{100.0} & 81.2 & 86.0 & 0.0 & 87.6 & 0.0 & 0.0 & 609.6 & 600.1 & 777.5 & 691.2 & 460.8 & 688.3 & 792.4 & 634.1 & 485.1 & 624.5 \\
\midrule
\multirow{4}{*}{HyperPang} & $\dot{w}$ & \textbf{100.0} & \textbf{100.0} & 51.7 & \textbf{100.0} & 87.6 & 52.0 & 98.0 & 87.6 & 0.0 & 0.0 & 616.7 & 600.0 & 637.8 & 749.6 & 591.3 & 706.7 & 785.9 & 637.5 & 573.3 & 583.0 \\
 & $\dot{x}$ & \textbf{100.0} & \textbf{100.0} & 72.0 & \textbf{100.0} & 81.7 & 54.0 & 12.0 & 51.3 & 0.0 & 0.0 & 613.5 & 600.1 & 737.7 & 737.4 & 516.4 & 725.4 & 833.5 & 626.7 & 600.7 & 696.1 \\
 & $\dot{y}$ & 88.9 & \textbf{100.0} & 19.2 & 57.9 & 0.0 & 84.0 & 28.0 & 0.0 & 0.0 & 0.0 & 607.3 & 600.1 & 814.3 & 738.7 & 552.3 & 717.5 & 795.3 & 625.2 & 556.8 & 693.2 \\
 & $\dot{z}$ & 88.9 & 88.9 & 72.0 & \textbf{100.0} & 52.8 & 0.0 & 0.0 & 87.6 & 0.0 & 0.0 & 620.0 & 600.1 & 832.4 & 669.3 & 528.4 & 706.8 & 790.6 & 632.0 & 562.7 & 617.5 \\
\midrule
\multirow{3}{*}{GenesioTesi} & $\dot{x}$ & 88.9 & \textbf{100.0} & 20.5 & \textbf{100.0} & 87.6 & 68.0 & 60.0 & 87.6 & 51.3 & 0.0 & 615.4 & 600.0 & 620.2 & 693.1 & 625.5 & 664.3 & 763.7 & 641.8 & 813.8 & 630.3 \\
 & $\dot{y}$ & 88.9 & \textbf{100.0} & 87.6 & \textbf{100.0} & 87.6 & 66.0 & 40.0 & 87.6 & 0.0 & 0.0 & 615.2 & 600.1 & 620.6 & 698.4 & 622.8 & 679.9 & 779.4 & 636.8 & 796.8 & 628.3 \\
 & $\dot{z}$ & \textbf{100.0} & 44.4 & 0.0 & 10.5 & 0.0 & 10.0 & 0.0 & 0.0 & 0.0 & 0.0 & 610.6 & 600.0 & 699.2 & 1148.9 & 633.5 & 713.8 & 795.1 & 642.0 & 837.0 & 666.6 \\
\midrule
\multirow{3}{*}{NewtonLiepnik} & $\dot{x}$ & \textbf{100.0} & 88.9 & 51.7 & 57.9 & 67.5 & 84.0 & 70.0 & 0.0 & 0.0 & 0.0 & 608.8 & 600.0 & 776.0 & 877.2 & 612.4 & 725.4 & 821.5 & 630.3 & 902.4 & 586.7 \\
 & $\dot{y}$ & \textbf{100.0} & \textbf{100.0} & 0.0 & \textbf{100.0} & 0.0 & 0.0 & 0.0 & 0.0 & 0.0 & 0.0 & 613.8 & 600.0 & 802.0 & 791.6 & 621.5 & 717.5 & 830.0 & 634.3 & 874.3 & 674.4 \\
 & $\dot{z}$ & 77.8 & \textbf{100.0} & 17.9 & \textbf{100.0} & 18.6 & 0.0 & 0.0 & 0.0 & 0.0 & 0.0 & 611.7 & 600.1 & 752.6 & 768.8 & 627.9 & 677.6 & 794.7 & 644.0 & 907.8 & 602.4 \\
\midrule
\multirow{3}{*}{DequanLi} & $\dot{x}$ & \textbf{100.0} & \textbf{100.0} & 0.0 & 16.7 & 0.0 & 18.0 & 0.0 & 0.0 & 0.0 & 0.0 & 610.6 & 600.0 & 763.9 & 748.0 & 549.1 & 688.3 & 789.8 & 627.3 & 562.8 & 649.3 \\
 & $\dot{y}$ & \textbf{100.0} & \textbf{100.0} & 19.2 & \textbf{100.0} & 53.8 & 76.0 & 0.0 & 87.6 & 0.0 & 0.0 & 621.5 & 600.0 & 804.0 & 665.7 & 570.6 & 669.9 & 758.7 & 638.0 & 592.3 & 665.2 \\
 & $\dot{z}$ & 87.5 & \textbf{100.0} & 0.0 & \textbf{100.0} & 64.5 & 24.0 & 0.0 & 51.3 & 0.0 & 0.0 & 613.1 & 600.0 & 725.0 & 683.3 & 509.0 & 706.1 & 794.3 & 623.3 & 604.8 & 740.4 \\
\midrule
\multirow{3}{*}{Finance} & $\dot{x}$ & \textbf{100.0} & \textbf{100.0} & 67.5 & 66.7 & 52.8 & 94.0 & 0.0 & 31.9 & 0.0 & 0.0 & 615.1 & 600.1 & 783.9 & 927.2 & 629.9 & 714.5 & 825.6 & 638.0 & 713.9 & 631.5 \\
 & $\dot{y}$ & 25.0 & 0.0 & 0.0 & \textbf{100.0} & 55.2 & 0.0 & 0.0 & 22.2 & 0.0 & 0.0 & 610.2 & 600.1 & 813.4 & 953.3 & 622.6 & 660.6 & 762.6 & 634.1 & 727.8 & 669.3 \\
 & $\dot{z}$ & 87.5 & \textbf{100.0} & 8.4 & \textbf{100.0} & 87.6 & 0.0 & 0.0 & 20.5 & 0.0 & 0.0 & 616.6 & 600.0 & 667.9 & 791.1 & 634.4 & 715.8 & 842.7 & 596.6 & 705.9 & 588.4 \\
\midrule
\multirow{3}{*}{Duffing} & $\dot{x}$ & \textbf{100.0} & \textbf{100.0} & 53.4 & \textbf{100.0} & 87.6 & 80.0 & 74.0 & 87.6 & 15.6 & 0.0 & 621.2 & 600.0 & 626.3 & 757.2 & 636.0 & 693.6 & 812.8 & 630.4 & 679.8 & 648.3 \\
 & $\dot{y}$ & \textbf{100.0} & 0.0 & 64.5 & 31.6 & 0.0 & 38.0 & 0.0 & 0.0 & 0.0 & 0.0 & 620.8 & 600.0 & 823.4 & 856.7 & 639.1 & 701.7 & 762.6 & 631.2 & 710.2 & 611.8 \\
 & $\dot{z}$ & \textbf{100.0} & 66.7 & 53.0 & \textbf{100.0} & 23.1 & 6.0 & 50.0 & 51.7 & 19.2 & 0.0 & 616.0 & 600.1 & 621.6 & 972.8 & 596.9 & 638.8 & 755.5 & 598.2 & 683.0 & 636.2 \\
\midrule
\multirow{3}{*}{Laser} & $\dot{x}$ & \textbf{100.0} & 55.6 & 18.6 & 21.1 & 0.0 & 68.0 & 32.0 & 72.0 & 0.0 & 0.0 & 618.4 & 600.0 & 821.9 & 819.7 & 623.3 & 714.9 & 803.2 & 631.5 & 714.7 & 655.7 \\
 & $\dot{y}$ & \textbf{100.0} & 55.6 & 0.0 & \textbf{100.0} & 51.7 & 14.0 & 38.0 & 87.6 & 0.0 & 0.0 & 614.8 & 600.1 & 826.6 & 719.0 & 636.6 & 688.4 & 791.6 & 628.7 & 744.3 & 637.2 \\
 & $\dot{z}$ & \textbf{100.0} & \textbf{100.0} & 22.2 & 94.7 & 19.9 & 14.0 & 0.0 & 51.3 & 0.0 & 0.0 & 613.4 & 600.1 & 798.8 & 763.3 & 628.8 & 726.5 & 771.8 & 640.4 & 742.3 & 677.7 \\
\midrule
\multirow{3}{*}{Hadley} & $\dot{x}$ & \textbf{100.0} & 25.0 & 27.6 & 11.1 & 0.0 & 0.0 & 0.0 & 0.0 & 0.0 & 0.0 & 612.6 & 600.1 & 824.7 & 989.3 & 637.1 & 661.5 & 768.8 & 625.3 & 641.2 & 603.8 \\
 & $\dot{y}$ & \textbf{100.0} & \textbf{100.0} & 18.6 & \textbf{100.0} & 0.0 & 0.0 & 0.0 & 0.0 & 0.0 & 0.0 & 620.1 & 600.1 & 793.2 & 814.5 & 626.6 & 676.1 & 741.9 & 627.7 & 664.9 & 681.2 \\
 & $\dot{z}$ & \textbf{100.0} & \textbf{100.0} & 0.0 & \textbf{100.0} & 64.5 & 0.0 & 0.0 & 17.9 & 0.0 & 0.0 & 612.0 & 600.1 & 756.8 & 812.3 & 635.6 & 690.5 & 826.9 & 636.2 & 676.8 & 709.8 \\
\midrule
\multirow{3}{*}{Brusselator} & $\dot{x}$ & 55.6 & 44.4 & 0.0 & 0.0 & 0.0 & 16.0 & 0.0 & 0.0 & 0.0 & 0.0 & 630.9 & 600.0 & 816.8 & 1156.9 & 649.1 & 659.7 & 812.5 & 609.1 & 823.3 & 652.5 \\
 & $\dot{y}$ & \textbf{100.0} & 33.3 & 0.0 & 94.7 & 27.6 & 36.0 & 0.0 & 0.0 & 0.0 & 0.0 & 617.6 & 600.1 & 807.1 & 904.6 & 635.0 & 644.1 & 782.3 & 611.4 & 847.3 & 656.3 \\
 & $\dot{z}$ & \textbf{100.0} & 22.2 & 22.2 & \textbf{100.0} & 42.6 & 58.0 & 2.0 & 18.6 & 0.0 & 0.0 & 625.1 & 600.1 & 636.8 & 1096.7 & 617.5 & 673.8 & 791.4 & 632.8 & 848.6 & 659.3 \\
\midrule
\multirow{3}{*}{FitzHughNagumo} & $\dot{x}$ & 87.5 & 0.0 & 0.0 & 0.0 & 0.0 & 0.0 & 0.0 & 0.0 & 0.0 & 0.0 & 607.3 & 600.0 & 781.6 & 1175.0 & 620.6 & 665.7 & 803.0 & 1738.1 & 594.8 & 626.1 \\
 & $\dot{y}$ & \textbf{100.0} & \textbf{100.0} & 0.0 & \textbf{100.0} & 38.5 & 2.0 & 0.0 & 0.0 & 0.0 & 0.0 & 623.9 & 600.1 & 776.7 & 875.5 & 640.6 & 668.7 & 809.7 & 605.6 & 588.2 & 640.5 \\
 & $\dot{z}$ & \textbf{100.0} & \textbf{100.0} & 67.4 & \textbf{100.0} & 20.5 & 60.0 & 68.0 & 53.4 & 22.2 & 35.0 & 622.4 & 600.1 & 619.7 & 652.6 & 606.4 & 722.5 & 814.8 & 618.9 & 606.7 & 593.0 \\
\midrule
\multirow{3}{*}{KawczynskiStrizhak} & $\dot{x}$ & 55.6 & 0.0 & 0.0 & 0.0 & 0.0 & 0.0 & 0.0 & 0.0 & 0.0 & 0.0 & 627.1 & 600.0 & 819.2 & 1117.7 & 628.4 & 689.7 & 800.0 & 629.7 & 772.5 & 591.3 \\
 & $\dot{y}$ & 55.6 & \textbf{100.0} & 15.6 & 0.0 & 15.6 & 0.0 & 0.0 & 19.2 & 0.0 & 0.0 & 623.6 & 600.0 & 807.1 & 839.5 & 613.5 & 706.1 & 811.8 & 637.2 & 792.3 & 670.8 \\
 & $\dot{z}$ & 88.9 & \textbf{100.0} & 67.4 & \textbf{100.0} & 87.6 & 34.0 & 0.0 & 87.6 & 0.0 & 0.0 & 622.0 & 600.1 & 704.8 & 802.5 & 637.3 & 701.0 & 785.1 & 629.7 & 792.4 & 627.0 \\
\midrule
\bfseries Average 95\% CI & & \cellcolor[RGB]{211,169,206}\makecell{\bfseries 80.4 \\ \bfseries $\pm$ 10.8} & \cellcolor[RGB]{211,169,206}\makecell{\bfseries 55.6 \\ \bfseries $\pm$ 22.1} & \cellcolor[RGB]{211,169,206}\makecell{\bfseries 6.5 \\ \bfseries $\pm$ 8.1} & \cellcolor[RGB]{211,169,206}\makecell{\bfseries 36.8 \\ \bfseries $\pm$ 18.3} & \cellcolor[RGB]{211,169,206}\makecell{\bfseries 6.1 \\ \bfseries $\pm$ 7.8} & \cellcolor[RGB]{211,169,206}\makecell{\bfseries 9.4 \\ \bfseries $\pm$ 6.4} & \cellcolor[RGB]{211,169,206}\makecell{\bfseries 3.8 \\ \bfseries $\pm$ 5.5} & \cellcolor[RGB]{211,169,206}\makecell{\bfseries 9.3 \\ \bfseries $\pm$ 9.9} & \cellcolor[RGB]{211,169,206}\makecell{\bfseries 0.0 \\ \bfseries $\pm$ 0.0} & \cellcolor[RGB]{211,169,206}\makecell{\bfseries 0.0 \\ \bfseries $\pm$ 0.0} & \cellcolor[RGB]{102,204,255}\makecell{\bfseries 616.38 \\ \bfseries $\pm$ 1.43} & \cellcolor[RGB]{102,204,255}\makecell{\bfseries 600.05 \\ \bfseries $\pm$ 0.00} & \cellcolor[RGB]{102,204,255}\makecell{\bfseries 739.94 \\ \bfseries $\pm$ 20.98} & \cellcolor[RGB]{102,204,255}\makecell{\bfseries 819.96 \\ \bfseries $\pm$ 40.83} & \cellcolor[RGB]{102,204,255}\makecell{\bfseries 590.92 \\ \bfseries $\pm$ 15.53} & \cellcolor[RGB]{102,204,255}\makecell{\bfseries 695.50 \\ \bfseries $\pm$ 6.42} & \cellcolor[RGB]{102,204,255}\makecell{\bfseries 798.71 \\ \bfseries $\pm$ 6.65} & \cellcolor[RGB]{102,204,255}\makecell{\bfseries 650.54 \\ \bfseries $\pm$ 43.80} & \cellcolor[RGB]{102,204,255}\makecell{\bfseries 686.61 \\ \bfseries $\pm$ 31.84} & \cellcolor[RGB]{102,204,255}\makecell{\bfseries 641.98 \\ \bfseries $\pm$ 9.88} \\
\bottomrule
\end{tabular}}

\label{table:cs_result_0_05}
\end{table}


\begin{table}[htbp]
\caption{Recovery rate and time cost on chaotic dynamics task (10.0\% noise)
}
\centering
\resizebox{\textwidth}{!}{
\begin{tabular}{llcccccccccccccccccccc}
\toprule
\multirow{2}{*}{Chaotic System} & \multirow{2}{*}{Variant} & \multicolumn{10}{c}{Recovery Rate (\%)} & \multicolumn{10}{c}{Avg. Time Cost (s)}\\
\cmidrule(lr){3-12} \cmidrule(lr){13-22}
& & PSE & Operon & BMS & PySR & NGGP & wAIC1 & wAIC2 & DGSR & uDSR & TPSR & PSE & Operon & BMS & PySR & NGGP & wAIC1 & wAIC2 & DGSR & uDSR & TPSR \\
\midrule
\multirow{3}{*}{ShimizuMorioka} & $\dot{x}$ & \textbf{100.0} & \textbf{100.0} & 87.4 & \textbf{100.0} & 75.6 & 28.0 & 16.0 & 75.6 & 75.2 & 0.0 & 621.9 & 600.0 & 616.7 & 707.4 & 624.9 & 695.3 & 776.6 & 629.5 & 757.5 & 683.7 \\
 & $\dot{y}$ & 88.9 & \textbf{100.0} & 0.0 & \textbf{100.0} & 0.0 & 14.0 & 0.0 & 0.0 & 0.0 & 0.0 & 613.9 & 600.1 & 741.9 & 905.4 & 628.6 & 742.7 & 798.6 & 641.3 & 762.8 & 640.0 \\
 & $\dot{z}$ & 77.8 & 11.1 & 0.0 & \textbf{100.0} & 75.2 & 16.0 & 0.0 & 16.6 & 0.0 & 0.0 & 610.3 & 600.1 & 779.5 & 879.4 & 637.0 & 716.8 & 825.5 & 624.4 & 782.2 & 652.0 \\
\midrule
\multirow{3}{*}{Rucklidge} & $\dot{x}$ & \textbf{100.0} & \textbf{100.0} & 0.0 & 72.2 & 25.3 & 74.0 & 12.0 & 32.8 & 0.0 & 0.0 & 617.9 & 600.0 & 684.4 & 869.6 & 559.2 & 698.7 & 828.3 & 624.3 & 676.5 & 604.9 \\
 & $\dot{y}$ & 75.0 & \textbf{100.0} & 75.6 & \textbf{100.0} & 75.6 & 76.0 & 68.0 & 75.6 & 0.0 & 0.0 & 614.2 & 600.1 & 629.8 & 676.3 & 552.9 & 700.0 & 793.7 & 625.2 & 704.6 & 661.2 \\
 & $\dot{z}$ & \textbf{100.0} & \textbf{100.0} & 0.0 & \textbf{100.0} & 16.6 & 12.0 & 6.0 & 75.6 & 0.0 & 0.0 & 610.8 & 600.0 & 757.6 & 711.1 & 568.0 & 735.2 & 815.5 & 629.7 & 697.8 & 610.3 \\
\midrule
\multirow{3}{*}{SprottJerk} & $\dot{x}$ & 87.5 & \textbf{100.0} & 48.2 & \textbf{100.0} & 75.6 & 0.0 & 4.0 & 75.6 & 75.6 & 0.0 & 611.0 & 600.0 & 621.9 & 667.9 & 563.4 & 685.8 & 809.6 & 638.2 & 697.6 & 669.8 \\
 & $\dot{y}$ & 87.5 & \textbf{100.0} & 87.4 & \textbf{100.0} & 75.6 & 2.0 & 20.0 & 75.6 & 0.0 & 0.0 & 620.7 & 600.1 & 618.5 & 661.4 & 619.0 & 712.7 & 839.4 & 617.8 & 734.5 & 675.7 \\
 & $\dot{z}$ & 25.0 & 0.0 & 0.0 & 0.0 & 0.0 & 0.0 & 6.0 & 0.0 & 0.0 & 0.0 & 620.2 & 600.0 & 753.7 & 1139.7 & 630.9 & 719.4 & 837.5 & 624.9 & 773.7 & 587.2 \\
\midrule
\multirow{4}{*}{HyperLorenz} & $\dot{w}$ & 88.9 & \textbf{100.0} & 0.0 & \textbf{100.0} & 0.0 & 28.0 & 48.0 & 0.0 & 0.0 & 0.0 & 612.9 & 600.1 & 752.6 & 760.5 & 516.7 & 698.1 & 779.3 & 623.8 & 537.7 & 664.0 \\
 & $\dot{x}$ & \textbf{100.0} & \textbf{100.0} & 0.0 & \textbf{100.0} & 0.0 & 96.0 & 2.0 & 0.0 & 0.0 & 0.0 & 613.2 & 600.1 & 777.9 & 693.1 & 514.2 & 706.5 & 797.9 & 615.3 & 550.6 & 619.4 \\
 & $\dot{y}$ & 88.9 & 77.8 & 32.8 & 84.2 & 25.3 & 32.0 & 0.0 & 18.3 & 0.0 & 0.0 & 611.0 & 600.1 & 834.8 & 754.5 & 560.9 & 704.0 & 789.8 & 602.9 & 534.4 & 592.5 \\
 & $\dot{z}$ & \textbf{100.0} & \textbf{100.0} & 0.0 & \textbf{100.0} & 75.2 & 30.0 & 2.0 & 75.6 & 0.0 & 40.0 & 614.9 & 600.0 & 818.3 & 688.7 & 541.4 & 656.3 & 790.0 & 627.9 & 515.7 & 704.6 \\
\midrule
\multirow{4}{*}{HyperJha} & $\dot{w}$ & 77.8 & \textbf{100.0} & 44.1 & \textbf{100.0} & 0.0 & 38.0 & 0.0 & 0.0 & 0.0 & 0.0 & 615.2 & 600.1 & 665.7 & 765.8 & 446.4 & 680.1 & 807.3 & 633.1 & 492.2 & 638.8 \\
 & $\dot{x}$ & 88.9 & \textbf{100.0} & 60.2 & \textbf{100.0} & 0.0 & 98.0 & 0.0 & 0.0 & 0.0 & 0.0 & 618.9 & 600.1 & 688.6 & 702.7 & 449.4 & 704.5 & 812.2 & 617.7 & 554.6 & 635.4 \\
 & $\dot{y}$ & 88.9 & 33.3 & 0.0 & 78.9 & 25.3 & 16.0 & 0.0 & 0.0 & 0.0 & 0.0 & 617.5 & 600.1 & 742.5 & 803.5 & 491.7 & 707.2 & 804.8 & 628.4 & 537.8 & 656.3 \\
 & $\dot{z}$ & \textbf{100.0} & \textbf{100.0} & 44.1 & \textbf{100.0} & 87.8 & 22.0 & 0.0 & 75.6 & 0.0 & 0.0 & 613.9 & 600.1 & 766.1 & 698.5 & 459.8 & 679.3 & 798.3 & 624.3 & 482.1 & 673.1 \\
\midrule
\multirow{4}{*}{HyperPang} & $\dot{w}$ & 77.8 & \textbf{100.0} & 60.2 & \textbf{100.0} & 75.6 & 64.0 & 72.0 & 75.6 & 0.0 & 0.0 & 621.2 & 600.0 & 628.1 & 745.7 & 589.2 & 704.0 & 778.7 & 639.1 & 571.2 & 622.4 \\
 & $\dot{x}$ & \textbf{100.0} & \textbf{100.0} & 48.3 & \textbf{100.0} & 43.6 & 64.0 & 2.0 & 21.7 & 0.0 & 0.0 & 628.9 & 600.0 & 731.6 & 744.6 & 518.0 & 727.2 & 831.4 & 636.2 & 591.7 & 623.9 \\
 & $\dot{y}$ & 55.6 & \textbf{100.0} & 30.5 & 31.6 & 0.0 & 92.0 & 2.0 & 0.0 & 0.0 & 0.0 & 614.4 & 600.1 & 826.9 & 732.3 & 551.1 & 703.6 & 808.6 & 639.0 & 558.5 & 629.2 \\
 & $\dot{z}$ & \textbf{100.0} & 88.9 & 48.3 & \textbf{100.0} & 71.1 & 0.0 & 0.0 & 75.6 & 0.0 & 0.0 & 612.0 & 600.1 & 825.3 & 659.9 & 519.5 & 704.0 & 789.5 & 639.2 & 564.3 & 576.3 \\
\midrule
\multirow{3}{*}{GenesioTesi} & $\dot{x}$ & 88.9 & \textbf{100.0} & 0.0 & \textbf{100.0} & 75.6 & 40.0 & 30.0 & 75.6 & 21.7 & 0.0 & 618.4 & 600.0 & 627.4 & 693.1 & 626.3 & 677.1 & 768.3 & 640.3 & 809.0 & 716.4 \\
 & $\dot{y}$ & 77.8 & \textbf{100.0} & 75.6 & \textbf{100.0} & 75.6 & 22.0 & 2.0 & 75.6 & 0.0 & 0.0 & 622.3 & 600.1 & 617.1 & 685.4 & 614.7 & 670.4 & 766.1 & 633.1 & 792.6 & 692.7 \\
 & $\dot{z}$ & 77.8 & 0.0 & 0.0 & 0.0 & 0.0 & 0.0 & 0.0 & 0.0 & 0.0 & 0.0 & 626.3 & 600.1 & 704.3 & 1132.8 & 625.3 & 726.7 & 810.0 & 626.7 & 831.6 & 631.9 \\
\midrule
\multirow{3}{*}{NewtonLiepnik} & $\dot{x}$ & \textbf{100.0} & 77.8 & 60.2 & 42.1 & 29.9 & 34.0 & 34.0 & 0.0 & 0.0 & 0.0 & 613.6 & 600.0 & 772.9 & 869.0 & 623.2 & 720.4 & 831.7 & 635.8 & 897.2 & 642.4 \\
 & $\dot{y}$ & \textbf{100.0} & \textbf{100.0} & 0.0 & \textbf{100.0} & 0.0 & 2.0 & 2.0 & 0.0 & 0.0 & 0.0 & 610.4 & 600.1 & 807.5 & 783.5 & 634.5 & 706.6 & 832.2 & 620.5 & 881.5 & 634.2 \\
 & $\dot{z}$ & \textbf{100.0} & \textbf{100.0} & 15.6 & \textbf{100.0} & 0.0 & 0.0 & 0.0 & 0.0 & 0.0 & 0.0 & 612.0 & 600.1 & 750.9 & 764.0 & 622.3 & 672.8 & 786.8 & 643.3 & 914.6 & 639.3 \\
\midrule
\multirow{3}{*}{DequanLi} & $\dot{x}$ & \textbf{100.0} & \textbf{100.0} & 0.0 & 44.4 & 0.0 & 18.0 & 0.0 & 0.0 & 0.0 & 0.0 & 612.6 & 600.0 & 769.2 & 746.0 & 548.6 & 699.3 & 791.1 & 620.3 & 574.1 & 603.7 \\
 & $\dot{y}$ & \textbf{100.0} & \textbf{100.0} & 30.5 & \textbf{100.0} & 52.3 & 38.0 & 0.0 & 75.6 & 0.0 & 0.0 & 617.8 & 600.1 & 791.2 & 679.3 & 575.4 & 672.9 & 776.8 & 644.0 & 590.1 & 637.5 \\
 & $\dot{z}$ & 50.0 & 62.5 & 0.0 & 88.9 & 17.8 & 4.0 & 0.0 & 21.7 & 0.0 & 0.0 & 616.6 & 600.0 & 733.3 & 694.2 & 515.3 & 703.8 & 791.6 & 626.1 & 600.6 & 643.7 \\
\midrule
\multirow{3}{*}{Finance} & $\dot{x}$ & 87.5 & 0.0 & 29.9 & 77.8 & 71.1 & 4.0 & 0.0 & 21.5 & 0.0 & 0.0 & 614.2 & 600.0 & 788.7 & 926.3 & 628.7 & 712.6 & 833.6 & 632.2 & 713.8 & 679.9 \\
 & $\dot{y}$ & 50.0 & 0.0 & 0.0 & \textbf{100.0} & 32.8 & 0.0 & 0.0 & 25.3 & 0.0 & 0.0 & 607.6 & 600.1 & 808.4 & 961.2 & 625.5 & 674.0 & 750.5 & 628.5 & 732.1 & 602.4 \\
 & $\dot{z}$ & 75.0 & \textbf{100.0} & 0.0 & \textbf{100.0} & 75.6 & 0.0 & 0.0 & 0.0 & 0.0 & 0.0 & 631.7 & 600.1 & 684.3 & 800.8 & 638.0 & 727.2 & 833.3 & 593.9 & 712.6 & 680.9 \\
\midrule
\multirow{3}{*}{Duffing} & $\dot{x}$ & 66.7 & \textbf{100.0} & 48.2 & \textbf{100.0} & 75.6 & 40.0 & 16.0 & 75.6 & 0.0 & 0.0 & 620.2 & 600.0 & 632.7 & 747.1 & 632.0 & 691.2 & 815.1 & 627.2 & 670.9 & 740.3 \\
 & $\dot{y}$ & 77.8 & 0.0 & 17.8 & 21.1 & 0.0 & 2.0 & 0.0 & 0.0 & 0.0 & 0.0 & 620.0 & 600.1 & 816.0 & 874.3 & 633.0 & 702.5 & 764.1 & 642.6 & 719.7 & 621.8 \\
 & $\dot{z}$ & \textbf{100.0} & 44.4 & 0.0 & \textbf{100.0} & 16.6 & 26.0 & 72.0 & 60.2 & 30.5 & 0.0 & 624.1 & 600.1 & 627.0 & 978.3 & 598.1 & 642.2 & 775.4 & 600.2 & 680.1 & 687.7 \\
\midrule
\multirow{3}{*}{Laser} & $\dot{x}$ & 88.9 & 11.1 & 0.0 & 10.5 & 0.0 & 18.0 & 8.0 & 48.3 & 0.0 & 0.0 & 623.9 & 600.1 & 819.1 & 817.5 & 623.3 & 710.9 & 807.4 & 630.3 & 728.0 & 677.9 \\
 & $\dot{y}$ & \textbf{100.0} & 55.6 & 0.0 & \textbf{100.0} & 60.2 & 0.0 & 2.0 & 75.6 & 0.0 & 0.0 & 613.3 & 600.1 & 824.0 & 732.4 & 626.2 & 697.2 & 800.9 & 634.9 & 729.4 & 650.2 \\
 & $\dot{z}$ & \textbf{100.0} & 22.2 & 25.3 & 94.7 & 44.1 & 0.0 & 0.0 & 21.7 & 0.0 & 0.0 & 617.7 & 600.0 & 815.4 & 780.9 & 625.4 & 708.4 & 768.6 & 635.9 & 749.8 & 687.3 \\
\midrule
\multirow{3}{*}{Hadley} & $\dot{x}$ & 87.5 & 0.0 & 18.3 & 0.0 & 0.0 & 0.0 & 0.0 & 0.0 & 0.0 & 0.0 & 609.8 & 600.1 & 828.4 & 978.6 & 629.4 & 659.2 & 774.7 & 633.6 & 654.0 & 622.5 \\
 & $\dot{y}$ & \textbf{100.0} & \textbf{100.0} & 0.0 & 88.9 & 0.0 & 0.0 & 0.0 & 0.0 & 0.0 & 0.0 & 615.2 & 600.1 & 795.4 & 822.4 & 625.3 & 670.7 & 740.7 & 628.4 & 674.0 & 601.4 \\
 & $\dot{z}$ & \textbf{100.0} & \textbf{100.0} & 0.0 & \textbf{100.0} & 17.8 & 2.0 & 0.0 & 15.6 & 0.0 & 0.0 & 618.5 & 600.1 & 770.5 & 813.8 & 638.6 & 708.1 & 824.3 & 632.7 & 688.9 & 632.4 \\
\midrule
\multirow{3}{*}{Brusselator} & $\dot{x}$ & 55.6 & 11.1 & 0.0 & 0.0 & 0.0 & 4.0 & 0.0 & 0.0 & 0.0 & 0.0 & 623.3 & 600.1 & 829.5 & 1160.0 & 641.4 & 659.2 & 805.7 & 610.3 & 831.3 & 602.6 \\
 & $\dot{y}$ & 66.7 & 0.0 & 0.0 & 89.5 & 18.3 & 0.0 & 0.0 & 0.0 & 0.0 & 0.0 & 617.2 & 600.1 & 803.2 & 891.8 & 637.4 & 637.5 & 785.5 & 606.0 & 845.3 & 543.6 \\
 & $\dot{z}$ & \textbf{100.0} & 33.3 & 0.0 & \textbf{100.0} & 0.0 & 60.0 & 64.0 & 0.0 & 0.0 & 0.0 & 619.4 & 600.1 & 623.3 & 1085.3 & 628.9 & 672.7 & 801.9 & 632.3 & 853.3 & 674.3 \\
\midrule
\multirow{3}{*}{FitzHughNagumo} & $\dot{x}$ & 0.0 & 0.0 & 0.0 & 0.0 & 0.0 & 0.0 & 0.0 & 0.0 & 0.0 & 0.0 & 611.9 & 600.0 & 773.6 & 1177.5 & 628.1 & 662.8 & 802.9 & 1756.0 & 606.7 & 579.8 \\
 & $\dot{y}$ & \textbf{100.0} & \textbf{100.0} & 0.0 & \textbf{100.0} & 0.0 & 0.0 & 0.0 & 0.0 & 0.0 & 0.0 & 620.3 & 600.0 & 784.2 & 894.6 & 630.9 & 675.4 & 815.9 & 601.6 & 596.9 & 668.0 \\
 & $\dot{z}$ & \textbf{100.0} & 88.9 & 75.2 & \textbf{100.0} & 0.0 & 74.0 & 76.0 & 48.2 & 0.0 & 35.0 & 614.8 & 600.1 & 622.2 & 660.5 & 612.8 & 713.7 & 801.6 & 619.9 & 620.1 & 622.6 \\
\midrule
\multirow{3}{*}{KawczynskiStrizhak} & $\dot{x}$ & 0.0 & 0.0 & 0.0 & 0.0 & 0.0 & 0.0 & 0.0 & 0.0 & 0.0 & 0.0 & 622.4 & 600.0 & 832.4 & 1121.0 & 623.8 & 704.5 & 795.7 & 625.2 & 784.3 & 630.9 \\
 & $\dot{y}$ & 55.6 & \textbf{100.0} & 0.0 & 11.1 & 0.0 & 0.0 & 0.0 & 30.5 & 0.0 & 0.0 & 618.6 & 600.0 & 807.7 & 818.7 & 614.9 & 714.0 & 825.9 & 632.2 & 799.7 & 630.8 \\
 & $\dot{z}$ & 33.3 & \textbf{100.0} & 75.2 & \textbf{100.0} & 75.6 & 50.0 & 0.0 & 75.6 & 0.0 & 0.0 & 618.4 & 600.0 & 702.8 & 810.2 & 625.1 & 694.8 & 797.6 & 635.5 & 782.9 & 674.8 \\
\midrule
\bfseries Average 95\% CI & & \cellcolor[RGB]{211,169,206}\makecell{\bfseries 61.0 \\ \bfseries $\pm$ 16.2} & \cellcolor[RGB]{211,169,206}\makecell{\bfseries 28.9 \\ \bfseries $\pm$ 20.4} & \cellcolor[RGB]{211,169,206}\makecell{\bfseries 1.9 \\ \bfseries $\pm$ 4.1} & \cellcolor[RGB]{211,169,206}\makecell{\bfseries 35.2 \\ \bfseries $\pm$ 19.5} & \cellcolor[RGB]{211,169,206}\makecell{\bfseries 3.1 \\ \bfseries $\pm$ 4.8} & \cellcolor[RGB]{211,169,206}\makecell{\bfseries 4.8 \\ \bfseries $\pm$ 4.4} & \cellcolor[RGB]{211,169,206}\makecell{\bfseries 0.6 \\ \bfseries $\pm$ 0.9} & \cellcolor[RGB]{211,169,206}\makecell{\bfseries 3.4 \\ \bfseries $\pm$ 5.1} & \cellcolor[RGB]{211,169,206}\makecell{\bfseries 0.0 \\ \bfseries $\pm$ 0.0} & \cellcolor[RGB]{211,169,206}\makecell{\bfseries 0.0 \\ \bfseries $\pm$ 0.0} & \cellcolor[RGB]{102,204,255}\makecell{\bfseries 617.01 \\ \bfseries $\pm$ 1.44} & \cellcolor[RGB]{102,204,255}\makecell{\bfseries 600.05 \\ \bfseries $\pm$ 0.00} & \cellcolor[RGB]{102,204,255}\makecell{\bfseries 741.67 \\ \bfseries $\pm$ 21.07} & \cellcolor[RGB]{102,204,255}\makecell{\bfseries 820.95 \\ \bfseries $\pm$ 40.69} & \cellcolor[RGB]{102,204,255}\makecell{\bfseries 590.09 \\ \bfseries $\pm$ 15.37} & \cellcolor[RGB]{102,204,255}\makecell{\bfseries 695.50 \\ \bfseries $\pm$ 6.57} & \cellcolor[RGB]{102,204,255}\makecell{\bfseries 799.77 \\ \bfseries $\pm$ 6.51} & \cellcolor[RGB]{102,204,255}\makecell{\bfseries 649.25 \\ \bfseries $\pm$ 44.58} & \cellcolor[RGB]{102,204,255}\makecell{\bfseries 689.08 \\ \bfseries $\pm$ 31.95} & \cellcolor[RGB]{102,204,255}\makecell{\bfseries 642.64 \\ \bfseries $\pm$ 10.80} \\
\bottomrule
\end{tabular}}
\label{table:cs_result_0_1}
\end{table}

\clearpage
\section{Additional Comparison}

\begin{figure}[t!]

  \hspace*{-0.0in}
  \centering
\includegraphics[width=1.0\textwidth]{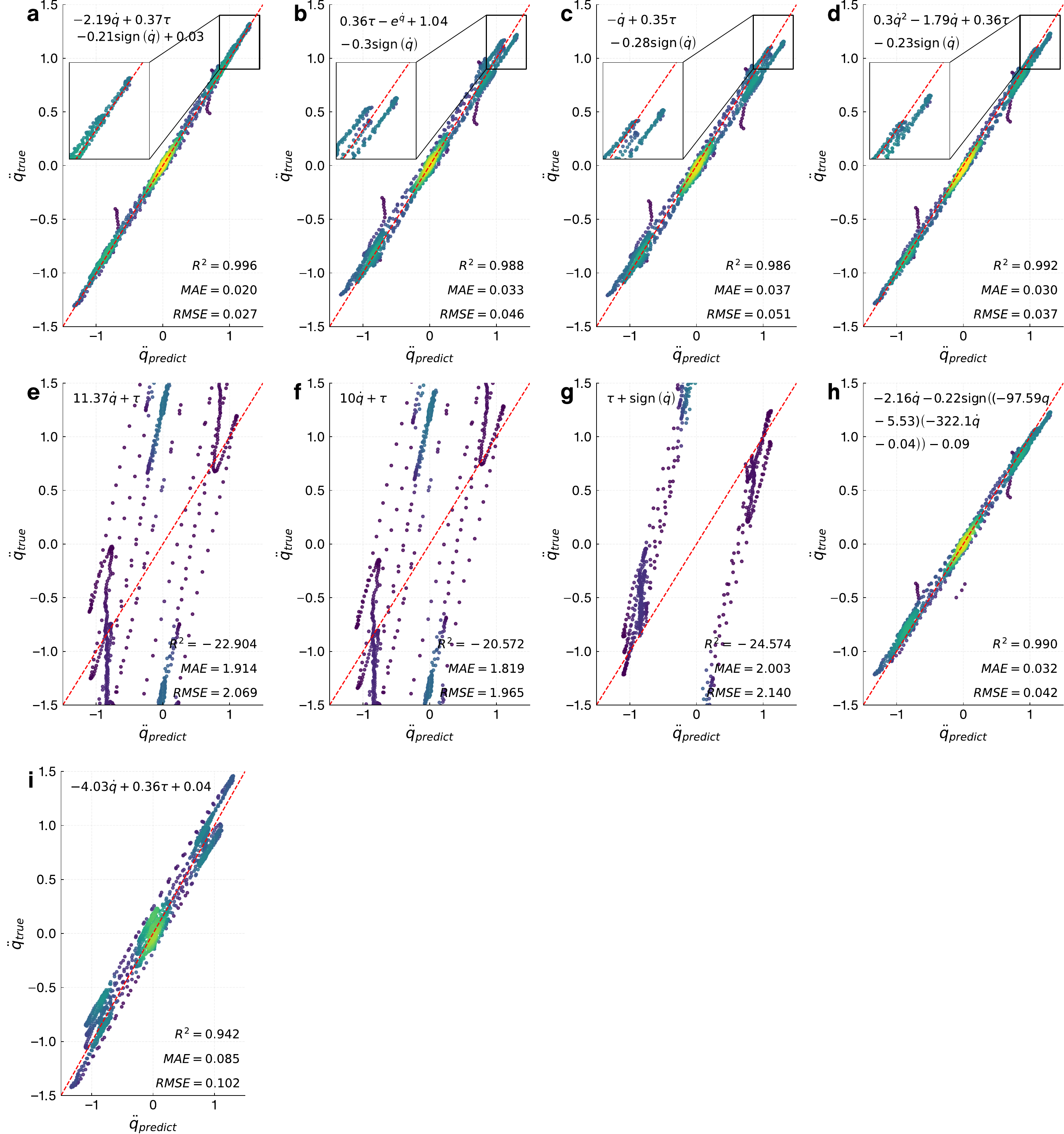}
  \caption{\kredit{The prediction performance along with \hsedit{typical} governing equations of EMPS discovered by: \textbf{a}.~PSE, \textbf{b}.~PySR, \textbf{c}.~BMS, \textbf{d}.~wAIC, \textbf{e}.~NGGP, \textbf{f}.~DGSR, \textbf{g}.~uDSR, \textbf{h}.~TPSR, \textbf{i}.~Operon.}}
\label{fig:EMPS_all}
\end{figure}

\kredit{
In this section, we present additional comparative analyses between our PSE and other methods. Table \ref{tab:sr_gpt_additional_compare} demonstrates comprehensive comparisons with SR-GPT, DGSR-MCTS, TPSR, and NeSymReS, where our approach consistently achieves superior performance. Furthermore, in Fig. \ref{fig:EMPS_all}, we provide an extensive evaluation of EMPS results encompassing multiple baseline methods. The results consistently demonstrate that our PSE outperforms these existing approaches by a significant margin.
}

\begin{table}[htbp]
\kredit{
\caption{\kredit{Additional comparison with SR-GPT \cite{li2024discoveringmathematicalformulasdata}, DGSR-MCTS \cite{kamienny2023deep}, TPSR \cite{Shojaee_reviewer_NIPS2023_TPSR}, and NeSymReS \cite{2021BiggioNeural_NeSymReS}. Except for our method, all data in this table are replicated from \cite{li2024discoveringmathematicalformulasdata}.}}
\label{tab:sr_gpt_additional_compare}
\begin{center}
{\footnotesize
\begin{tabular}{cccHcccc}
\toprule
Dataset & SR-GPT & DGSR-MCTS & SPL & TPSR & NeSymReS & PSE\\
\midrule
Nguyen  & $96_{\pm1.27}\%$ & $93_{\pm2.52}\%$ & $91_{\pm3.46}\%$ & $91_{\pm4.17}\%$ & $56_{\pm3.67}\%$ & $\mathbf{100}_{\pm0.00}\%$ \\
Livermore  & ${91}_{\pm1.07}\%$ & $83_{\pm3.13}\%$ & $81_{\pm2.40}\%$ & $83_{\pm2.17}\%$ & $28_{\pm3.58}\%$ & $\mathbf{100}_{\pm0.00}\%$\\
R  & ${50}_{\pm1.82}\%$ & $45_{\pm1.81}\%$ & $32_{\pm2.61}\%$ & $40_{\pm1.33}\%$ & $4_{\pm3.18}\%$ & $\mathbf{100}_{\pm0.00}\%$\\
AIFeynman  & ${65}_{\pm1.22}\%$ & $57_{\pm2.57}\%$ & $44_{\pm4.10}\%$ & $49_{\pm3.19}\%$ & $18_{\pm4.62}\%$ & $\mathbf{73}_{\pm4.07}\%$\\
\midrule
Average & ${74.90}_{\pm10.62}\%$ & $68.5_{\pm11.1}\%$ & $63.1_{\pm13.4}\%$ & $65.1_{\pm10.8}\%$ & $25.1_{\pm10.74}\%$ & $\mathbf{93.3}_{\pm1.02}\%$ \\
\bottomrule
\end{tabular}}
\end{center}}

\end{table}

\clearpage
\newpage
\section{Algorithm Pseudocode}

The pseudocodes for the proposed method are summarized in Algorithms \ref{alg:psrn_eval_derive}--\ref{alg:random_generator}.

\begin{algorithm}[htbp]
\caption{PSRN Evaluation and Symbolic Derivation}
\label{alg:psrn_eval_derive}
\begin{algorithmic}[1]
\Require PSRN $\psi_{\text{psrn}}$, Leaf symbols $s_0$, Leaf values $\mathbf{X}_0$, Target $\mathbf{y}$, Top count $k$.
\Ensure Top-$k$ expressions $\mathcal{F}$, Raw errors $\mathcal{E}$.

\State \Comment{1. PSRN Forward Pass}
\State $\mathbf{H} \leftarrow \psi_{\text{psrn}}(\mathbf{X}_0)$ \Comment{Numerical outputs from $\psi_{\text{psrn}}$'s final layer, $\mathbf{h}^{(L)}$}

\State \Comment{2. Error Computation}
\State $\mathbf{E} \leftarrow \text{MSE}(\mathbf{H}, \mathbf{y})$ \Comment{Error for each candidate}

\State \Comment{3. Identify Top-$k$ Candidates}
\State $\mathcal{E}, \text{Indices} \leftarrow \text{topk}(\mathbf{E}, k)$

\State \Comment{4. Recursive Symbolic Reconstruction}
\Function{Reconstruct}{$i, l, s_0, \psi_{\text{psrn}}$}
    \If{$l = 0$} \Comment{Base case: leaf layer}
        \State \Return $s_0[i]$ \Comment{Symbolic form of the $i$-th leaf}
    \EndIf
    
    \State $(\text{op}, \text{children\_indices}) \leftarrow {\psi_{\text{psrn}}}_l.\text{structure}(i)$ 
    \Comment{Retrieve operator and child using $\mathbf{\Theta}^{(l)}$}
    
    \State $\text{args} \leftarrow []$
    \For{each $c\_idx$ in $children\_indices$}
        \State Append \Call{Reconstruct}{$c\_idx, l-1, s_0, \psi_{\text{psrn}}$} to args.
    \EndFor   
    \State \Return $\text{op}(\text{args})$ \Comment{Combine arguments symbolically}
\EndFunction

\State $\mathcal{F} \leftarrow \emptyset$
\State $L \leftarrow \psi_{\text{psrn}}.\text{depth}$
\For{each $idx \in \text{Indices}$}
    \State $f \leftarrow \Call{Reconstruct}{idx, L, s_0, \psi_{\text{psrn}}}$
    \State Add $f$ to $\mathcal{F}$.
\EndFor

\State \Return $\mathcal{F}, \mathcal{E}$
\end{algorithmic}
\end{algorithm}

\begin{algorithm}[htbp]
\caption{Parallel Symbolic Enumeration (PSE)}
\label{alg:pse}
\begin{algorithmic}[1]
\Require Dataset $\mathcal{D} = (\mathbf{X}, \mathbf{y})$, Operator set $\mathcal{O}$, Variable set $\mathcal{V}=\{v_1, \ldots, v_m\}$, Max iterations $T_{\max}$, PSRN number of input slots $N_{\text{in}}$, Num. sampled variables $N_{\text{vars}}$, Choice of token generator type.
\Ensure Pareto front $\mathcal{P}$ of (expression, error, complexity) tuples.

\State Initialize Token Generator $\pi$ (e.g., $\pi_{\text{GP}}$ (Alg. \ref{alg:gp_generator}), $\pi_{\text{MCTS}}$ (Alg. \ref{alg:mcts_generator_compact_final_fix}), or $\pi_{\text{Random}}$ (Alg. \ref{alg:random_generator})) with $\mathcal{O}, \mathcal{V}$.
\State Initialize PSRN $\psi_{\text{psrn}}$ with $N_{\text{in}}$, operators $\mathcal{O}$, depth $L_{\text{depth}}$.
\State Initialize Pareto front $\mathcal{P} \leftarrow \emptyset$.
\State Initialize base symbols for $\pi$: $s_{\text{base}} \leftarrow \mathcal{V}$.

\For{$t = 1$ to $T_{\max}$ (or while time limit not exceeded)}
    \State \Comment{1. Token Generation by selected $\pi$}
    \State $s_{\text{tokens}} \leftarrow \pi.\Call{Step}{N_{\text{in}} - N_{\text{vars}}, \mathbf{X}, \mathbf{y}, s_{\text{base}}}$.

    \State \Comment{2. Prepare PSRN Inputs}
    \State $s_{\text{vars}} \leftarrow \text{RandomSelect}(N_{\text{vars}} \text{ from } \mathcal{V})$.
    \State $s_0 \leftarrow s_{\text{vars}} \cup s_{\text{tokens}}$. \Comment{Leaf symbols for PSRN, size $N_{\text{in}}$}
    \State $\mathbf{X}_0 \leftarrow \text{EvaluateLeaves}(s_0, \mathbf{X})$. \Comment{Numerical values of $s_0$}

    \State \Comment{3. PSRN Evaluation and Derivation (using Alg. \ref{alg:psrn_eval_derive} logic)}
    \State $\mathcal{F}_{\text{raw}}, \mathcal{E}_{\text{raw}} \leftarrow \Call{PSRN\_Evaluate\_Derive}{\psi_{\text{psrn}}, s_0, \mathbf{X}_0, \mathbf{y}, k_{\text{top}}}$. 
    
    \State \Comment{4. Refine Expressions and Evaluate}
    \State $\mathcal{C}_{\text{candidates}} \leftarrow \emptyset$. \Comment{Collection of (expr, reward, error, complexity)}
    \State $R_{\text{current}} \leftarrow []$. \Comment{Rewards for this iteration}
    \For{each $f_{\text{raw}} \in \mathcal{F}_{\text{raw}}$}
        \State $f_{\text{simp}} \leftarrow \text{SymbolicSimplify}(f_{\text{raw}})$.
        \If{using constants}
            \State $f_{\text{final}}, \epsilon_{\text{final}} \leftarrow \text{TuneCoefficients}(f_{\text{simp}}, \mathbf{X}, \mathbf{y})$.
        \Else
            \State $f_{\text{final}} \leftarrow f_{\text{simp}}$.
            \State $\epsilon_{\text{final}} \leftarrow \text{CalculateError}(f_{\text{final}}, \mathbf{X}, \mathbf{y})$.
        \EndIf
        \State $c_{\text{complexity}} \leftarrow \text{CalculateComplexity}(f_{\text{final}})$.
        \State $r_{\text{reward}} \leftarrow \text{CalculateReward}(\epsilon_{\text{final}}, c_{\text{complexity}})$.
        \State Add $(f_{\text{final}}, r_{\text{reward}}, \epsilon_{\text{final}}, c_{\text{complexity}})$ to $\mathcal{C}_{\text{candidates}}$.
        \State Append $r_{\text{reward}}$ to $R_{\text{current}}$.
    \EndFor
    
    \State \Comment{5. Update Pareto Front and Feedback to Token Generator}
    \State $\text{terminated} \leftarrow \text{UpdateParetoFront}(\mathcal{P}, \mathcal{C}_{\text{candidates}})$.
    \State $s_{\text{base}} \leftarrow \text{ExtractPromisingSymbols}(\mathcal{P})$. \Comment{Update base symbols for $\pi$}
    \If{$R_{\text{current}}$ is not empty}
        \State $\pi.\Call{Reward}{\max(R_{\text{current}}), s_{\text{base}}}$.
    \EndIf
    
    \If{$\text{terminated}$ OR other stopping criteria met}
        \State \textbf{break}
    \EndIf
\EndFor
\State \Return $\mathcal{P}$
\end{algorithmic}
\end{algorithm}

\begin{algorithm}[htbp]
\caption{GP Token Generator ($\pi_{\text{GP}}$)}
\label{alg:gp_generator}
\begin{algorithmic}[1]
\Require Operator set $\mathcal{O}$, Variable set $\mathcal{V}$. 
\Statex \textit{Internal state: Population $P$, GP parameters ($N_{\text{pop}}$, $G_{\text{gen}}$, $p_m$, $p_c$)}.
\Ensure New set of $N_{\text{target\_tokens}}$ token strings $s_{\text{new\_tokens}}$.

\Function{InitializePopulation}{$N_{\text{pop}}, \mathcal{O}, \mathcal{V}, s_{\text{initial\_pool}}$}
    \State $P \leftarrow \emptyset$.
    \State Define Primitive Set $\text{pset}$ using $\mathcal{O}$, $\mathcal{V}$, and elements from $s_{\text{initial\_pool}}$.
    \For{$i = 1$ to $N_{\text{pop}}$}
        \State Create random expression tree $e_i$ using $\text{pset}$.
        \State Add $e_i$ to $P$.
    \EndFor
    \State \Return $P$.
\EndFunction

\Function{Step}{$N_{\text{target\_tokens}}, \mathbf{X}, \mathbf{y}, s_{\text{base}}$}
    \If{population $P$ is not initialized or reset is flagged}
        \State $P \leftarrow \Call{InitializePopulation}{N_{\text{pop}}, \mathcal{O}, \mathcal{V}, s_{\text{base}}}$.
    \EndIf

    \State \Comment{Run GP evolution}
    \State $P_{\text{evolved}} \leftarrow \Call{RunGA}{P, G_{\text{gen}}}$.
    \State $P \leftarrow P_{\text{evolved}}$.

    \State \Comment{Process evolved expressions to generate tokens}
    \State $s_{\text{candidate\_exprs}} \leftarrow \Call{ExtractExpressionsFromPopulationAndHOF}{P}$.
    \State $s_{\text{processed\_tokens}} \leftarrow \Call{ProcessExpressionsToTokens}{s_{\text{candidate\_exprs}}, \mathcal{V}}$.
    \State $s_{\text{new\_tokens}} \leftarrow \Call{SampleTokens}{s_{\text{processed\_tokens}}, N_{\text{target\_tokens}}, \mathcal{V}}$.
    \State \Return $s_{\text{new\_tokens}}$.
\EndFunction

\Function{Reward}{$\text{max\_reward\_signal}, s_{\text{base\_updated}}$}
    \State Add expressions from $s_{\text{base\_updated}}$ to internal buffer to seed future GP runs.
    \State Adapt GP evolution based on $\text{max\_reward\_signal}$.
\EndFunction
\end{algorithmic}
\end{algorithm}

\begin{algorithm}[htbp]
\caption{MCTS Token Generator ($\pi_{\text{MCTS}}$)}
\label{alg:mcts_generator_compact_final_fix}
\begin{algorithmic}[1]
\Require Operator set $\mathcal{O}$, Initial symbols $\mathcal{V}_{\text{init}}$.
\Ensure New set of $N_{\text{target\_tokens}}$ token strings $s_{\text{new\_tokens}}$.
\Statex \textit{Internal state: MCTS tree root $\text{rootNode}$, MCTS parameters ($N_{\text{sim}}$, $c_{\text{UCB}}$), Path for last chosen tokens $\text{lastPath}$}.

\Function{Step}{$N_{\text{target\_tokens}}, s_{\text{base}}$}
    \If{$\text{rootNode}$ is not initialized OR $s_{\text{base}}$ changed}
        \State $\text{rootNode} \leftarrow \Call{NewNode}{\text{state} \leftarrow s_{\text{base}}}$
    \EndIf

    \For{$k = 1$ to $N_{\text{sim}}$} \Comment{MCTS iterations}
        \State $\text{node} \leftarrow \text{rootNode}$
        \State $\text{path} \leftarrow [\text{node}]$
        \While{$\text{node}$ is not leaf AND $\Call{IsFullyExpanded}{\text{node}}$}
            \State $\text{node} \leftarrow \Call{SelectChildUCB}{\text{node}, c_{\text{UCB}}}$
            \State Append $\text{node}$ to $\text{path}$.
        \EndWhile
        \If{NOT $\Call{IsTerminal}{\text{node}}$ AND NOT $\Call{IsFullyExpanded}{\text{node}}$}
            \State $\text{node} \leftarrow \Call{Expand}{\text{node}, \mathcal{O}, \mathcal{V}_{\text{init}}}$
            \State Append $\text{node}$ to $\text{path}$.
        \EndIf
        \State $r_{\text{sim}} \leftarrow \Call{SimulateAndEvaluate}{\text{node}, \mathcal{O}, \mathcal{V}_{\text{init}}}$
        \State \Call{BackpropagateReward}{\text{path}, $r_{\text{sim}}$}
    \EndFor

    \State $s_{\text{new\_tokens}}, \text{chosenPath} \leftarrow \Call{ExtractPromisingTokens}{\text{rootNode}, N_{\text{target\_tokens}}}$
    \State $\text{lastPath} \leftarrow \text{chosenPath}$
    \State \Return $s_{\text{new\_tokens}}$
\EndFunction

\Function{Reward}{$\text{actualReward}, s_{\text{base\_updated}}$}
    \If{$\text{lastPath}$ is set}
      \State \Call{BackpropagateReward}{\text{lastPath}, \text{actualReward}}
    \EndIf
    \State \Comment{$s_{\text{base\_updated}}$ informs root for next step.}
\EndFunction
\end{algorithmic}
\end{algorithm}

\begin{algorithm}[htbp]
\caption{Random Token Generator ($\pi_{\text{Random}}$)}
\label{alg:random_generator}
\begin{algorithmic}[1]
\Require Operator set $\mathcal{O}$, Variable set $\mathcal{V}$. 
\Statex \textit{Internal state: Max depth for random trees $D_{\text{rand\_max}}$}.
\Ensure New set of $N_{\text{target\_tokens}}$ token strings $s_{\text{new\_tokens}}$.

\Function{GenerateRandomTree}{$\text{current\_depth}, D_{\text{rand\_max}}, \mathcal{O}, s_{\text{pool}}$}
    \If{$\text{current\_depth} == D_{\text{rand\_max}}$ OR ($\text{current\_depth} > 0$ AND random stop condition)}
        \State \Return $\Call{RandomlySelect}{s_{\text{pool}}}$.
    \EndIf
    
    \State $\text{op} \leftarrow \Call{RandomlySelect}{\mathcal{O}}$.
    \If{$\text{op}$ is unary}
        \State $\text{operand} \leftarrow \Call{GenerateRandomTree}{\text{current\_depth}+1, D_{\text{rand\_max}}, \mathcal{O}, s_{\text{pool}}}$.
        \State \Return $\text{op}(\text{operand})$.
    \Else \Comment{$\text{op}$ is binary}
        \State $\text{operand1} \leftarrow \Call{GenerateRandomTree}{\text{current\_depth}+1, D_{\text{rand\_max}}, \mathcal{O}, s_{\text{pool}}}$.
        \State $\text{operand2} \leftarrow \Call{GenerateRandomTree}{\text{current\_depth}+1, D_{\text{rand\_max}}, \mathcal{O}, s_{\text{pool}}}$.
        \State \Return $\text{op}(\text{operand1}, \text{operand2})$.
    \EndIf
\EndFunction

\Function{Step}{$N_{\text{target\_tokens}}, \mathbf{X}, \mathbf{y}, s_{\text{base}}$}
    \State $s_{\text{new\_tokens}} \leftarrow \emptyset$.
    \State $\text{TokenPool} \leftarrow s_{\text{base}} \cup \mathcal{V}$.
    \If{using constants}
        \For{$j=1$ to $N_{\text{const\_to\_generate}}$}
            \State Add $\Call{SampleRandomConstant}{}$ to $\text{TokenPool}$.
        \EndFor
    \EndIf

    \For{$i = 1$ to $N_{\text{target\_tokens}}$}
        \State $\text{newexpr} \leftarrow \Call{GenerateRandomTree}{0, D_{\text{rand\_max}}, \mathcal{O}, \text{TokenPool}}$.
        \State Add $\text{newexpr}$ to $s_{\text{new\_tokens}}$.
    \EndFor
    \State \Return $s_{\text{new\_tokens}}$.
\EndFunction

\Function{Reward}{$\text{max\_reward\_signal}, s_{\text{base\_updated}}$}
    \State \Comment{No specific update needed for purely random generator.}
\EndFunction
\end{algorithmic}
\end{algorithm}


\newpage
\section{Summary of Problem Sets, Noise, Constant Settings}
\label{sec:exp_summary}

The configurations summarized in Table \ref{tab:exp_summary} are based on the following standard procedures. For experiments involving synthetic noise, we use an additive Gaussian model where the noisy target data, $y_{\text{noisy}}$, is generated from the clean data, $y_{\text{clean}}$, according to the formula: $y_{\text{noisy}} = y_{\text{clean}} + \text{noise\_level} \times \text{std}(y_{\text{clean}}) \times \mathcal{N}(0,1)$. Here, $\mathcal{N}(0,1)$ denotes a standard normal distribution.

The ``Token Constants + Coefficients Refinement'' column in the table indicates whether the constant handling mechanism for each algorithm was enabled or disabled for a given problem set.

\begin{table}[h!]
\centering

\caption{Summary of Experimental Configurations for All Datasets. This table provides an overview of the key settings for each experiment, including the number of variables, data points, noise characteristics, and the method used for handling numerical constants in the discovered expressions.}
\label{tab:exp_summary}
\resizebox{\textwidth}{!}{%
\begin{threeparttable}
\begin{tabular}{@{}lcccc@{}}
\toprule
\textbf{Problem Set} & \textbf{\# Variables} & \textbf{\# Data Points} & \textbf{Noise Level \& Type} & \textbf{Token Constants + Coefficients Refinement} \\
\midrule
\multicolumn{5}{l}{\textit{--- Problem Sets without Coefficients ---}} \\
Nguyen & 1--2 & 20 & Noiseless  & Disabled \\
R & 1 & 20 & Noiseless  & Disabled \\
R$^*$ & 1 & 20 & Noiseless  & Disabled \\
Livermore & 1--2 & 20--1000 & Noiseless  & Disabled \\
Feynman & 2--5 & 20--50 & Noiseless  & Disabled \\
SRBench & 1--10 & 50000 & \makecell[c]{0\%, 0.1\%, 1\%, and 10\% \\ Additive Gaussian} & Disabled \\
\midrule
\multicolumn{5}{l}{\textit{--- Problem Sets with Coefficients ---}} \\
Nguyen-c & 1--2 & 20 & Noiseless  & Enabled \\

Chaotic Dynamics & 3--4 & 1000 & \makecell[c]{1\%, 2\%, 5\%, and 10\%\\ Additive Gaussian} & Enabled \\
EMPS & 3 & 5000 & Real-world experimental noise & Enabled \\
Turbulent Friction & 1 (transformed) & 362 & Real-world experimental noise & Enabled \\
\bottomrule
\end{tabular}
\end{threeparttable}
}
\end{table}


\newpage

\section{High-dimensional Synthetic Problem Sets}

To evaluate the model's performance on high-dimensional symbolic regression and to showcase its ability to handle a large number of variables, we generated a benchmark suite of 20 synthetic problems. Each problem is defined by a ground-truth expression constructed from 12 active variables, which were randomly sampled without replacement from a pool of 50 variables ($x_1, \dots, x_{50}$). The remaining 38 variables act as distractors. For each problem, we generated a dataset of 500 samples where all input variables were drawn from a uniform distribution, $U(-1, 1)$. The entire generation process, including variable selection and data sampling, was governed by a fixed random seed to ensure full reproducibility.

The symbolic expressions were generated under several constraints to ensure complexity and consistency. Each expression was constructed to include each of the four basic arithmetic operators $\{+, -, \times, \div\}$) at least once and to use each of the 12 active variables exactly once. The operators were chosen from a categorical distribution with weights of 4, 3, 2, and 1 for addition, multiplication, subtraction, and division, respectively, corresponding to sampling probabilities of 0.4, 0.3, 0.2, and 0.1. Furthermore, to introduce structural diversity, parentheses were inserted at each potential branching point in the expression tree with a probability of 0.15. Finally, all expressions were validated to ensure they did not produce undefined (\texttt{NaN}) or infinite (\texttt{Inf}) values on the generated dataset.

\begin{table}[ht]
\centering
\caption{\textbf{High-dimensional synthetic regression problems} ($d=$50). The ground truth equations are embedded in a 50-dimensional space, where unused variables act as distractors. The variable indices for each equation are randomly sampled from $\{1, ..., 50\}$ to create a more challenging feature selection task. Using operator set $\mathcal{O} = \{+, -, \times, \div\}$.}
\label{tab:synthetic_50d}

\begin{tabular}{lll}
\toprule
\textbf{Name} & \textbf{Equation} & \textbf{Dataset} \\
\midrule
Problem-1 & $x_{37} + x_{44} / x_{30} + x_{27} / x_{6} + x_{20} + x_{26} - x_{13} + x_{50} \times x_{18} \times x_{47} + x_{34}$ & $U(-1, 1, 500)$ \\
Problem-2 & $x_{9} - x_{19} \times x_{49} \times x_{10} / x_{43} + x_{34} + x_{20} - x_{23} / x_{30} \times x_{13} + x_{44} + x_{42}$ & $U(-1, 1, 500)$ \\
Problem-3 & $x_{29} \times x_{21} / x_{34} + x_{43} \times x_{28} + x_{17} - x_{13} \times x_{49} \times x_{37} + x_{32} - x_{50} / x_{44}$ & $U(-1, 1, 500)$ \\
Problem-4 & $x_{13} \times x_{37} + x_{20} \times x_{10} \times x_{15} - x_{17} \times x_{42} / x_{22} + x_{29} \times x_{12} / x_{40} - x_{41}$ & $U(-1, 1, 500)$ \\
Problem-5 & $x_{37} / x_{43} \times x_{20} + x_{18} + x_{34} \times x_{7} - x_{30} \times x_{5} + x_{41} + x_{10} + x_{40} \times x_{42}$ & $U(-1, 1, 500)$ \\
Problem-6 & $x_{9} \times x_{36} + x_{42} / x_{30} \times x_{33} + x_{23} - x_{17} + x_{15} + x_{43} \times x_{13} / x_{11} \times x_{16}$ & $U(-1, 1, 500)$ \\
Problem-7 & $x_{45} \times x_{50} - x_{20} - x_{28} / x_{33} \times x_{36} - x_{22} + x_{18} + x_{35} \times x_{16} + x_{12} + x_{46}$ & $U(-1, 1, 500)$ \\
Problem-8 & $x_{39} \times x_{20} + x_{31} \times x_{23} + x_{38} + x_{32} + x_{5} - x_{7} / x_{27} + x_{18} + x_{35} - x_{36}$ & $U(-1, 1, 500)$ \\
Problem-9 & $x_{41} - x_{37} - x_{22} \times x_{29} + x_{26} / x_{49} + x_{32} + x_{12} \times x_{50} / x_{28} + x_{38} + x_{21}$ & $U(-1, 1, 500)$ \\
Problem-10 & $x_{38} \times x_{44} - x_{48} - x_{20} + x_{19} \times x_{37} / x_{9} - x_{45} + x_{6} + x_{33} + x_{23} + x_{18}$ & $U(-1, 1, 500)$ \\
Problem-11 & $x_{34} / x_{40} - x_{41} + x_{15} - x_{9} + x_{28} \times x_{46} / x_{20} + x_{24} + x_{14} / x_{47} \times x_{36}$ & $U(-1, 1, 500)$ \\
Problem-12 & $x_{36} / x_{28} + x_{13} - x_{9} + x_{49} + x_{12} + x_{34} \times x_{46} + x_{42} + x_{45} \times x_{22} + x_{33}$ & $U(-1, 1, 500)$ \\
Problem-13 & $x_{15} \times x_{45} + x_{27} + x_{22} + x_{40} \times x_{35} / x_{26} + x_{42} - x_{6} + x_{5} + x_{25} \times x_{13}$ & $U(-1, 1, 500)$ \\
Problem-14 & $x_{17} - x_{50} / x_{34} - x_{24} \times x_{36} - x_{49} - x_{44} + x_{16} + x_{31} - x_{7} / x_{43} \times x_{19}$ & $U(-1, 1, 500)$ \\
Problem-15 & $x_{19} + x_{37} - x_{28} + x_{16} \times x_{43} - x_{22} \times x_{44} - x_{39} + x_{21} \times x_{35} / x_{25} \times x_{15}$ & $U(-1, 1, 500)$ \\
Problem-16 & $x_{30} + x_{38} \times x_{31} \times x_{27} / x_{34} + x_{50} \times x_{15} \times x_{7} - x_{26} / x_{39} / x_{48} + x_{24}$ & $U(-1, 1, 500)$ \\
Problem-17 & $x_{25} + x_{34} - x_{30} \times x_{36} / x_{38} - x_{47} + x_{19} / x_{20} \times x_{50} + x_{28} / x_{14} - x_{26}$ & $U(-1, 1, 500)$ \\
Problem-18 & $x_{45} + x_{15} \times x_{47} + x_{25} + x_{28} - x_{16} - x_{12} + x_{18} + x_{37} - x_{7} \times x_{20} / x_{11}$ & $U(-1, 1, 500)$ \\
Problem-19 & $x_{45} / x_{10} + x_{25} / x_{26} + x_{39} - x_{7} - x_{49} + x_{13} + x_{47} \times x_{12} - x_{29} + x_{36}$ & $U(-1, 1, 500)$ \\
Problem-20 & $x_{12} - x_{39} \times x_{49} - x_{34} - x_{45} / x_{24} + x_{26} \times x_{5} \times x_{47} + x_{18} + x_{8} \times x_{32}$ & $U(-1, 1, 500)$ \\
\bottomrule
\end{tabular}
\end{table}

\newpage
We conducted a comprehensive benchmark to compare our method, PSE, against two state-of-the-art baselines: PySR~\cite{cranmer2023interpretable_PySR} and Operon~\cite{Operon}. All algorithms were tested on the 20 high-dimensional problems detailed in Table~\ref{tab:synthetic_50d}, with a time limit of 10 minutes per problem. Notably, our implementation of PSE, which utilizes a 2-layer PSRN with 70 inputs, required only 8.5 GB of GPU memory, making it accessible on most modern consumer-grade GPUs.

The comparative results are detailed in Table~\ref{tab:full_comparison}. A stark performance gap is evident between PSE and the baseline methods. As the table demonstrates, both PySR and Operon consistently achieved a 0\% recovery rate, failing to find any correct solutions within the allocated time. In contrast, PSE successfully recovered a substantial fraction of the ground-truth expressions, underscoring its robustness in navigating high-dimensional search spaces with numerous distractor variables.

\begin{table}[htbp]
\centering
\caption{\textbf{Recovery rate comparison on 20 high-dimensional synthetic problems.} The table shows the symbolic recovery rate for each algorithm. Recovery rates are aggregated from 20 independent runs for each problem.}

\label{tab:full_comparison}
\begin{tabular}{l c c c}
\toprule
\textbf{Problem} & \textbf{PSE (Ours)} & \textbf{PySR} & \textbf{Operon} \\
\midrule
Problem-1  & \textbf{10\%} & 0\% & 0\% \\
Problem-2  & \textbf{0\%} & 0\% & 0\% \\
Problem-3  & \textbf{10\%} & 0\% & 0\% \\
Problem-4  & \textbf{10\%} & 0\% & 0\% \\
Problem-5  & \textbf{90\%} & 0\% & 0\% \\
Problem-6  & \textbf{5\%} & 0\% & 0\% \\
Problem-7  & \textbf{100\%} & 0\% & 0\% \\
Problem-8  & \textbf{90\%} & 0\% & 0\% \\
Problem-9  & \textbf{60\%} & 0\% & 0\% \\
Problem-10 & \textbf{95\%} & 0\% & 0\% \\
Problem-11 & \textbf{10\%} & 0\% & 0\% \\
Problem-12 & \textbf{100\%} & 0\% & 0\% \\
Problem-13 & \textbf{95\%} & 0\% & 0\% \\
Problem-14 & \textbf{80\%} & 0\% & 0\% \\
Problem-15 & \textbf{100\%} & 0\% & 0\% \\
Problem-16 & \textbf{0\%} & 0\% & 0\% \\
Problem-17 & \textbf{45\%} & 0\% & 0\% \\
Problem-18 & \textbf{100\%} & 0\% & 0\% \\
Problem-19 & \textbf{95\%} & 0\% & 0\% \\
Problem-20 & \textbf{10\%} & 0\% & 0\% \\
\midrule
\multicolumn{4}{c}{\textbf{Summary}} \\
\midrule
\textbf{Average} & \textbf{40\%} (160/400) & \textbf{0\%} & \textbf{0\%} \\
\bottomrule
\end{tabular}

\end{table}

\newpage
\section{Analysis of Performance Limitations and Failure Modes}

In the main text, we discussed the conceptual limitations of PSE. Here, we provide a concrete analysis of these boundaries by examining specific cases where the algorithm faces significant challenges. These examples serve to illustrate the conditions under which PSE, like other data-driven discovery methods, may struggle to identify the ground-truth expression.

\subsection{Multi-level Nested Complex Expressions}

\begin{figure}[b!]
  \hspace*{-0.0in}
  \centering
  \includegraphics[width=6.5in]{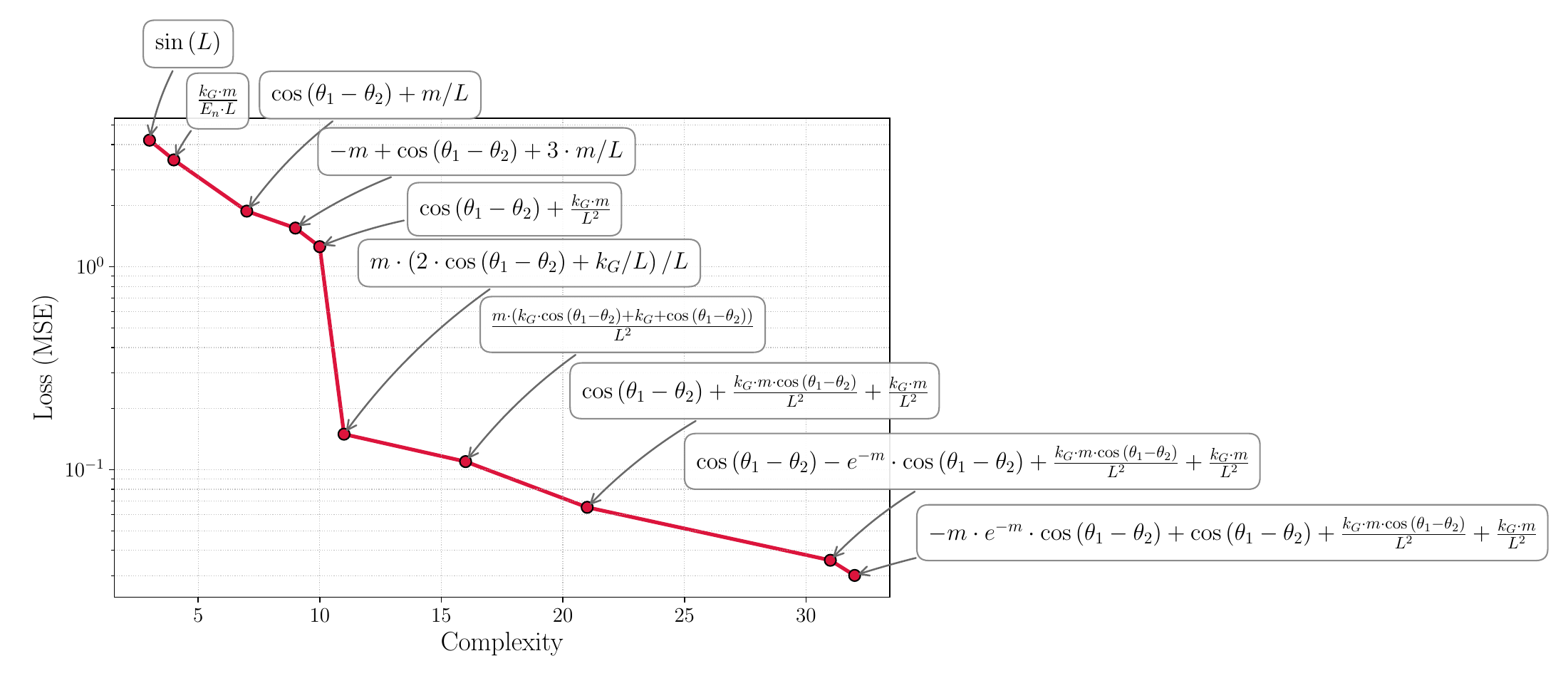}
  \caption{PSE's Pareto Front on Feynman test 2 Problem when the running time is 1 hour.}
  \label{fig_feynman_test_2}
\end{figure}

\begin{equation}
\label{feynman-test-2}
    \frac{m \cdot k_G}{L^2} \left(1 + \sqrt{1 + \frac{2 E_n L^2}{m k_G^2}} \cdot \cos(\theta_1 - \theta_2)\right)
\end{equation}

Although our PSE achieves a high average recovery rate in SRBench \cite{SRBench_la2021contemporary} (see Fig. \textcolor{blue}{6\textbf{d}--\textbf{f}} in the \textcolor{blue}{Main text}), it struggles with more challenging cases, such as the Feynman-test-2 example shown here in Eq. \ref{feynman-test-2}. Within a one-hour time limit, PSE fails to correctly recover this expression. Here, we display the Pareto front obtained after running PSE for one hour—while it successfully identifies the terms \(\frac{m \cdot k_G}{L^2}\) and \(\cos(\theta_1 - \theta_2)\), it fails to recover the highly nonlinear and complex term \(\sqrt{1 + \frac{2 E_n L^2}{m k_G^2}}\). (See Fig. \ref{fig_feynman_test_2})

This specific failure mode is instructive, as it highlights the current frontier of the PSE framework and the intricate interplay between its components. The challenge lies not in the enumeration capacity of the PSRN module itself, but in the stochastic discovery process of the token generator (e.g., GP). For PSE to succeed on such a problem, the generator is required to construct a highly nested, multi-variable sub-expression like $\sqrt{1 + \frac{2 E_n L^2}{m k_G^2}}$ as a single, coherent token. The search landscape for such problems can be deceptive; simpler expressions found on the Pareto front, such as those involving only linear combinations of $\cos(\theta_1 - \theta_2)$ and other terms, offer a compellingly low error, creating strong local optima that are difficult for a heuristic search to escape.

This represents a classic challenge in symbolic regression: balancing exploration of structurally complex regions of the search space against exploitation of ``good-enough'' simpler solutions. The marginal improvement in accuracy offered by the true, complex term may not provide a sufficient reward signal to guide the evolutionary process toward its discovery within a practical timeframe. Future work could address this by developing more sophisticated token generation strategies, such as incorporating domain-specific structural priors, integrating multiple powerful token generator, or employing multi-objective rewards that explicitly value structural novelty. This would enhance the synergy between the heuristic search and parallel enumeration, thereby empower PSE to more systematically conquer these problems, where solutions are characterized by deep compositional structure and deceptive fitness landscapes.

\subsection{Token Constants Sensitivity}

\begin{figure}[t!]
    \centering
    \begin{minipage}[b]{0.28\textwidth}
        \includegraphics[width=\textwidth]{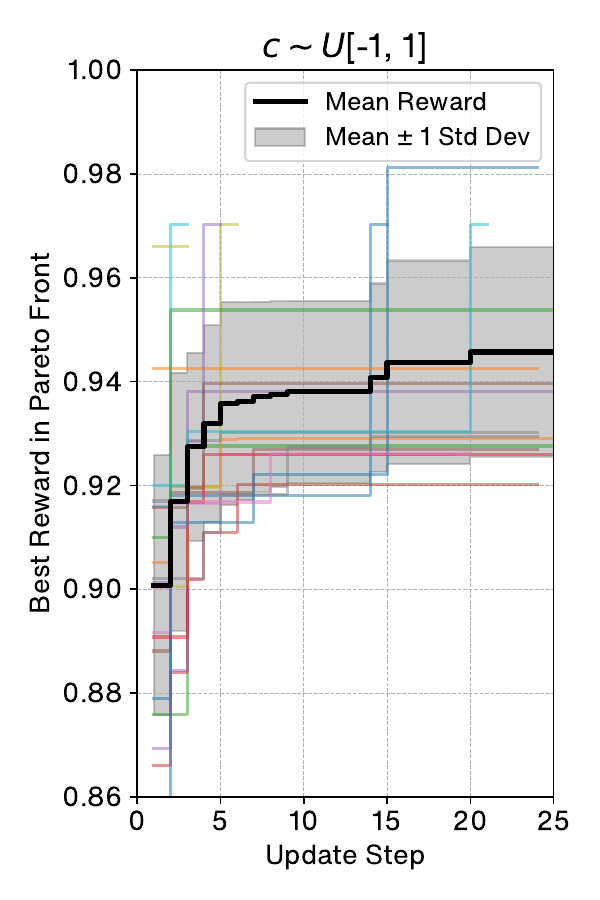}
    \end{minipage}
    \begin{minipage}[b]{0.28\textwidth}
        \includegraphics[width=\textwidth]{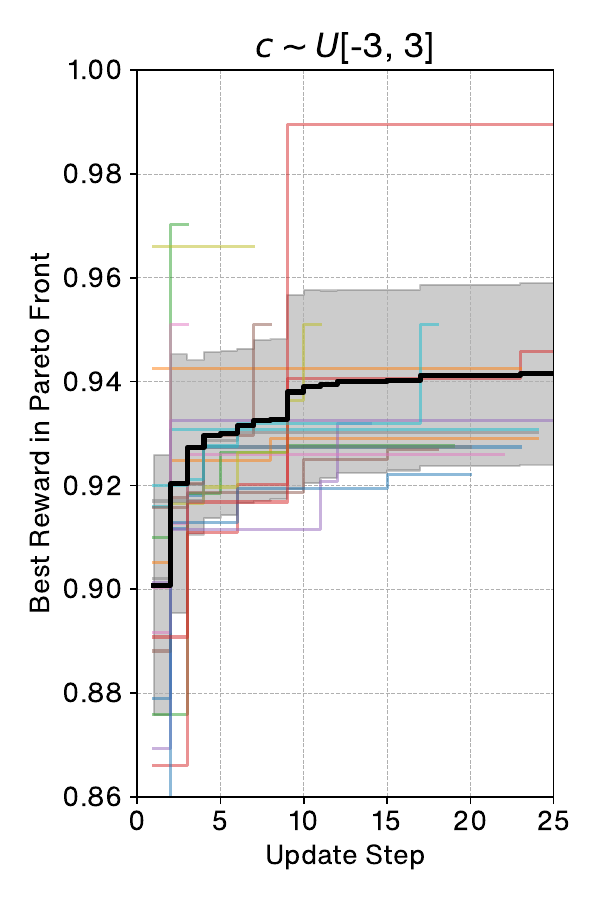}
    \end{minipage}
    \begin{minipage}[b]{0.28\textwidth}
        \includegraphics[width=\textwidth]{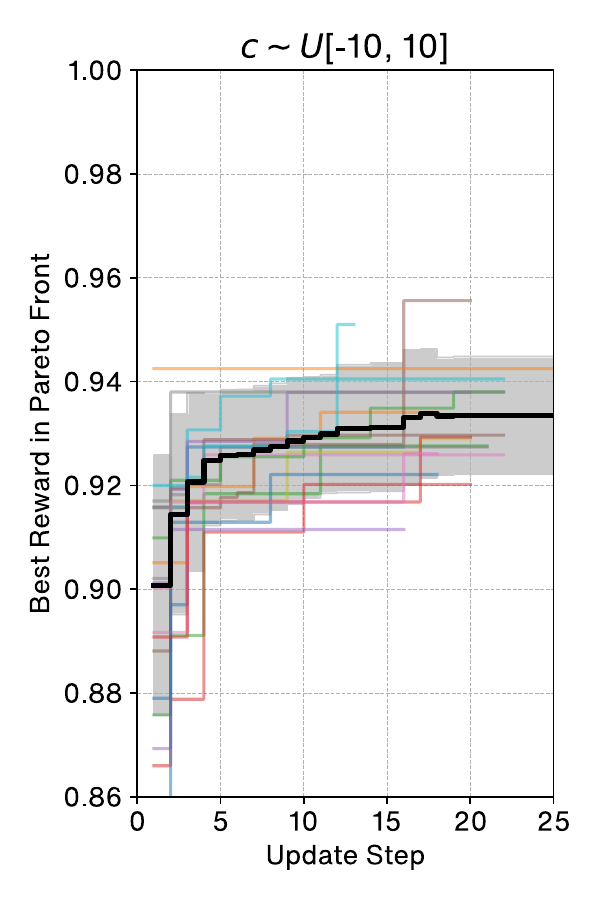}
    \end{minipage}
    \caption{Reward values of optimal expressions across PSE iterations for Nguyen-8c under three token constant range ($U[-1, 1]$, $U[-3, 3]$, and $U[-10, 10]$). Each colored thin line represents an independent run (20 runs total), with the mean and one standard deviation interval indicated by the black line and shaded area respectively.}
    \label{fig_SI_nguyen8c}
\end{figure}

Here we provide a more in-depth analysis of how the sampling range of token constants affects recovery speed in our ablation experiments. Taking Nguyen-8c as an example, Fig. \ref{fig_SI_nguyen8c} demonstrates that varying the sampling range of token constants (specifically $U[-1, 1]$, $U[-3, 3]$, and $U[-10, 10]$ in this case) influences the efficiency of PSE in discovering the ground truth. 

While the discovery efficiency shows minimal difference between the first two ranges, we observe a slight decline in recovery speed when the constant sampling range expands further from the true coefficients to $U[-10, 10]$. This is evidenced by the mean curve (black line) across multiple runs, where the reward of optimal expressions in the Pareto frontier becomes marginally lower at equivalent iteration steps compared to $U[-1, 1]$ and $U[-3, 3]$.

\subsection{PSRN's Affinity for Fractional Structures}

\begin{figure}[t!]
  \hspace*{-0.0in}
  \centering
  \includegraphics[width=4.5in]{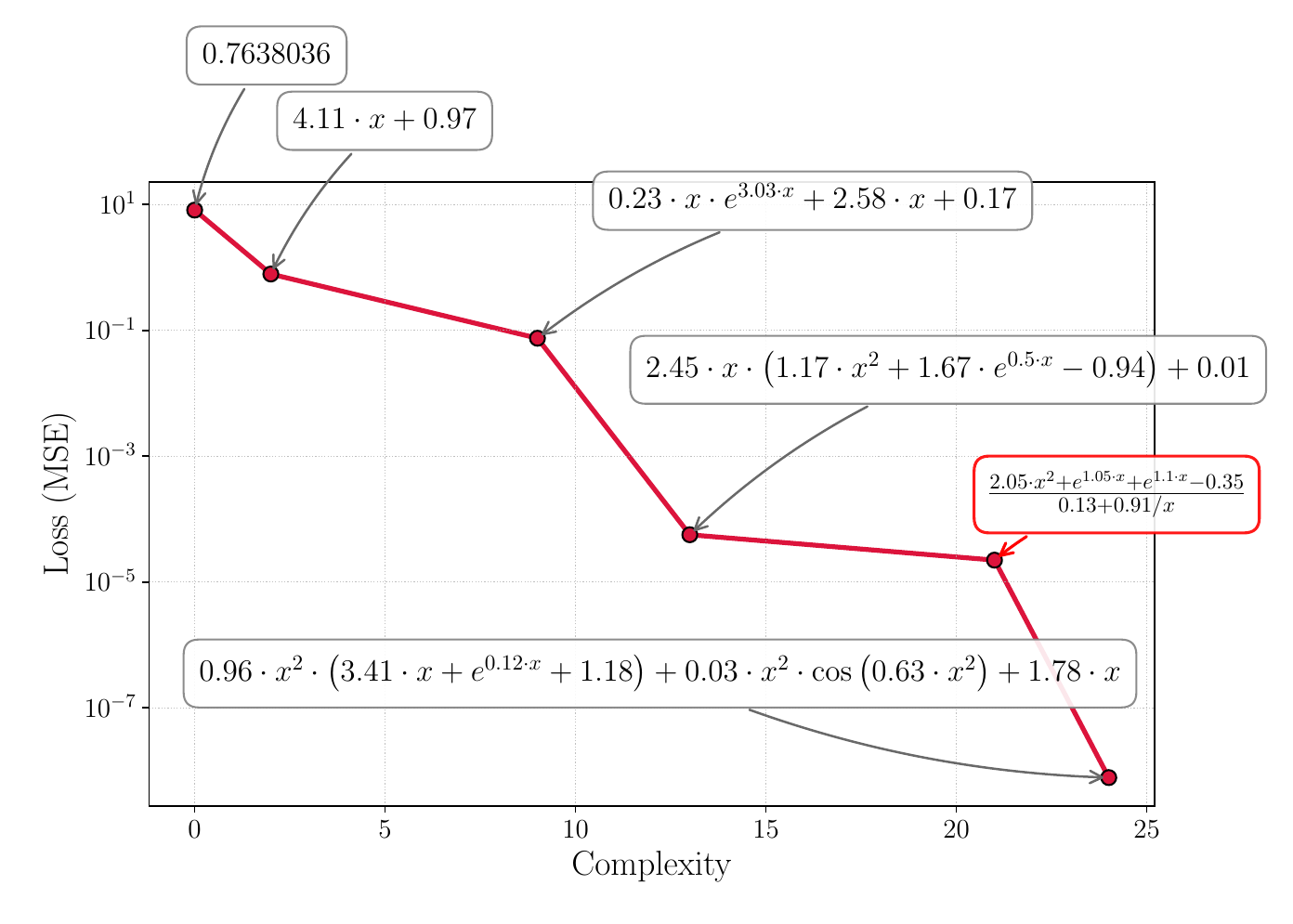}
  \caption{PSE's Pareto Front on Nguyen-1c Problem when the PSE step is at 2 (running time is approximately at 10 seconds). The red box highlights one of the fractional structures found by PSE.}
  \label{fig_Fractional_pareto_Nguyen_1c}
\end{figure}

\begin{figure}[t!]
  \hspace*{-0.0in}
  \centering
  \includegraphics[width=4.5in]{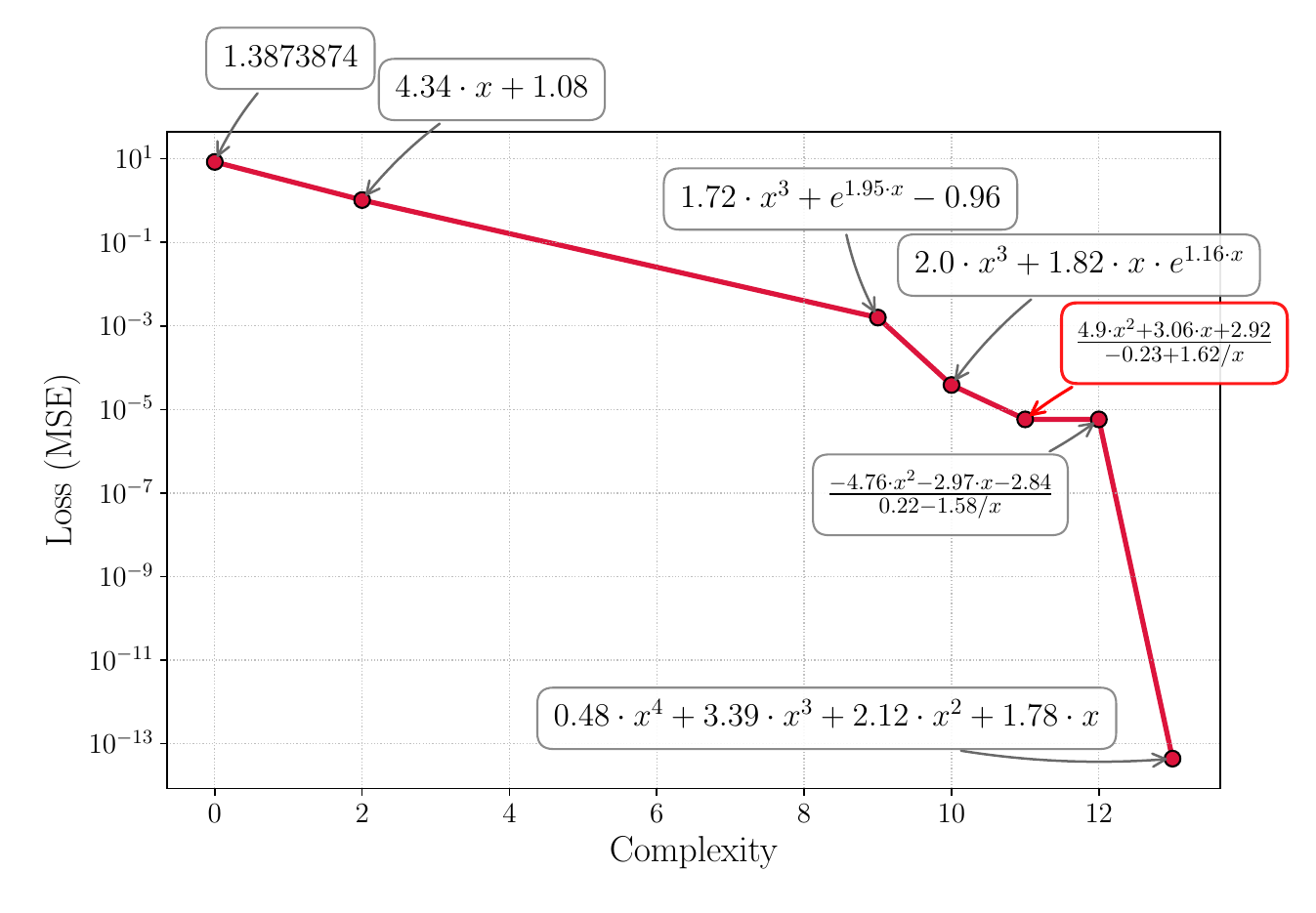}
  \caption{PSE's Pareto Front on Nguyen-2c Problem when the PSE step is at 7 (running time is approximately at 35 seconds). The red box highlights one of the fractional structures found by PSE.}
  \label{fig_Fractional_pareto_Nguyen_2c}
\end{figure}

\begin{figure}[t!]
  \hspace*{-0.0in}
  \centering
  \includegraphics[width=4.5in]{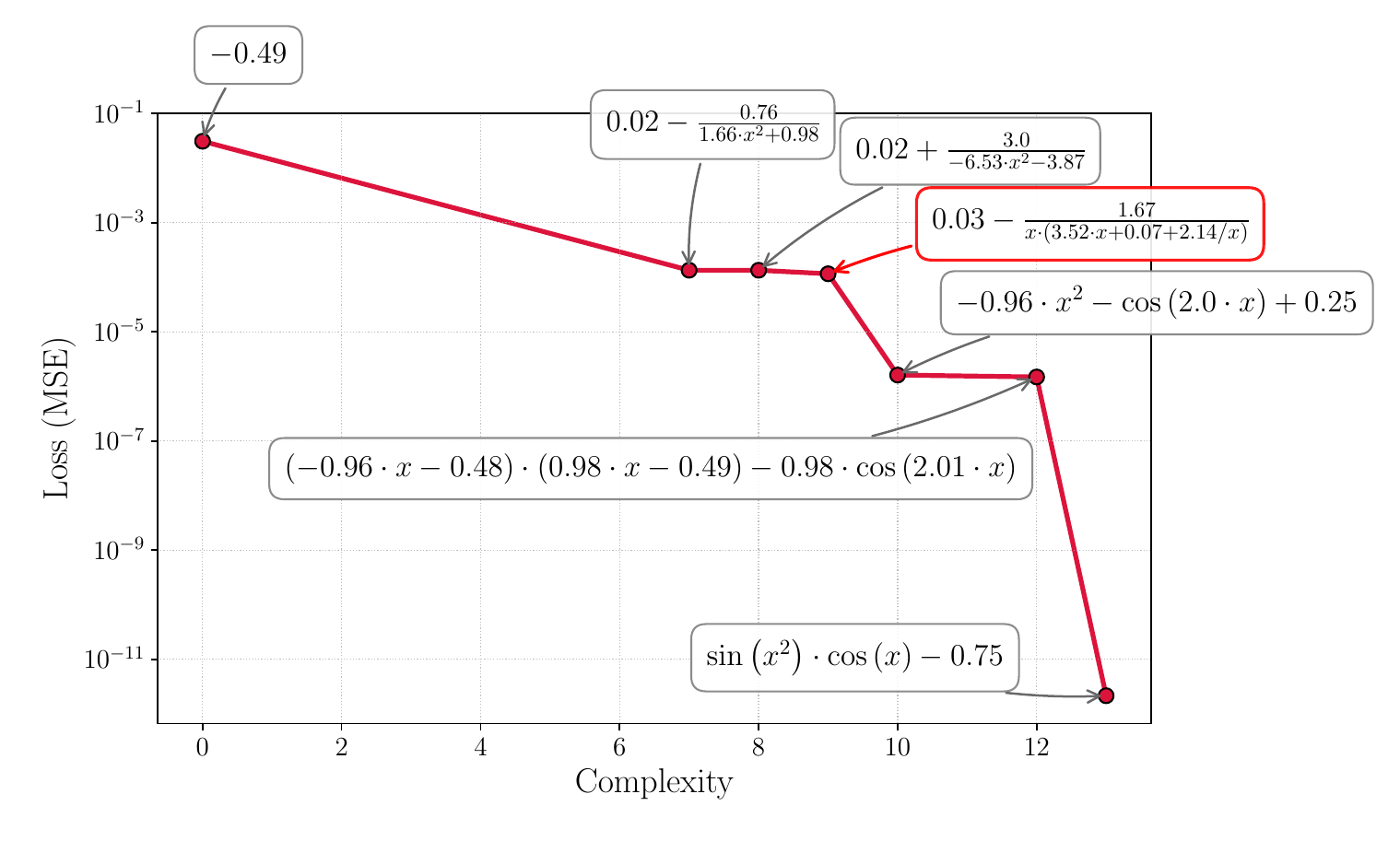}
  \caption{PSE's Pareto Front on Nguyen-5c Problem when the PSE step is at 6 (running time is approximately at 30 seconds). The red box highlights one of the fractional structures found by PSE.}
  \label{fig_Fractional_pareto_Nguyen_5c}
\end{figure}

Another notable characteristic of PSRN is its strong tendency to discover expressions in fractional forms. This behavior is reminiscent of the Padé approximant, a method that uses rational functions (ratios of polynomials) to approximate complex functions, i.e., approximating a function \( f(x) \) as the ratio of two polynomials \( R_{[m/n]}(x) = \frac{P_m(x)}{Q_n(x)} \). Here, we demonstrate the Pareto frontiers of PSE when searching for Nguyen-1c, Nguyen-2c, and Nguyen-5c. Figs. \ref{fig_Fractional_pareto_Nguyen_1c}--\ref{fig_Fractional_pareto_Nguyen_5c} show that patterns resembling a fractional structure frequently appear in PSE's Pareto frontiers. This behavior aligns with PSE's high recovery rate (close to 100\%) on R-1, R-2, R-3, as well as R-1$^{*}$, R-2$^{*}$, and R-3$^{*}$, while the recovery rates of other methods approach 0\%. This is because PSRN's exhaustive nature allows it to directly explore complex fractional structures, which fundamentally differs from mainstream GP-based methods (which typically evolve sub-structures of expressions in a more incremental manner). On the other hand, this could also become a unique limitation of PSRN—for instance, under high-noise conditions, its overly strong fitting capability might mistakenly identify noise patterns as fractional structures.

\newpage
\section{Comparison with ESR}

We conducted a direct comparison with Exhaustive Symbolic Regression (ESR) \cite{ESR_Bartlett_2022}, another enumeration-based method. However, this comparison is constrained by a key architectural limitation of ESR: it is designed exclusively for single-variable problems. Consequently, we could only evaluate its performance on the subset of our benchmarks that fit this criterion. For all single-variable problems from the main text, we ran both algorithms under a one-hour time limit per problem. Here, all conditions, such as the operator used ($\mathcal{O}_{\text{Koza}}$) and the time budget (1 hour), are kept identical to those in the main text to ensure a fair comparison.

The results, summarized in Table \ref{tab:performance_comparison}, reveal a significant performance gap. PSE achieved a 100\% recovery rate across all 37 single-variable problems, whereas ESR successfully recovered only 5 expressions, resulting in an average recovery rate of 13.5\%. Furthermore, PSE was substantially more efficient, finding solutions in an average of 83 seconds, while ESR consistently utilized the full one-hour time limit.

This performance disparity stems from a fundamental architectural difference in their enumeration strategies. ESR performs an \textbf{explicit enumeration}: it exhaustively constructs every possible symbolic expression as a SymPy object and then evaluates each one. This process is inherently time-consuming, especially as expression complexity increases. In contrast, PSE's PSRN module performs an \textbf{implicit enumeration}. By leveraging parallel computation and common subtree reuse, it first numerically evaluates a vast space of expressions to identify the top-performing candidates based on their loss. Only then does it recursively reconstruct the symbolic form of these few optimal expressions. This ``evaluate first, reconstruct second'' approach avoids the computational bottleneck of building millions of symbolic objects.

A crucial step for ESR is the pre-generation of a file containing all unique equations up to a certain complexity for a given operator set. As shown in Table \ref{tab:time_generate}, generating this file for the standard Koza library can take more than two days. In contrast, generating the equivalent DR Mask for PSE, which serves a similar purpose of eliminating redundancy, takes less than a minute. This significant difference in setup time underscores the practical efficiency and scalability advantages of PSE's implicit enumeration approach.

\begin{table}[htbp]
        \centering
        \caption{Performance Comparison between PSE and ESR} 
        \label{tab:performance_comparison}
        \scalebox{0.8}{

        \begin{tabular}{ccccc}
            \toprule
            
            Problem & \multicolumn{2}{c}{Recovery Rate (\%)} & \multicolumn{2}{c}{Time (s)} \\
            
            \cmidrule(lr){2-3} \cmidrule(lr){4-5}

            & PSE & ESR & PSE & ESR \\
            \midrule
            
            Nguyen-1 & 100\% & 0\% & 1.26 & 3601.09 \\
            Nguyen-2 & 100\% & 0\% & 1.32 & 3600.70 \\
            Nguyen-3 & 100\% & 0\% & 2.93 & 3600.73 \\
            Nguyen-4 & 100\% & 0\% & 3.22 & 3601.13 \\
            Nguyen-5 & 100\% & 0\% & 3.25 & 3600.58 \\
            Nguyen-6 & 100\% & 100\% & 2.78 & 3600.88 \\
            Nguyen-7 & 100\% & 0\% & 17.97 & 3601.18 \\
            Nguyen-8 & 100\% & 100\% & 3.26 & 3600.82 \\
            \midrule
            Nguyen-1c & 100\% & 0\% & 25.62 & 3601.06 \\
            Nguyen-2c & 100\% & 0\% & 51.32 & 3601.12 \\
            Nguyen-5c & 100\% & 100\% & 13.89 & 3600.99 \\
            Nguyen-8c & 100\% & 100\% & 704.40 & 3601.61 \\
            \midrule
            R-1 & 100\% & 0\% & 133.00 & 3601.72 \\
            R-2 & 100\% & 0\% & 98.22 & 3601.78 \\
            R-3 & 100\% & 0\% & 767.96 & 3601.28 \\
            \midrule
            R*-1 & 100\% & 0\% & 44.35 & 3600.95 \\
            R*-2 & 100\% & 0\% & 36.42 & 3601.61 \\
            R*-3 & 100\% & 0\% & 24.09 & 3600.90 \\
            \midrule
            Livermore-1 & 100\% & 0\% & 6.19 & 3601.25 \\
            Livermore-2 & 100\% & 0\% & 2.81 & 3601.45 \\
            Livermore-3 & 100\% & 0\% & 65.57 & 3601.21 \\
            Livermore-4 & 100\% & 0\% & 21.71 & 3601.26 \\
            Livermore-5 & 100\% & 0\% & 7.20 & 3600.89 \\
            Livermore-6 & 100\% & 0\% & 15.13 & 3601.53 \\
            Livermore-7 & 100\% & 0\% & 2.29 & 3600.31 \\
            Livermore-8 & 100\% & 0\% & 2.19 & 3601.29 \\
            Livermore-9 & 100\% & 0\% & 51.88 & 3601.18 \\
            Livermore-13 & 100\% & 0\% & 56.52 & 3600.72 \\
            Livermore-14 & 100\% & 0\% & 3.05 & 3601.75 \\
            Livermore-15 & 100\% & 0\% & 225.37 & 3601.77 \\
            Livermore-16 & 100\% & 0\% & 550.75 & 3601.09 \\
            Livermore-18 & 100\% & 0\% & 15.93 & 3600.66 \\
            Livermore-19 & 100\% & 0\% & 4.51 & 3600.88 \\
            Livermore-20 & 100\% & 100\% & 2.09 & 3600.69 \\
            Livermore-21 & 100\% & 0\% & 15.71 & 3600.50 \\
            Livermore-22 & 100\% & 0\% & 4.53 & 3601.06 \\
            \midrule
            \textbf{Average} & \textbf{100}\% & \textbf{13.5}\% & \textbf{82.96} & \textbf{3601.10} \\
            \bottomrule
            
        \end{tabular}
        }
\end{table}

\begin{table}[htbp]
        \centering
        \caption{Time to Generate DR mask or Unique Expressions for $\mathcal{O}_{\text{Koza}} = \{+,-,\times, \div, \sin, \cos, \exp, \log\}$} 
        \label{tab:time_generate}
        \vspace{6pt}
        
        \begin{tabular}{lcc}
          \toprule
          & PSE & ESR \\
          \midrule
        Time & $<$ 1 minute & $>$ 2 days \\
          \bottomrule
        \end{tabular}
        
\end{table}


\clearpage

\bibliographystyle{unsrt}
\footnotesize
\bibliography{references_supp}